\documentclass[10pt,journal,compsoc]{IEEEtran}

\usepackage{graphicx}
\usepackage{amsmath}
\usepackage{amsfonts}
\usepackage{amsthm}
\usepackage{booktabs}
\usepackage{algorithm}
\usepackage{algorithmic}
\usepackage{subfigure}
\usepackage{epsfig}

\usepackage{microtype}
\usepackage{epsfig}
\usepackage[table,xcdraw]{xcolor}
\usepackage{float}
\usepackage{placeins}
\usepackage{color, colortbl}
\usepackage{stfloats}
\usepackage{enumitem}
\usepackage{tabularx}
\usepackage{xstring}
\usepackage{multirow}
\usepackage{xspace}
\usepackage{url}
\usepackage{xcolor}
\usepackage[pagebackref=true,breaklinks=true,letterpaper=true,colorlinks,bookmarks=false]{hyperref}

\begin{document}

\title{Explicit Visual Prompting for Universal Foreground Segmentations}

\author{Weihuang Liu, Xi Shen, Chi-Man Pun,~\IEEEmembership{Senior Member,~IEEE}, and~Xiaodong Cun

\thanks{

Weihuang Liu and Chi-Man Pun are with the Department of Computer and Information Science, Faculty of Science and Technology, University of Macau, Taipa, Macau~(e-mail: nifangbaage@gmail.com; cmpun@umac.mo).

Xi Shen and Xiaodong Cun are with Tencent AI Lab~(e-mail: shenxiluc@gmail.com; vinthony@gmail.com)

Corresponding author: Chi-Man Pun and Xiaodong Cun}
}

\hypersetup{
    colorlinks = true,
    linkbordercolor = {white},
    citecolor=blue
}

\IEEEtitleabstractindextext{%

\begin{abstract}

Foreground segmentation is a fundamental problem in computer vision, which includes salient object detection, forgery detection, defocus blur detection, shadow detection, and camouflage object detection.
Previous works have typically relied on domain-specific solutions to address accuracy and robustness issues in those applications.
In this paper, we present a unified framework for a number of foreground segmentation tasks without any task-specific designs.
We take inspiration from the widely-used pre-training and then prompt tuning protocols in NLP and propose a new visual prompting model, named Explicit Visual Prompting~(EVP). 
Different from the previous visual prompting which is typically a dataset-level implicit embedding, our key insight is to enforce the tunable parameters focusing on the explicit visual content from each individual image, \emph{i.e.}, the features from frozen patch embeddings and high-frequency components.
Our method freezes a pre-trained model and then learns task-specific knowledge using a few extra parameters.
Despite introducing only a small number of tunable parameters, EVP achieves superior performance than full fine-tuning and other parameter-efficient fine-tuning methods. 
Experiments in fourteen datasets across five tasks show the proposed method outperforms other task-specific methods while being considerably simple. 
The proposed method demonstrates the scalability in different architectures, pre-trained weights, and tasks.
The code is available at:~\url{https://github.com/NiFangBaAGe/Explicit-Visual-Prompt}.
\end{abstract}

}

\maketitle

\section{Introduction}
\label{sec:intro}

\IEEEPARstart{F}oreground segmentation is one of the fundamental research topics for scene understanding, with broad applications across other computer vision tasks. It divides the images into two non-overlapping sub-regions, which refer to the foreground and background, according to the intra-class similarities and inter-class differences. Foreground segmentation contains many sub-tasks, including detecting salient objects~\cite{liu2019simple,wei2020label,liu2021visual}, segmenting the manipulated parts~\cite{zhou2018learning,wu2017deep,cun2018image}, identifying out-of-focus pixels~\cite{zhao2019btbnet,tang2019defusionnet,cun2020defocus}, separating shadow regions~\cite{shi2014discriminative,zhu2018bidirectional,zheng2019distraction}, and discovering concealed objects~\cite{fan2020camouflaged,le2019anabranch,mei2021camouflaged}, \emph{etc}. Compared with image classification and object detection, these tasks provide more geometrically accurate descriptions of the targets, hence shown to be beneficial to numerous other computer vision tasks, including auto-refocus~\cite{bae2007defocus}, image retargeting~\cite{karaali2016image}, and object tracking~\cite{mikic2000moving}, \emph{etc}.

At present, these tasks are typically addressed by domain-specific solutions, which are designed carefully as unique network architectures to learn task-related characteristics. 
Meanwhile, the above tasks have the same input/output format and the same purpose that segments the foreground, resulting in similar encoder-decoder models. Hence, the designed model for segmenting specific targets can also be applied to others. 
Nevertheless, there is no unified framework to solve these similar problems yet.

We introduce a novel approach for foreground segmentation, which uses a unified framework to segment foregrounds with different characteristics according to task-specific knowledge.
 We take inspiration from recent advances of \textit{prompting}~\cite{vpt,chen2022adaptformer,bar2022visual}, which is a concept that initially emerged in natural language processing (NLP)~\cite{brown2020language}. The basic idea is to efficiently adapt a frozen large foundation model to many downstream tasks with the minimum extra trainable parameters. As the foundation model has already been trained on a large-scale dataset, prompting often leads to better model generalization on the downstream tasks~\cite{brown2020language}. Prompting also significantly saves the storage of models since it only needs to save a shared basic model and task-aware promptings.
In addition, we observe that the recent advances in downstream computer vision tasks usually fine-tune the pre-trained model for using the ability of scene understanding learned from large-scale datasets. 
These observations encourage us to develop a universal framework for foreground segmentations by learning a limited number of task-specific promptings. The model is trained on a large dataset to capture general visual knowledge, and task-specific promptings are learned to capture specific features related to each task. This approach allows the model to adapt to a wide range of foreground segmentation tasks by using a universal framework, thereby reducing the need for task-specific architectures. 
We take a model pre-trained on a large-scale dataset and freeze its parameters. 
To adapt to each task, we learn promptings by a few extra parameters.

Another insight comes from the effectiveness of hand-crafted image features, such as SIFT~\cite{huang2008detection}, illumination~\cite{okabe2009attached}, noise~\cite{luo2010jpeg}.
These features play important roles in traditional methods in foreground segmentations~\cite{sengottuvelan2008performance,pike2018quantifying,huang2008detection}.
Moreover, recent advances in deep-learning-based methods~\cite{park2017unified,zhou2018learning, wang2022objectformer} combine these hand-crafted features with learned features to improve performance.
Based on this observation, we propose \textit{explicit visual prompting~(EVP)}, 
which learns the task-specific knowledge from the features of each individual image itself.
Particularly, the tuning performance can be hugely improved via the re-modulation of image features.  Note that this is different from VPT~\cite{vpt}, which learns implicit prompts. 
Specifically, we consider two kinds of features for our task. 
The first is the features from the frozen patch embedding, which is critical since we need to shift the distribution of the original model.
Another is high-frequency features since the pre-trained model is learned to be invariant to these features via data augmentation. 


Our previous conference work~\cite{liu2023explicit}~(EVPv1) has shown the effectiveness of explicit visual prompting. EVPv1 first tunes the frozen patch embedding features with a linear layer. Then, it applies Fast Fourier Transform~(FFT) to the input image and a fixed mask to the spectrum to extract the high-frequency components~(HFC) and learns an extra embedding for HFC. Finally, the Adaptor performs adaptation in all the layers by considering features from the image embeddings and HFC. EVPv1 shows superior performance against other parameter-efficient fine-tuning methods and task-specific methods in different tasks.

Despite these advancements, EVPv1 still requires manual extraction of high-frequency components with appropriate parameters.
In this work, we follow the idea of explicit visual prompting and provide an end-to-end solution, EVPv2. We introduce Fourier MLP, which learns high-frequency features directly from the image embeddings. Fourier MLP performs FFT for the image embeddings, then generates the adaptive mask for the spectrum and reconstructs the high-frequency features. The image embeddings and high-frequency features are fed to the Frequency Enhanced Adaptor to generate promptings and attached to the corresponding transformer layers. 
In addition, this paper provides more discussions and results in terms of target task, pre-trained model, and prompt understanding.

Comprehensive experiments are conducted to validate the effectiveness of the proposed approach. 
We evaluate it on a total of fourteen different datasets, covering five different tasks that are critical in foreground segmentation, which include salient object detection, forgery detection, shadow detection, defocus blur detection, and camouflaged object detection. Our unified framework is compared to task-specific solutions to show its superiority. 
The result also indicates that our method outperforms other parameter-efficient fine-tuning approaches and full fine-tuning.

In conclusion, 
the proposed method offers a unified framework for foreground segmentations, providing a new perspective for reducing the need for task-specific architectures and improving the reusability of the framework to the related tasks. 
It delivers a commendable performance that is better than other parameter-efficient fine-tuning methods and full fine-tuning.
This simple approach shows comparable or even better performance than other complicated task-specific models.
The results provide strong pieces of evidence of its potential to address a broad range of problems.
We believe our simple but effective framework will be widely used in various foreground segmentation tasks, and play an important role in developing unified frameworks for other computer vision tasks.


In summary, our main contributions are as follows:
\begin{itemize}
    \item To the best of our knowledge, we are the first to design a unified approach for a number of foreground segmentation tasks, including salient object detection, forgery detection, defocus blur detection, shadow detection, and camouflaged object detection. 
    
    \item We propose explicit visual prompting~(EVP), which takes the frozen patch embedding and high-frequency features as prompting. 
    It outperforms other parameter-efficient tuning methods and task-specific methods. 
    
    \item The proposed method shows the scalability in different architectures, pre-trained models, and foreground segmentation tasks.
    
\end{itemize}
\section{Related Work} 
\label{sec:related}

\subsection{Visual Prompting Tuning} Prompting is initially proposed in NLP~\cite{brown2020language,liu2021pre}. Brown~\emph{et al.}~\cite{brown2020language} demonstrates strong generalization to downstream transfer learning
tasks even in the few-shot or zero-shot settings with manually chosen
prompts in GPT-3. Recently, prompting has been adapted to vision tasks~\cite{sandler2022fine,bahng2022exploring,vpt}. Sandler~\emph{et al.}~\cite{sandler2022fine} proposes memory tokens which is a set of learnable embedding vectors for each transformer layer. 
Bahng~\emph{et al.}~\cite{bahng2022exploring} add learnable pixels in the input pixel space.
VPT~\cite{vpt} proposes similar ideas and investigates the generality and feasibility of visual prompting via extensive experiments spanning multiple kinds of recognition tasks across multiple domains and backbone architectures. 
Unlike the above methods whose main focuses are on recognition tasks, our work aims at exploring optimal visual content for foreground segmentation. 

\subsection{Salient Object Detection} Salient object detection aims at detecting visually salient object regions in images. Classical approaches utilize low-level clues such as color, luminance, and texture for salient object detection~\cite{zhang2017salient,scharfenberger2013statistical,achanta2009frequency}. 
Recently, deep learning-based methods~\cite{liu2019simple,wei2020label,yun2022selfreformer,ma2023boosting,liu2021visual} have shown superior performance than traditional approaches.
PoolNet~\cite{liu2019simple} designs a global guidance module and feature aggregation module for pyramid pooling.
LDF~\cite{wei2020label} decouples the saliency map into the body map and detail map to address the issue of the unbalanced edge pixels.
VST~\cite{liu2021visual} first employs transformer for salient object detection, which constructs patch-level features and models long-range dependencies via self-attention.
Ma~\emph{et al.}~\cite{ma2023boosting} splits the global semantic and local detail information to capture large and small objects.

\subsection{Forgery Detection} The goal of forgery detection is to detect pixels that are manually manipulated, such as pixels that are removed, replaced, or edited. Early approaches~\cite{mahdian2009using,fridrich2012rich} detect region splicing through inconsistencies in local noise levels, based on the fact that images of different origins might contain different noise characteristics introduced by the sensors or post-processing steps. Other clues are found to be helpful, such as SIFT~\cite{huang2008detection}, JPEG compression artifacts~\cite{luo2010jpeg} and re-sampling artifacts~\cite{feng2012normalized}. Recently, approaches have moved towards end-to-end deep learning methods for solving specific forensics tasks using labeled training data~\cite{zhou2018learning,hao2021transforensics,wu2017deep,psccnet,hu2020span}. 
Recently, TransForensic~\cite{hao2021transforensics} leverages vision transformers~\cite{dosovitskiy2020image} to tackle the problem. High-frequency components still served as useful prior in this field. RGB-N~\cite{zhou2018learning} designs an additional noise stream. ObjectFormer~\cite{wang2022objectformer} extracts high-frequency features as complementary signals to visual content. But unlike ObjectFormer, our main focus is to leverage high-frequency components as a prompting design to efficiently and effectively adapt to different low-level segmentation tasks.

\subsection{Defocus Blur Detection} Given an image, defocus blur detection aims at separating in-focus and out-of-focus regions, which could be potentially useful for auto-refocus~\cite{bae2007defocus}, salient object detection~\cite{jiang2013salient} and image retargeting~\cite{karaali2016image}. Traditional approaches mainly focus on designing hand-crafted features based on gradient~\cite{shi2014discriminative} or edge~\cite{karaali2017edge}. In the deep era, most methods delve into CNN architectures~\cite{park2017unified,zhao2019btbnet,tang2019defusionnet,zhao2019cenet}. Park~\emph{et al.}~\cite{park2017unified} proposes the first CNN-based method using both hand-crafted and deep features. BTBNet~\cite{zhao2019btbnet} develops a fully convolutional network to integrate low-level clues and high-level semantic information. DeFusionNet~\cite{tang2019defusionnet} recurrently fuses and refines multi-scale deep features for defocus blur detection. CENet~\cite{zhao2019cenet} learns multiple smaller defocus blur detectors and ensembles them to enhance diversity. Cun~\emph{et al.}~\cite{cun2020defocus} further employs the depth information as additional supervision and proposes a joint learning framework inspired by knowledge distillation. Zhao~\emph{et al.}~\cite{zhao2021defocus} explores deep ensemble networks for defocus blur detection.
DAD~\cite{zhao2021self} proposes to learn a generator to generate masks in an adversarial manner.

\subsection{Shadow Detection} Shadows occur frequently in natural scenes, and have hints for scene geometry~\cite{okabe2009attached}, light conditions~\cite{okabe2009attached} and camera location ~\cite{junejo2008estimating} and lead to challenging cases in many vision tasks. 
Early attempts explore illumination~\cite{finlayson2005removal,finlayson2009entropy} and hand-crafted features~\cite{huang2011characterizes,lalonde2010detecting}. In the deep era, some methods mainly focus on the design of CNN architectures~\cite{zhu2018bidirectional,cun2020towards} or involving the attention modules~(\emph{e.g.}, the direction-aware attention~\cite{hu2018direction}, distraction-aware module~\cite{zheng2019distraction}). Recent works utilize the lighting as additional prior, for example, ADNet~\cite{le2018a+} generates the adversarial training samples for better detection and FDRNet~\cite{zhu2021mitigating} arguments the training samples by additionally adjusted brightness. MTMT~\cite{mtmt} leverages the mean teacher model to explore unlabeled data for semi-supervised shadow detection.

\subsection{Camouflaged Object Detection} 
Detecting camouflaged objects is a challenging task as foreground objects are often with visual similar patterns to the background. Early works distinguish the foreground and background through low-level clues such as texture~\cite{sengottuvelan2008performance}, brightness~\cite{pike2018quantifying}, and color~\cite{hou2011detection}. Recently, deep learning-based methods~\cite{fan2020camouflaged,mei2021camouflaged,li2021uncertainty,lv2021simultaneously,jiaying2022frequency} show their strong ability in detecting complex camouflage objects. Le~\emph{et al.}~\cite{le2019anabranch} proposes the first end-to-end network for camouflaged object detection, which is composed of a classification branch and a segmentation branch. Fan \emph{et al.}~\cite{fan2020camouflaged} develops a search-identification network and the largest camouflaged object detection dataset. PFNet~\cite{mei2021camouflaged} is a bio-inspired framework that mimics the process of positioning and identification in predation. FBNet~\cite{jiaying2022frequency} suggests disentangling frequency modeling and enhancing the important frequency component.

\subsection{Universal Image Segmentation} 
Various computer vision tasks were divided and thousands of task-specific solutions were born in the past decades.
Recent advances~\cite{cheng2022masked,zhang2021k,jain2022oneformer,bar2022visual,wang2022images,kirillov2023segment} have shown that similar tasks can be solved within a unified framework.
Mask2Former~\cite{cheng2022masked} and K-Net\cite{zhang2021k} unify semantic, instance, and panoptic segmentation in a single framework. 
OneFormer~\cite{jain2022oneformer} designs a task-conditioned joint training strategy that enables a single model to perform these three tasks.
In-context learning~\cite{brown2020language} based methods~\cite{bar2022visual,wang2022images} perform a specific task by providing task examples, hence can handle multiple kinds of tasks.
SAM~\cite{kirillov2023segment} design a prompt encoder that enables the model to segment anything interactively and has a strong few-shot segmentation performance by training on a huge dataset.



\section{Method}
\label{sec:method}
We propose Explicit Visual Prompting~(EVP) for adapting pre-trained Vision Transformers to foreground segmentations. 
EVP keeps the backbone frozen and only contains a small number of tunable parameters to learn task-specific knowledge from the features of each individual
image itself. 
The overview of the proposed model is shown in Figure~\ref{fig:overview}. 
Below, we first present Vision Transformer and Visual Prompting in Section~\ref{Preliminary}.
Then, we describe the proposed Explicit Visual Prompting in Section~\ref{sec:explicit_visual_prompting_for_foreground_segmentation}, as well as the details of two variants in Section~\ref{sec:explicit_visual_prompting_with_adaptor} and Section~\ref{sec:explicit_visual_prompting_with_frequency_enhanced_adaptor}.

\subsection{Preliminaries}
\label{Preliminary}

\subsubsection{Vision Transformer}
Vision Transformer is first proposed in~\cite{dosovitskiy2020image}. 
A plain Vision Transformer consists of a patch embedding layer and some transformer blocks.
The input image $I \in \mathbb{R}^{H \times W \times 3}$ is first divided into several patches $p \in \mathbb{R}^{N \times h \times w \times 3}$, where $H$ and $W$ represents the height and width of the input image, $h$ and $w$ is the height and width of the patch, and $N = \frac{H\times W}{h\times w}$ is the number of patches.
The patches are then flattened and projected into $d$ dimensional tokens $x \in \mathbb{R}^{N \times d}$ by the patch embedding layer.
An extra learnable classification token [CLS] and positional encoding are combined with the embedded tokens and fed to the transformer blocks.
Each transformer block is composed of the Multi-head Self-Attention~(MSA) block and the Multi-Layer Perceptron~(MLP) block.
In the MSA block, the tokens are mapped into three vectors, namely Q, K, and V, through a linear transformation. Then, the self-attention calculation is performed by calculating the scaled dot-product attention of each token:
\begin{equation}
x=\mathtt{Attention}(Q,K,V)=\mathtt{Softmax}(QK^{T}/\sqrt{d})V.
\end{equation}

The output tokens $x$ are further sent to the MLP block which consists of two linear layers and a $\mathtt{GELU}$ activation~\cite{hendrycks2016gaussian}, which can be formulated as:
\begin{equation}
x=\mathtt{MLP}(\mathtt{LN}(x))+x,
\end{equation}
where $\mathtt{LN}$ represents LayerNorm~\cite{ba2016layer}.

After several transformer blocks, [CLS] is used for final vision recognition.

SegFormer~\cite{xie2021segformer} is a hierarchical transformer-based structure with a much simpler decoder for semantic segmentation. Similar to the traditional CNN backbone, SegFormer captures multi-stale features via several stages. Differently, each stage is built via the feature embedding layers\footnote{SegFormer has a different definition of patch embedding in ViT~\cite{dosovitskiy2020image}. It uses the overlapped patch embedding to extract the denser features and will merge the embedding to a smaller spatial size at the beginning of each stage.} and vision transformer blocks~\cite{dosovitskiy2020image}. As for the decoder, it leverages the multi-scale features from the encoder and MLP layers for decoding to the specific classes.

\subsubsection{Visual Prompting}
Prompting~\cite{brown2020language} has emerged as a powerful approach for fine-tuning models that are pre-trained on large-scale datasets to downstream tasks, without the need to modify the original model weights. 
The pre-trained model is guided to perform a specific task by introducing a set of instructions.
Recently, visual prompting~\cite{vpt,bahng2022exploring,bar2022visual}, which refers to the introduction of additional input to the input space, has emerged as a novel approach and has aroused considerable attention in computer vision.
This supplementary input can be the tokens, images, or other forms of media, and can provide context for downstream tasks that require additional information. 
This novel approach has garnered significant interest since it can reduce the computational resources necessary for fine-tuning while simultaneously maintaining model performance.

\begin{figure*}[!t]
    \centering
    \includegraphics[width=\linewidth]{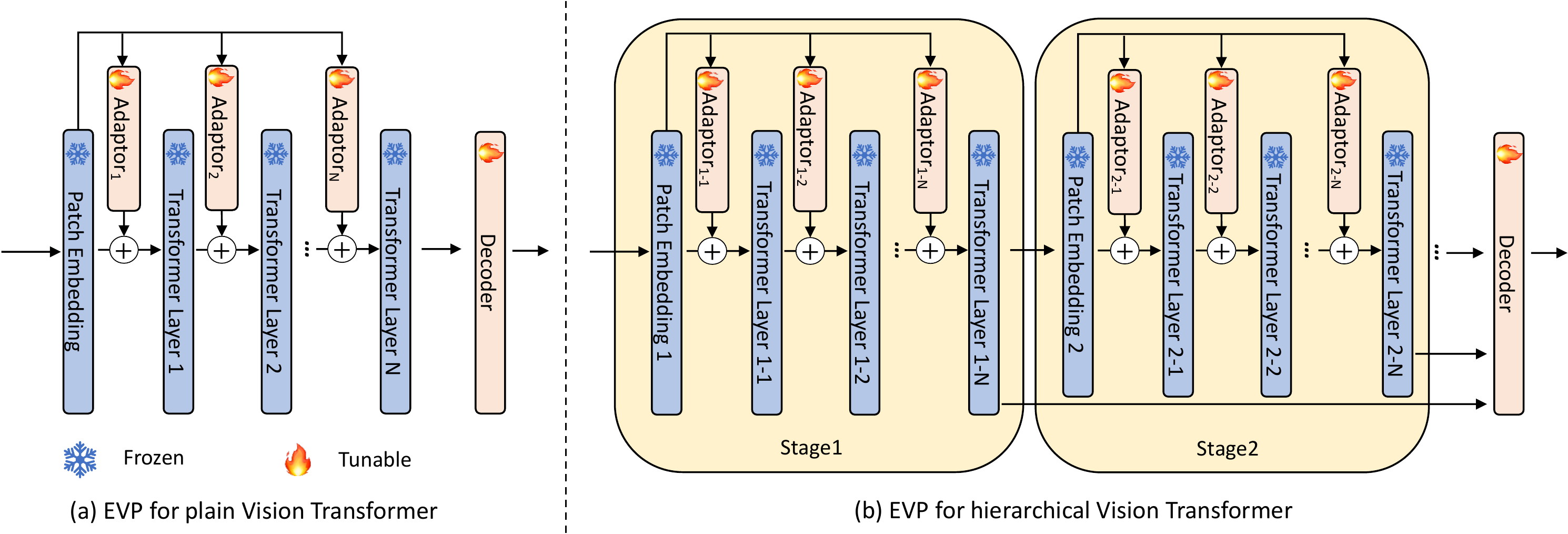}
    \vspace{-1em}
    \caption{Overview of the proposed Explicit Visual Prompting~(EVP) for universal foreground segmentation.
    EVP is applicable to both plain Vision Transformer~(a) and hierarchical Vision Transformer~(b).
    EVP learns explicit prompts for each individual image and attaches the prompts to the transformer blocks in the prompting fashion.
    }
    \label{fig:overview}
\end{figure*}

\subsection{Explicit Visual Prompting for Foreground Segmentation}
\label{sec:explicit_visual_prompting_for_foreground_segmentation}
Typically, a collection of $K$ different foreground segmentation tasks can be solved with $K$ task-specific methods:
\begin{align}
	M_{pred}^k = \mathtt{\phi}^k(I^k), \quad k=1,2,...,K
\end{align}
where $\mathtt{\phi}$ denotes the task-specific architectues and $M_{pred}$ is the prediction.
The models are fully fine-tuned in each task to adapt to these tasks.

We present Explicit Visual Prompting~(EVP) for foreground segmentation. EVP is a universal framework that can handle various foreground segmentation tasks without any task-specific design.
Formally, the proposed explicit visual prompting framework 
can be written as:
\begin{align}
	M_{pred}^k = \mathtt{\phi}_{pt}(I^k) + \mathtt{\phi}_{vp}^k(I^k), \quad k=1,2,...,K
\end{align}
where $\mathtt{\phi}_{pt}$ denotes the frozen pre-trained backbone and $\mathtt{\phi}_{vp}$ is a set of tunable parameters. 
EVP handles various tasks using a common frozen backbone by learning explicit prompts for each individual image with a limited number of tunable parameters.

Our key insight is to learn explicit prompts from the image embeddings and frequency domain.
We learn the former to shift the distribution from the pre-train dataset to the target dataset. 
And the motivation to learn the latter is that the pre-trained model is learned to be invariant to high-frequency features through data augmentation.
In each transformer block, we generate the prompt according to the input image and attach it to the transformer features for further processing.
As shown in Figure~\ref{fig:overview}, the proposed method is scalable for different vision transformer architectures.
For the single-stage Vision Transformer architecture, we use the patch embedding layer to generate prompts.
As for the multi-stage Vision Transformer architecture, the prompts at different stages are generated by the patch embedding layer at the corresponding stage.

Below, we introduce two variants. The first is Explicit Visual Prompting with Adaptor. It manually separates the high-frequency components~(HFC) from the input image and learns the extra patch embedding layer for HFC, followed by the lightweight Adaptor to efficiently generate the prompts. Another one is Explicit Visual Prompting with Frequency Enhanced Adaptor. It automatically extracts frequency features from the image embeddings using the proposed Fourier MLP and then generates the prompts with the Frequency Enhanced Adaptor.

\begin{figure}[!t]
    \centering
    \includegraphics[width=\linewidth]{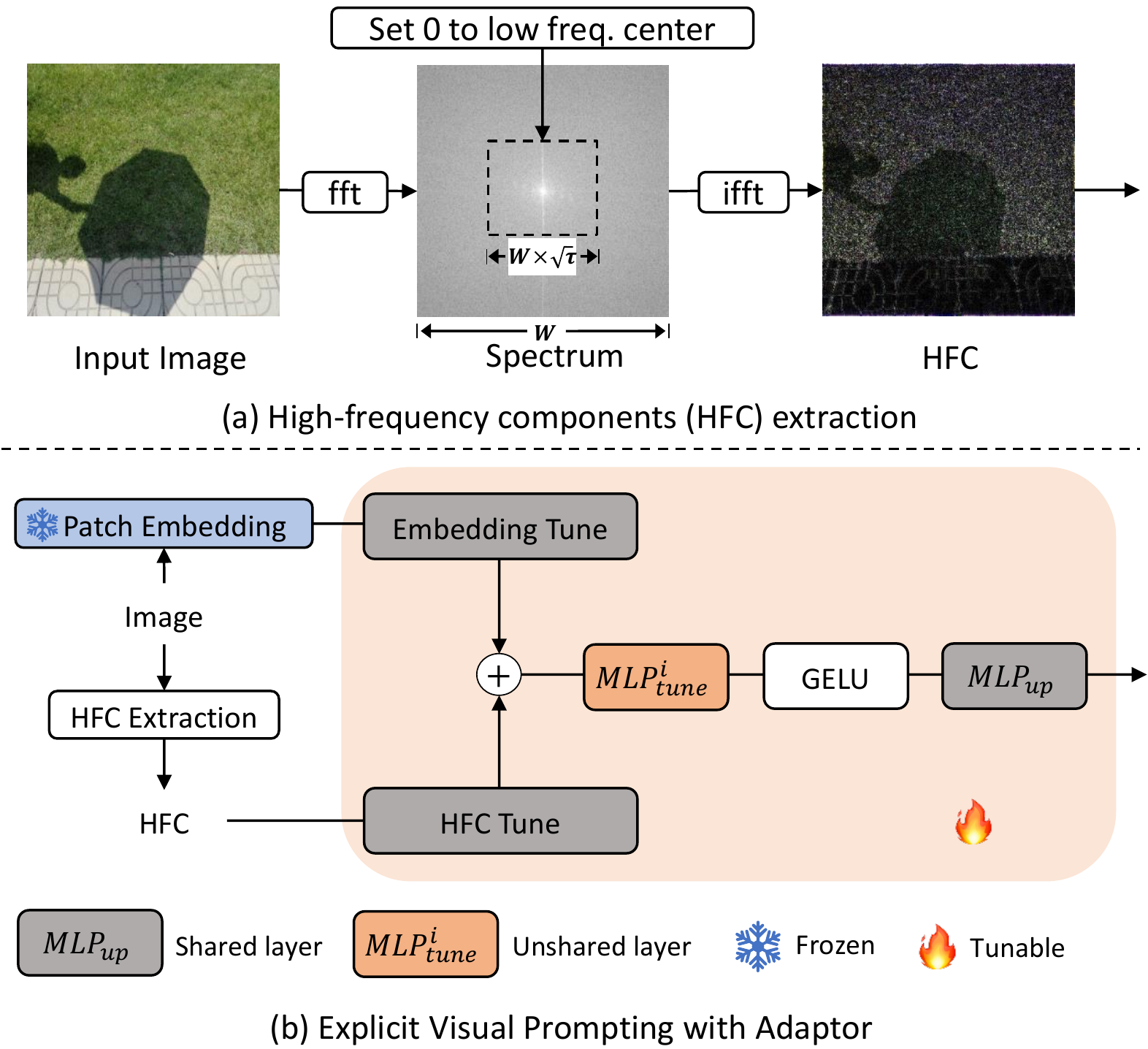}
    \vspace{-1em}
    \caption{
    The procedure to generate high-frequency components~(top). We apply a fixed mask to the spectrum of the input image to obtain the high-frequency components.
    The architecture of the proposed Explicit Visual Prompting with Adaptor~(bottom). We use the \textit{Embedding Tune} and \textit{HFC Tune} to tune the extracted features. The Adaptor is designed to merge these features.}
    \label{fig:adaptor}
\end{figure}

\subsubsection{Explicit Visual Prompting with Adaptor}
\label{sec:explicit_visual_prompting_with_adaptor}
In this section, we present the proposed Explicit Visual Prompting with Adaptor. 
The proposed method is shown in Figure~\ref{fig:adaptor}.
Below, we describe the method with three modules: patch embedding tune, high-frequency components~(HFC) tune as well as Adaptor. The patch embedding tune and high-frequency components tune aims to initialize the embeddings for Adaptor, while Adaptor is used to generate prompts for different transformer layers efficiently.

\textbf{Patch embedding tune.} This module aims at tuning the pre-trained patch embeddings. 
In the pre-trained Vision Transformer,  the input image is projected to $d$-dimension embeddings $x \in \mathbb{R}^{N \times d}$
via the frozen patch embedding layer. We freeze this projection and add a tunable linear layer $\mathtt{L_{pe}}$ to project the original embeddings into $c$-dimension features $F_{pe} \in \mathbb{R}^{N \times c}$. 

\begin{align}
	F_{pe} = \mathtt{L_{pe}}(x) \text{, with } c = \frac{d}{r} \label{eqn:fpe}
\end{align}
where we introduce the scale factor $r$ to control the tunable parameters.

\textbf{High-frequency components tune.} 
\label{sec:high_frequency_extraction}
This module aims at tuning the pre-trained model in the frequency domain. We first extract the high-frequency components and then learn the extra patch embedding layer.

As shown in Figure~\ref{fig:adaptor}-a, for an image $I$, we can decompose it into low-frequency components $I_l$ (LFC) and high-frequency components $I_h$ (HFC), \textit{i.e.} $I = \{I_l, I_h\}$. Denoting $\mathtt{fft}$ and $\mathtt{ifft}$ as the Fast Fourier Transform~(FFT) and its inverse respectively, we use $z$ to represent the frequency component of $I$. Therefore we have $z = \mathtt{fft}(I)$ and $I = \mathtt{ifft}(z)$. We shift low frequency coefficients to the center $(\frac{H}{2}, \frac{W}{2})$. To obtain HFC, a binary mask $\mathbf{M}_h \in \{0, 1\}^{H \times W}$ is generated and applied on $z$ depending on a mask ratio $\tau$: 

\begin{equation}
\mathbf{M}^{i,j}_h(\tau) = \begin{cases} 
	0, & \frac{4|(i - \frac{H}{2})(j - \frac{W}{2})|}{HW} \le \tau\\ 
	1, & \text{otherwise} 
\end{cases}
\end{equation}
$\tau$ indicates the surface ratio of the masked regions. HFC can be computed:
\begin{align}
	I_{hfc} =  \mathtt{ifft}(z \mathbf{M}_h(\tau))
\end{align}


Note that for RGB images, we compute the above process on every channel of pixels independently.

For the high-frequency components $I_{hfc}$, we learn the extra patch embedding layer similar to the backbone. Formally,
$I_{hfc}$ is divided into small patches with the same patch size as backbone. Denoting these patches $p_{hfc} \in \mathbb{R}^{N \times h \times w \times 3}$, we learn a linear layer $\mathtt{L_{hfc}}$ to project these patches into a $c$-dimension feature $F_{hfc} \in \mathbb{R}^{N \times c}$. 

\begin{align}
	F_{hfc} =  \mathtt{L_{hfc}}(p_{hfc})
\end{align}

\textbf{Adaptor.} The goal of Adaptor is to efficiently and effectively perform adaptation in all the layers by considering features from the image embeddings and high-frequency components.
For the $i$-th Adaptor, we take $F_{pe}$ and $F_{hfc}$ as input and obtain the prompting $P^i$: 
\begin{align}
   	P^i & =  \mathtt{MLP_{up}}(\mathtt{GELU}(\mathtt{MLP^i_{tune}}(F_{pe}+F_{hfc}))) 
\end{align}
where $\mathtt{GELU}$ is GELU activation. $\mathtt{MLP^i_{tune}}$ is a linear layer for producing different prompts in each Adaptor. $\mathtt{MLP_{up}}$ is an up-projection layer shared across all the Adaptors for matching the dimension of transformer features. $P^i$ is the output prompting that attaches to the corresponding transformer layer.

\begin{figure}[!t]
    \centering
    \includegraphics[width=0.9\linewidth]{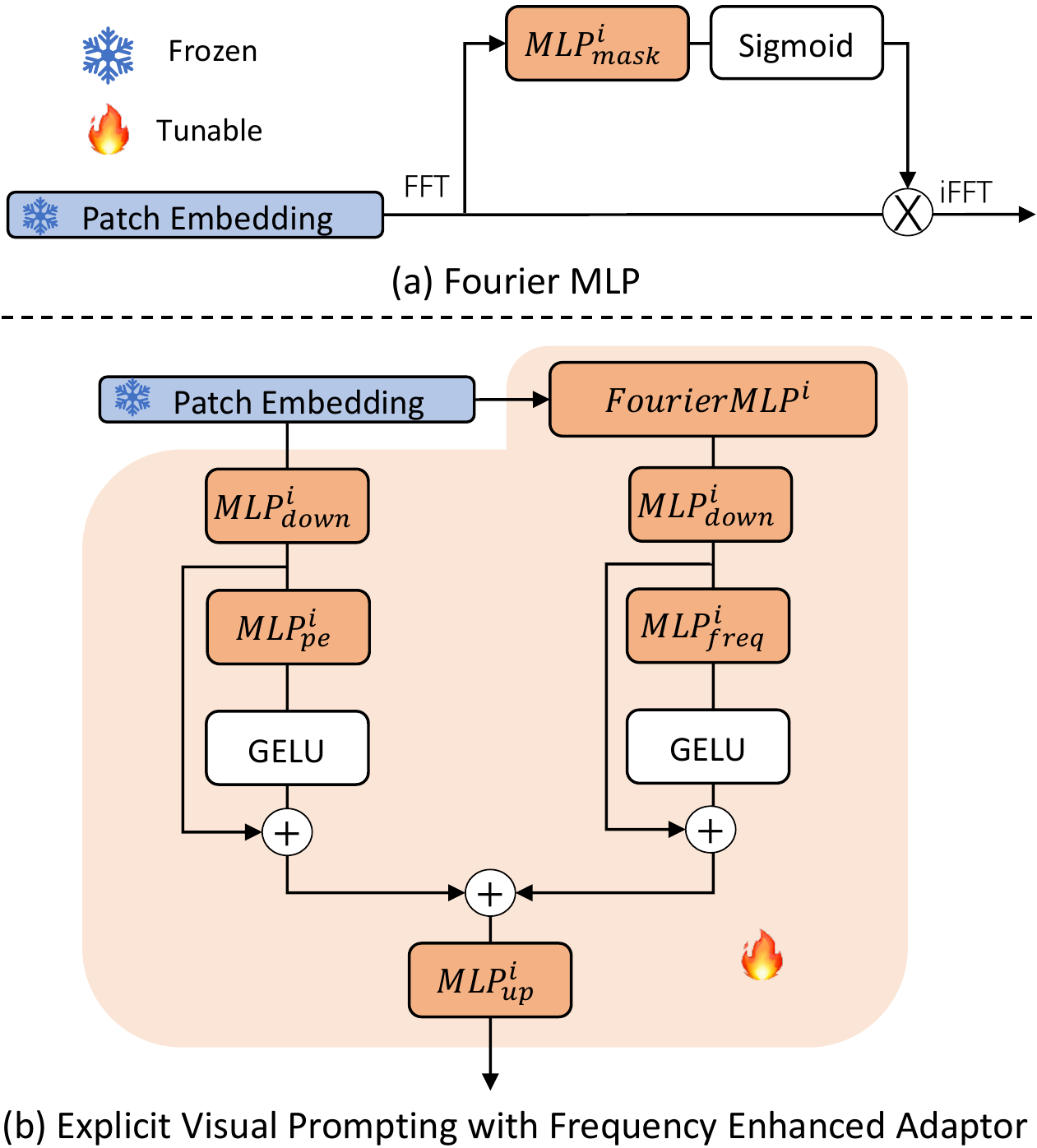}
    \vspace{-1em}
    \caption{
    The proposed Fourier MLP~(top). We perform Fast Fourier Transform for the image embeddings and generate the adaptive mask for the spectrum.
    The architecture of the proposed Explicit Visual Prompting with Frequency Enhanced Adaptor~(bottom).
    We use Fourier MLP to extract frequency features from the image embeddings and train the Frequency Enhanced Adaptor to obtain prompts.}
    \label{fig:freq_enhanced_adaptor}
\end{figure}

\subsubsection{Explicit Visual Prompting with Frequency Enhanced Adaptor}
\label{sec:explicit_visual_prompting_with_frequency_enhanced_adaptor}

In this section, we present the proposed Explicit Visual Prompting based on Frequency Enhanced Adaptor. 
Compared with the above one, we design Fourier MLP which learns frequency features directly from the image embeddings, yielding an end-to-end high-frequency feature extraction solution. The detail is shown in Figure~\ref{fig:freq_enhanced_adaptor}.


\textbf{Fourier MLP.} We design the Fourier MLP for feature extraction in the frequency domain.
Given the $d$-dimension embeddings $x \in \mathbb{R}^{N \times d}$ obtained by the patch embedding layer, 
we first perform $\mathtt{fft}$ to $x$:
\begin{equation}
x_{real}, x_{imag} = \mathtt{fft}(x)
\end{equation}
where $x_{real}$ and $x_{imag}$ denote the real part and imaginary part of $x$, respectively. 
Then, we generate the masks using two MLPs:
\begin{equation}
\begin{split}
M_{real} & = \mathtt{Sigmoid}(\mathtt{MLP_{real}}(x_{real})),\\
M_{\mathtt{imag}} & = \mathtt{Sigmoid}(\mathtt{MLP_{imag}}(x_{imag}))
\end{split}
\end{equation}
where $\mathtt{Sigmoid}$ is the sigmoid activation.
Finally, we reconstruct the real part and imaginary part via the corresponding masks and apply $\mathtt{ifft}$ to obtain the features in the frequency domain:
\begin{equation}
x_{freq} = \mathtt{ifft}(x_{real} \times M_{real}, x_{imag} \times M_{imag})
\end{equation}

\textbf{Frequency Enhanced Adaptor.}
We design Frequency Enhanced Adaptor to generate visual prompts without the necessity to manually extract high-frequency components and adjust the mask ratio in Section~\ref{sec:high_frequency_extraction}.

For the $i$-th Frequency Enhanced Adaptor, we take $x$ as input. We first generate the $i$-th frequency features $x_{freq}^{i}$ via the i-th Fourier MLP $\mathtt{FMLP^{i}}$: 
$x_{freq}^{i} = \mathtt{FMLP^{i}}(x)$.
We utilize a down-projection layer with parameters $\boldsymbol{W}_{down} \in \mathbb{R}^{{d} \times \frac{d}{r}}$ to $x$ and $x_{freq}^i$:
\begin{equation}
\begin{split}
x^{i} = \mathtt{MLP_{down}^i}(x),\\
x_{freq}^{i} = \mathtt{MLP_{down}^i}(x_{freq}^{i})
\end{split}
\end{equation}

The output embeddings are then sent to two light-weight MLPs:
\begin{equation}
\begin{split}
x^{i} = \mathtt{GELU(MLP_{pe}^i}(x)) + x^{i},\\
x_{freq}^{i} = \mathtt{GELU(MLP_{freq}^i}(x_{freq}^{i})) + x_{freq}^{i}
\end{split}
\end{equation}

Next, an up-projection layer with parameters $\boldsymbol{W}_{up} \in \mathbb{R}^{{\frac{d}{r}} \times {d}}$ obtains the output promptings:
\begin{align}
P^i & =  \mathtt{MLP_{up}^i}(x^{i} + x_{freq}^{i})
\end{align}
Finally, $P^i$ is attached to the corresponding transformer layer.

\section{Experiment}
\label{sec:exp}
This section shows the experimental results for foreground segmentation of our method and other state-of-the-art methods. 
Firstly, Section~\ref{sec:dataset_and_metric} introduces all the datasets and metrics for comparisons. 
Then, we introduce the details of the experiment in Section~\ref{sec:setting}.
We conduct both quantitative and qualitative comparisons with state-of-the-art methods in Section~\ref{sec:main_results}.
Finally, we verify the effectiveness of each component of the proposed method in Section~\ref{sec:ablation_study}.

\begin{table}
\caption{Summary of datasets considered in this work. We show the number of images in training (\textit{\# Train}) and testing set (\textit{\# Test}) for different datasets.}
\centering
\normalsize
\resizebox{\columnwidth}{!}
{
\begin{tabular}{c|c|c|c}
\toprule
Task                     & Dataset Name     & \# Train & \# Test \\ \hline

\multirow{2}{*}{\shortstack{ \\ \\ \\ Salient \\ Object \\ Detection}} & DUTS~\cite{wang2017learning}    & 10,553  & 5,019   \\ \cline{2-4} 
& ECSSD~\cite{shi2015hierarchical}    & -  & 1,000  \\ \cline{2-4} 
& HKU-IS~\cite{li2015visual}    & -  & 4,447  \\ \cline{2-4} 
& PASCAL-S~\cite{li2014secrets}     & -  & 850  \\ \cline{2-4} 
& DUT-OMRON~\cite{yang2013saliency}    & -  & 5,168  \\ \hline

\multirow{2}{*}{\shortstack{Forgery \\Detection}} & CAISA~\cite{dong2013casia}    & 5,123  & 921   \\ \cline{2-4} 
& IMD20~\cite{novozamsky2020imd2020}    & -  & 2,010  \\ \hline

\multirow{2}{*}{\shortstack{Shadow \\Detection}} 
& ISTD~\cite{wang2018stacked}     & 1,330  & 540  \\ \cline{2-4} 
                         & SBU~\cite{sbu}      & 4,089  & 638  \\ \hline
                         
\multirow{2}{*}{\shortstack{Defocus Blur \\Detection}} & CUHK~\cite{shi2014discriminative}     & 604   & 100  \\ \cline{2-4} 
                         &  DUT~\cite{zhao2018defocus}      & -     & 500  \\ \hline
                         
\multirow{3}{*}{ \shortstack{Camouflaged \\ Object Detection}} & COD10K~\cite{fan2020camouflaged}     & 3,040   & 2,026  \\ \cline{2-4} 
& CAMO~\cite{le2019anabranch}    & 1,000     & 250  \\ \cline{2-4}
&  CHAMELEON ~\cite{skurowski2018animal}      & -     & 76  \\ \bottomrule

\end{tabular}
}
\label{tab:dataset}
\end{table}

\subsection{Datasets and Metrics}
\label{sec:dataset_and_metric}
We evaluate our model on a variety of datasets across five tasks: salient object detection, forgery detection, shadow detection, defocus blur detection, and camouflaged object detection. A summary of the basic information of these datasets is illustrated in Table~\ref{tab:dataset}.

\textbf{Salient Object Detection.} 
We evaluate the proposed method on five commonly used datasets for salient object detection, including DUTS~\cite{wang2017learning}, ECSSD~\cite{shi2015hierarchical}, HKU-IS~\cite{li2015visual}, PASCAL-S~\cite{li2014secrets}, and DUT-OMRON~\cite{yang2013saliency}.
DUTS is a large benchmark composed of 10,553 training samples~(DUTS-TR) and 5,019 testing samples~(DUTS-TE).
ECSSD contains 1,000 images of large objects.
HKU-IS has 4,447 challenging images which contain multiple foregrounds.
PASCAL-S involves 850 images from the PASCAL VOC2010~\cite{everingham2010pascal} validation set.
DUT-OMRON has 5,168 images with complex backgrounds.
Following the previous work, we train the model on DUTS-TR and test on other datasets with $352 \times 352$ images. 
S-measure~($S_{m}$), mean E-measure~($E_\phi$), maximum F-measure~($F_{\beta}^{m}$), and mean absolute error~(MAE) are reported for evaluation.

\textbf{Forgery Detection.} CASIA~\cite{dong2013casia} is a large dataset for forgery detection, which is composed of 5,123 training and 921 testing spliced and copy-moved images. 
IMD20~\cite{novozamsky2020imd2020} is a real-life forgery image dataset that consists of 2,010 samples for testing.
We follow the protocol of previous works~\cite{liu2022pscc, hao2021transforensics,wang2022objectformer} to conduct the training and evaluation at the resolution of $256 \times 256$. We use pixel-level Area Under the Receiver Operating Characteristic Curve~(AUC) and $F_{1}$ score to evaluate the performance.

\textbf{Shadow Detection.} SBU~\cite{sbu} is the largest annotated shadow dataset which contains 4,089 training and 638 testing samples, respectively.
ISTD~\cite{wang2018stacked} contains triple samples for shadow detection and removal, we only use the shadowed image and shadow mask to train our method.
Following \cite{mtmt,zhu2018bidirectional,zhu2021mitigating}, we train and test both datasets with the size of $400 \times 400$. 
As for the evaluation metrics, We report the balance error rate~(BER).

\textbf{Defocus Blur Detection.} 
Following previous work~\cite{zhao2018defocus, cun2020defocus}, we train the defocus blur detection model in the CUHK dataset~\cite{shi2014discriminative}, which contains a total of 704 partial defocus samples. We train the network on the 604 images split from the CUHK dataset and test in DUT~\cite{zhao2018defocus} and the rest of the CUHK dataset. 
The images are resized into $320 \times 320$, following~\cite{cun2020defocus}. We report performances with commonly used metrics: F-measure~($F_{\beta}$) and MAE. 

\textbf{Camouflaged Object Detection.} 
COD10K~\cite{fan2020camouflaged} is the largest dataset for camouflaged object detection, which contains 3,040 training and 2,026 testing samples. CHAMELEON~\cite{skurowski2018animal} includes 76 images collected from the Internet for testing. CAMO~\cite{le2019anabranch} provides diverse images with naturally camouflaged objects and artificially camouflaged objects. Following~\cite{fan2020camouflaged,mei2021camouflaged}, we train on the combined dataset and test on the three datasets. We employ commonly used metrics: S-measure~($S_{m}$), mean E-measure~($E_\phi$), weighted F-measure~($F_\beta^w$), and MAE for evaluation. 

\textbf{Metrics.} 
AUC calculates the area of the ROC curve. ROC curve is a function of true positive rate~($\frac{tp}{tp + fn}$) in terms of false positive rate~($\frac{fp}{fp + tn}$), where $tp$, $tn$, $fp$, $fn$ represent the number of pixels which are classified as true positive, true negative, false positive, and false negative, respectively. 

$F_{1}$ score is defined as $F_{1} = \frac{2 \times precision \times recall}{precision + recall}$, where $precision = \frac{tp}{tp + fp}$ and $recall = \frac{tp}{tp + fn}$.

Balance error rate~(BER) $ = \left(1-\frac{1}{2}\left(\frac{tp}{tp+fn}+\frac{tn}{tn+fp}\right)\right) \times 100$.

Weighted F-measure ($F_\beta^w$) weighting the quantities TP, TN, FP, and FN according to the errors at their location and their neighborhood information:
$F_\beta^w = \frac{\left(1+\beta^2\right) \times precision^w \times recall^w}{\beta^2 \times precision^w + recall^w}$.

S-measure~($S_\alpha$) evaluates the structural similarity between the prediction map and the ground truth map, which considers object-aware ($S_o$) and region-aware ($S_r$) structure similarities:
$ S_\alpha=\alpha \times S_o + (1 - \alpha) \times S_r$,
where $\alpha$ is set to 0.5.

E-measure~($E_\phi$) jointly considers image statistics and local pixel matching:
$E_\phi=\frac{1}{W \times H} \sum_{i=1}^W \sum_{j=1}^H \phi_S(i, j)$,
where $\phi_S$ is the alignment matrix depending on the similarity of the prediction and ground truth.

F-measure~($F_{\beta}$) is calculated as $F_{\beta} = \frac{\left(1+\beta^2\right) \times precision \times recall}{\beta^2 \times precision + recall}$, where $\beta^2=0.3$. 

Mean absolute error~(MAE) computes the pixel-wise average distance between the prediction mask~$M_{pred} \in [0,1]^{W \times H} $ and ground truth mask~$M_{gt} \in [0,1]^{W \times H}$: MAE $= \frac{1}{W \times H} \sum_{i=1}^W \sum_{j=1}^H \left| M_{pred} - M_{gt} \right|$.

\subsection{Experiment Settings}
\label{sec:setting}
Our method contains a frozen backbone for feature extraction and a decoder for segmentation prediction. We initialize the weight of the backbone via the pre-trained model, and the weight of the decoder is randomly initialized. Below, we give the details of each method.

\textbf{Full-tuning.} We follow the basic setting above, and then, fine-tune all the parameters of the encoder and decoder.

\textbf{Only Decoder.} We follow the basic setting above, and then, fine-tune the parameters in the decoder only.

\textbf{VPT~\cite{vpt}.} We first initialize the model following the basic setting. Then, we concatenate the prompt tokens in each transformer block of the backbone only. Notice that, the prompt embeddings are implicitly shared across the whole dataset. We follow their  original paper and optimize the parameters in the prompt embeddings and the decoder. 

\textbf{AdaptFormer~\cite{chen2022adaptformer}.} We first initialize the model following the basic setting above. Then, the AdaptMLP is added to each transformer block of the backbone for feature adaptation. We fine-tune the parameters in the decoder and the newly introduced AdaptMLPs. 

\textbf{EVP.} We also initialize the weight following the basic setting. Then, we add the explicit prompting as described in Section~\ref{sec:explicit_visual_prompting_for_foreground_segmentation}. We fine-tune the parameters in the decoder and the newly introduced adaptors.

\textbf{Implementation Details.}
All the experiments are performed on a single NVIDIA Titan V GPU with 12G memory. AdamW~\cite{adam} optimizer is used for all the experiments. The initial learning rate is set to $2e^{-4}$ for salient object detection, defocus blur detection, and camouflaged object detection, and $5e^{-4}$ for others. Cosine decay is applied to the learning rate. The models are trained for 20 epochs for the SBU~\cite{sbu} dataset and camouflaged combined dataset~\cite{fan2020camouflaged,skurowski2018animal}, and 50 epochs for others. Random horizontal flipping is applied during training for data augmentation. The mini-batch is equal to 4. Binary cross-entropy (BCE) loss is used for defocus blur detection and forgery detection, balanced BCE loss is used for shadow detection, and BCE loss and IOU loss are used for salient object detection and camouflaged object detection.
We use two different architectures, plain vision transformer~(ViT) and hierarchical vision transformer~(SegFormer) for experiments. For ViT, we employ ViT-Base as the backbone and construct a decoder with a down-projection layer, four transformer blocks, and a linear layer. And we follow the original design for Segformer with the mit-b4 backbone.   

\newcommand{\tablestyle}[2]{\setlength{\tabcolsep}{#1}\renewcommand{\arraystretch}{#2}\centering
\footnotesize}


\begin{table*}[t]
	\centering
 
    \hspace{-5em}
	\begin{minipage}{1.\linewidth}
        \caption{Comparison with SOTA methods on salient object detection.}
		\centering
        \tablestyle{1pt}{1.}
        \begin{tabular}{l||cccc|cccc|cccc|cccc|cccc}
            \toprule
            \multirow{2}{*}{Method}&  \multicolumn{4}{c|}{DUTS~\cite{wang2017learning}}& \multicolumn{4}{c|}{DUT-OMRON~\cite{yang2013saliency}} & \multicolumn{4}{c|}{HKU-IS~\cite{li2015visual}}  & \multicolumn{4}{c|}{ECSSD~\cite{shi2015hierarchical}} & \multicolumn{4}{c}{PASCAL-S~\cite{li2014secrets}}     \\ \cline{2-21}
            & $S_\alpha \uparrow$ & $E_\phi$ $\uparrow$ & $F_\beta^m$ $\uparrow$ & MAE $\downarrow$  & $S_\alpha \uparrow$ & $E_\phi$ $\uparrow$ & $F_\beta^m$ $\uparrow$ & MAE $\downarrow$ & $S_\alpha \uparrow$ & $E_\phi$ $\uparrow$ & $F_\beta^m$ $\uparrow$ & MAE $\downarrow$  & $S_\alpha \uparrow$ & $E_\phi$ $\uparrow$ & $F_\beta^m$ $\uparrow$ & MAE $\downarrow$  & $S_\alpha \uparrow$ & $E_\phi$ $\uparrow$ & $F_\beta^m$ $\uparrow$ & MAE $\downarrow$\\ \hline
            PoolNet~\cite{liu2019simple} & .878 & .889 & .880 & .040 & .828 & .863 & .808 & .056 & .910 & .949 & .933 & .032 & .922 & .924 & .944 & .039 & .847 & .850 & .869 & .074 \\
            LDF~\cite{wei2020label}  & .892 & .910 & .898 & .034 & .838 & .873 & .820 & .051 & .919 & .954 & .939 & .027 & .924 & .925 & .950 & .034 & .856 & .865 & .874 & .059\\
            VST~\cite{liu2021visual}  & .896 & .892 & .890 & .037 & .850 & .861 & .825 & .058 & .928 & .953 & .942 & .029 & .932 & .918 & .951 & .033 & .865 & .837 & .875 & .061\\
            SelfReformer~\cite{yun2022selfreformer} & .911 & .920 & .916 & .026 & .856 & .886 & .836 & \bf.041 & .930 & .959 & .947 & .024 & .935 & .928 & .957 & .027 & .874 & .872 & \bf.894 & .050 \\
            BBRF~\cite{ma2023boosting} & .908 & .927 & .916 & \bf.025 & .855 & .887 & .843 & .042 & .935 & .965 & \bf.958 & \bf.020 & .939 & .934 & .963 & \bf.022 & .871 & .867 & .891 & \bf.049\\
            \hline
            
            EVPv1~(SegFormer~\cite{xie2021segformer}) & .913 & .947 & \bf.923 & .026 & .862 & .894 & .858 & .046 & .931 & .961 & .952 & .024 & .935 & .957 & .960 & .027 & .878 & \bf.917 & .872 & .054\\
            EVPv2~(ViT-CLS~\cite{dosovitskiy2020image}) & .897 & .922 & .898 & .034 & .856 & .875 & .841 & .048 & .921 & .947 & .939 & .030 & .926 & .945 & .951 & .035 & .870 & .900 & .859 & .060\\
            EVPv2~(ViT-MAE~\cite{cheng2022masked}) & .904 & .925 & .913 & .030 & .854 & .870 & .844 & .046 & .922 & .947 & .939 & .030 & .937 & .954 & .960 & .029 & .877 & .907 & .866 & .054\\
            EVPv2~(ViT-SAM~\cite{kirillov2023segment})  & \bf.915 & .944 & .921 & .026 & \bf.874 & \bf.902 & \bf.865 & \bf.040 & \bf.937 & \bf.966 & .957 & \bf.020 & \bf.944 & \bf.964 & \bf.967 & \bf.022 & .877 & .912 & .868 & .054 \\
            EVPv2~(SegFormer~\cite{xie2021segformer}) & \bf.915 & \bf.948 & \bf.923 & .027 & .862 & .895 & .857 & .047 & .932 & .963 & .953 & .023 & .935 & .957 & .958 & .028 & \bf.879 & \bf.917 & .869 & .053\\
            \bottomrule
        \end{tabular}
    \label{tab:sota_salient}
	\end{minipage}\quad
 
	\begin{minipage}{0.5\linewidth}
        \caption{Comparison with SOTA methods on defocus blur detection.}
		\centering
        \tablestyle{1pt}{1.}
        \begin{tabular}{l||cc|cc}
            \toprule
             \multirow{2}{*}{Method} &  \multicolumn{2}{c|}{DUT~\cite{zhao2018defocus}}   & \multicolumn{2}{c}{CUHK~\cite{shi2014discriminative}}  \\  \cline{2-5}
            & $F_{\beta}\uparrow$ & MAE$\downarrow$   & $F_{\beta}\uparrow$ & MAE$\downarrow$    \\ \hline            DeFusionNet~\cite{tang2019defusionnet}  &  .823    & .118  & .818 & .117 \\
            BTBNet~\cite{zhao2019btbnet}  &  .827    & .138  & .889 & .082 \\
            CENet~\cite{zhao2019cenet}    & .817    & .135  & .906 & .059 \\
            DAD~\cite{zhao2021self}     & .794    & .153  & .884    & .079  \\
            EFENet~\cite{zhao2021defocus} & .854    & .094  & .914   & .053  \\ \hline
            EVPv1~(SegFormer~\cite{xie2021segformer}) & \bf.890 & \bf.068 & .928 & .045 \\
            EVPv2~(ViT-CLS~\cite{dosovitskiy2020image}) & .868 & .083 & .904 & .058  \\ 
            EVPv2~(ViT-MAE~\cite{cheng2022masked}) & .883 & .078 & .923 & .047  \\ 
            EVPv2~(ViT-SAM~\cite{kirillov2023segment}) & .866 & .097 & \bf.932 & .043  \\ 
            EVPv2~(SegFormer~\cite{xie2021segformer}) & .887 & .070 & \bf.932 & \bf.042  \\ \bottomrule
        \end{tabular} 
    \label{tab:sota_defocus}
	\end{minipage}\quad
	\begin{minipage}{0.45\linewidth}
        \caption{Comparison with SOTA methods on forgery detection.}
		\centering
        \tablestyle{1pt}{1.}
        \begin{tabular}{l||cc|cc}
            \toprule
             \multirow{2}{*}{Method} & \multicolumn{2}{c|}{IMD20~\cite{novozamsky2020imd2020}}   & \multicolumn{2}{c}{CAISA~\cite{dong2013casia}}  \\ \cline{2-5}
            & F1$\uparrow$ & AUC$\uparrow$   & F1$\uparrow$ & AUC$\uparrow$      \\ \hline
            ManTra~\cite{wu2019mantra}    & -         & .748    &  -   & .817       \\
            SPAN~\cite{hu2020span}        &    -    & .750         & .382       & .838       \\
            PSCCNet~\cite{liu2022pscc}        &   -  & .806         & .554       & .875       \\
            TransForensics~\cite{hao2021transforensics} & -   & \bf.848    & .627    & .837       \\ 
            ObjectFormer~\cite{wang2022objectformer}    & -   & .821     & .579     & \bf.882     \\\hline
            EVPv1~(SegFormer~\cite{xie2021segformer})   & .443 & .807 &.636 & .862 \\
            EVPv2~(ViT-CLS~\cite{dosovitskiy2020image})  & .371 & .755 & .471 & .784  \\ 
            EVPv2~(ViT-MAE~\cite{cheng2022masked})    & .404 & .772 & .619 & .858  \\ 
            EVPv2~(ViT-SAM~\cite{kirillov2023segment})    & .418 & .792 & .614 & .842  \\ 
            EVPv2~(SegFormer~\cite{xie2021segformer})    & \bf.446 & .811 & \bf.654 & .876  \\  \bottomrule
        \end{tabular}
    \label{tab:sota_forgery}
	\end{minipage}\quad

    \hspace{-8em}
	\begin{minipage}{0.4\linewidth}
        \caption{Comparison with SOTA methods on shadow detection.}
		\centering
        \tablestyle{1pt}{1.}
        \begin{tabular}{l||c|c}
            \toprule
             \multirow{2}{*}{Method}  & ISTD~\cite{wang2018stacked} &  SBU~\cite{sbu}  \\   \cline{2-3}
            & BER$\downarrow$ & BER$\downarrow$    \\ \hline
            BDRAR~\cite{zhu2018bidirectional}  & 2.69   & 3.89   \\
            DSC~\cite{hu2018direction}          & 3.42      & 5.59   \\
            DSD~\cite{zheng2019distraction}                 & 2.17 & 3.45   \\
            MTMT~\cite{mtmt}   & 1.72 & 3.15 \\
            FDRNet~\cite{zhu2021mitigating} & 1.55 & \bf3.04 \\ \hline
            EVPv1~(SegFormer~\cite{xie2021segformer}) & \bf1.35 & 4.31 \\
            EVPv2~(ViT-CLS~\cite{dosovitskiy2020image})  & 1.80 & 5.16 \\ 
            EVPv2~(ViT-MAE~\cite{cheng2022masked})  & 1.65 & 4.13 \\ 
            EVPv2~(ViT-SAM~\cite{kirillov2023segment})  & 1.51 & 4.28 \\ 
            EVPv2~(SegFormer~\cite{xie2021segformer})  & \bf1.35 & 4.15 \\ \bottomrule
        \end{tabular}
    \label{tab:sota_shadow}
	\end{minipage}\quad
    \begin{minipage}{0.6\linewidth}
        \caption{Comparison with SOTA methods on camouflaged object detection.}
		\centering
        \footnotesize
		\tablestyle{1pt}{1.}
        \begin{tabular}{l||cccc|cccc|cccc}
            \toprule
            \multirow{2}{*}{Method}&  \multicolumn{4}{c|}{CHAMELEON~\cite{skurowski2018animal}} & \multicolumn{4}{c|}{CAMO~\cite{le2019anabranch}} & \multicolumn{4}{c}{COD10K~\cite{fan2020camouflaged}}    \\ \cline{2-13}
            & $S_\alpha$$\uparrow$ & $E_\phi$ $\uparrow$ & $F_\beta^w$ $\uparrow$ & MAE $\downarrow$  & $S_\alpha \uparrow$ & $E_\phi$ $\uparrow$ & $F_\beta^w$ $\uparrow$ & MAE $\downarrow$ & $S_\alpha \uparrow$ & $E_\phi$ $\uparrow$ & $F_\beta^w$ $\uparrow$ & MAE $\downarrow$\\ \hline
            SINet~\cite{fan2020camouflaged} & .869 & .891 & .740 & .044 & .751 & .771 & .606 & .100 & .771 & .806 & .551 & .051 \\
            RankNet~\cite{lv2021simultaneously}  & .846 & .913 & .767 & .045 & .712 & .791 & .583 & .104 & .767 & .861 & .611 & .045 \\
            JCOD~\cite{li2021uncertainty}  & .870 & .924 & - & .039 & .792 & .839 & - & .082 & .800 & .872 & - & .041 \\
            PFNet~\cite{mei2021camouflaged} & .882 & \bf.942 & .810 & .033 & .782 & .852 & .695 & .085 & .800 & .868 & .660 & .040  \\
            FBNet~\cite{jiaying2022frequency} & \bf.888 & .939 & \bf.828 & .032 & .783 & .839 & .702 & .081 & .809 & .889 & .684 & .035 \\
            \hline
            EVPv1~(SegFormer~\cite{xie2021segformer}) & .871 & .917 & .795 & .036 & .846 & .895 & .777 & .059 & \bf.843 & .907 & .742 & \bf.029\\
            EVPv2~(ViT-CLS~\cite{dosovitskiy2020image})& .829 & .899 & .732 & .048 & .781 & .837 & .683 & .082 & .768 & .852 & .622 & .047\\
            EVPv2~(ViT-MAE~\cite{he2022masked}) & .885 & .931 & .825 & \bf.030 & \bf.851 & .894 & \bf.797 & \bf.055 & .839 & .908 & \bf.747 & \bf.029\\
            EVPv2~(ViT-SAM~\cite{kirillov2023segment}) & .883 & .924 & .819 & \bf.030 & .833 & .876 & .768 & .066 & .826 & .887 & .722 & .034\\
            EVPv2~(SegFormer~\cite{xie2021segformer}) & .878 & .919 & .809 & .033 & .848 & \bf.899 & .786 & .058 & \bf.843 & \bf.909 & .746 & \bf.029\\
            \bottomrule
        \end{tabular}
    \label{tab:sota_cod}
	\end{minipage}\quad
\end{table*}

%

\newcommand{\ablationtablestyle}[2]{\setlength{\tabcolsep}{#1}\renewcommand{\arraystretch}{#2}\centering\normalsize}

\begin{table*}[t]
\caption{
Comparison with SOTA parameter-efficient tuning approaches. We conduct evaluations on five datasets for five different tasks.
The efficient tuning method which achieves better performance than full-tuning is marked as {\color{orange}{orange}}.
The best performance among all methods is shown as \textbf{blod}.
}
\centering
\ablationtablestyle{1pt}{1.}
{
\begin{tabular}{l||c|cccc|cc|c|cc|cccc}
\toprule
\multirow{3}{*}{Method}& Trainable & \multicolumn{4}{c|}{\textbf{Salient}} & \multicolumn{2}{c|}{\textbf{Defocus Blur}} & \textbf{Shadow} & \multicolumn{2}{c|}{\textbf{Forgery }} & \multicolumn{4}{c}{\textbf{Camouflaged}}\\
      & Param.& \multicolumn{4}{c|}{DUTS~\cite{wang2017learning}}&  \multicolumn{2}{c|}{CUHK~\cite{shi2014discriminative}} & ISTD~\cite{wang2018stacked}& \multicolumn{2}{c|}{CASIA~\cite{dong2013casia}} & \multicolumn{4}{c}{CAMO~\cite{le2019anabranch}}\\ 
      & (M) & $S_\alpha \uparrow$ & $E_\phi$ $\uparrow$  & $F_{\beta}^{m}$ $\uparrow$ & MAE $\downarrow$   &$F_{\beta}\uparrow$ & MAE $\downarrow$      & BER $\downarrow$    & $F_1\uparrow$ & AUC $\uparrow$  & $S_\alpha \uparrow$ & $E_\phi$ $\uparrow$  & $F_\beta^w$ $\uparrow$ & MAE $\downarrow$   \\ \hline
      \multicolumn{15}{c}{\textbf{Tuning with ViT-CLS~\cite{dosovitskiy2020image}}} \\ \hline
Full-tuning & 89.33 & .836 & .869 & .827  & .059  & .899 & .057 & 9.48 & .331 & .692 & .647 & .704 & .473 & .132 \\  
Only Decoder & 3.22 & .788 & .826 & .755  & .078  & .804  & .115 & 12.4 & .314 & .688 & .613 & .662 & .407 & .159 \\
VPT-Deep~\cite{vpt} & 3.68 & \color{orange}{.867} & \color{orange}{.893} & \color{orange}{.864}  & \color{orange}{.045} & .889  & .066 & \color{orange}{2.78} & \color{orange}{.400} & \color{orange}{.751} & \color{orange}{.734} & \color{orange}{.792} & \color{orange}{.605} & \color{orange}{.105} \\
AdaptFormer~\cite{chen2022adaptformer} & 3.69 & \color{orange}{.886} & \color{orange}{.913} & \color{orange}{.884} & \color{orange}{.038}  & .888 & .068  & \color{orange}{2.41} & \color{orange}{.393} & \color{orange}{.747} & \color{orange}{.756} & \color{orange}{.813} & \color{orange}{.650} & \color{orange}{.095}  \\
EVPv1 & 3.72 & \color{orange}{\textbf{.904}} & \color{orange}{\textbf{.924}} & \color{orange}{\textbf{.909}}  & \color{orange}{\textbf{.032}} & \color{orange}{.901}  & .059 & \color{orange}{2.34} & \color{orange}{.398} & \color{orange}{.740} & \color{orange}{.753} & \color{orange}{.812} & \color{orange}{.634} & \color{orange}{.097} \\ 
EVPv2 & 3.71 & \color{orange}{.897} & \color{orange}{.922} & \color{orange}{.898}  & \color{orange}{.034} & \color{orange}{\textbf{.905}}  & \color{orange}{\textbf{.057}} & \color{orange}{\textbf{1.80}} & \color{orange}{\textbf{.471}} & \color{orange}{\textbf{.784}} & \color{orange}{\textbf{.781}} & \color{orange}{\textbf{.837}} & \color{orange}{\textbf{.683}} & \color{orange}{\textbf{.082}} \\ \hline
      \multicolumn{15}{c}{\textbf{Tuning with SegFormer~\cite{xie2021segformer}}} \\ \hline
Full-tuning & 64.00 & .896 & .921 & .901  & .035  & \bf.935 & \bf.039 & 2.42 & .465 & .754 & .837 & .887 & .778 & .060 \\  
Only Decoder & 3.15 & .864 & .884 & .876  & .050  & .891  & .080 & 4.36 & .396 & .722 & .783 & .827 & .671 & .088 \\
VPT-Deep~\cite{vpt} & 3.72 & \color{orange}{.912} & \color{orange}{.932} & \color{orange}{.922}  & \color{orange}{.030} & .925  & .054 & \color{orange}{1.61} & \color{orange}{.621} & \color{orange}{.858} & \color{orange}{.840} & \color{orange}{.890} & \color{orange}{.762} & \color{orange}{.065} \\
AdaptFormer~\cite{chen2022adaptformer} & 3.71 & \color{orange}{.913} & \color{orange}{.937} & \color{orange}{.922} & \color{orange}{.028}  & .922 & .050  & \color{orange}{1.60} & \color{orange}{\textbf{.668}} & \color{orange}{\textbf{.880}} & \color{orange}{.844} & \color{orange}{.892} & \color{orange}{.776} & \color{orange}{.061}  \\
EVPv1 & 3.70 & \color{orange}{.913} & \color{orange}{.947} & \color{orange}{\textbf{.923}}  & \color{orange}{\textbf{.026}} & .928  & .045 & \color{orange}{\textbf{1.35}} & \color{orange}{.636} & \color{orange}{.862} & \color{orange}{.846} & \color{orange}{.895} & \color{orange}{.777} & \color{orange}{.059} \\ 
EVPv2 & 3.68 & \color{orange}{\textbf{.915}}  & \color{orange}{\textbf{.948}}  & \color{orange}{.923} & \color{orange}{.027}   & .932 & .042 & \color{orange}{\textbf{1.35}} & \color{orange}{.654} & \color{orange}{.876} & \color{orange}{\textbf{.848}} & \color{orange}{\textbf{.899}} & \color{orange}{\textbf{.786}} & \color{orange}{\textbf{.058}} \\  \bottomrule
\end{tabular}
}
\label{tab:sota_finetune}
\end{table*}

\subsection{Main Results}
\label{sec:main_results}
In this section, we evaluate the performance of the proposed method with other task-specific state-of-the-art methods. We also report the recent parameter-efficient fine-tuning methods for image classification which can be adapted to our task. 

\begin{figure*}[tp]
    \centering
    \includegraphics[width=0.9\linewidth]{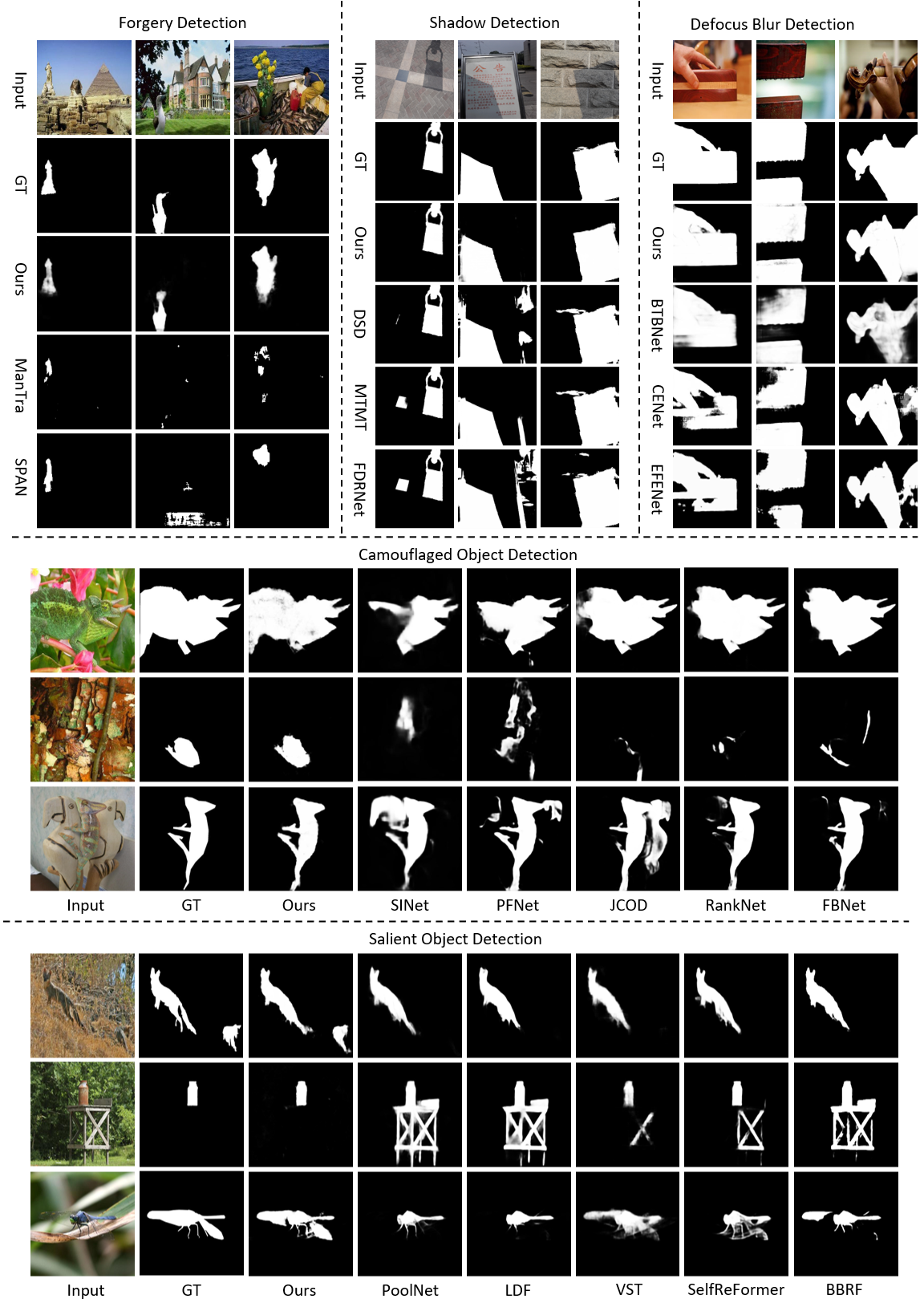}
    \caption{Comparisons with other task-specific methods. We show the results of: 
    (1)~ManTra~\cite{wu2019mantra} and SPAN~\cite{hu2020span} on CAISA~\cite{dong2013casia} dataset for forgery detection~(top-right); (2)~DSD~\cite{zheng2019distraction}, MTMT~\cite{mtmt}, and FDRNet~\cite{zhu2021mitigating} on ISTD~\cite{wang2018stacked} dataset for shadow detection~(top-middle); 
    (3)~BTBNet\cite{zhao2019btbnet}, CENet~\cite{zhao2019cenet}, and EFENet~\cite{zhao2021defocus} on CUHK~\cite{shi2014discriminative} dataset for defocus blur detection~(top-right);
    (4)~SINet~\cite{fan2020camouflaged}, PFNet~\cite{mei2021camouflaged},  JCOD~\cite{li2021uncertainty}, RankNet~\cite{lv2021simultaneously}, and FBNet~\cite{jiaying2022frequency} on CAMO~\cite{le2019anabranch} dataset for camouflaged object detection~(middle);
    and (5)~PoolNet~\cite{liu2019simple}, LDF~\cite{wei2020label}, VST~\cite{liu2021visual}, SelfReformer~\cite{yun2022selfreformer}, and BBRF~\cite{ma2023boosting} on DUTS~\cite{wang2017learning} dataset for salient object detection~(bottom).}
    \label{fig:sota_result}
\end{figure*}

\textbf{Comparison with the task-specific methods.}
We report the comparison of our methods and other task-specific methods in Table~\ref{tab:sota_salient}, Table~\ref{tab:sota_defocus}, Table~\ref{tab:sota_forgery}, Table~\ref{tab:sota_shadow}, and Table~\ref{tab:sota_cod}, which show that EVP performs well when compared with task-specific methods.
Note that some task-specific methods~(ManTra~\cite{wu2019mantra}, SPAN~\cite{hu2020span}, PSCCNet~\cite{liu2022pscc}, and ObjectFormer~\cite{wang2022objectformer} in Table~\ref{tab:sota_forgery}, and MTMT~\cite{mtmt} in Table~\ref{tab:sota_shadow}) use extra training data to get better performance, while we only use the training data from the standard datasets.
Despite utilizing a simple ViT~(ViT-CLS) as the backbone, EVP demonstrates good results, particularly when using two other pre-trained weights: the self-supervised model~(ViT-MAE~\cite{he2022masked}) and the recent foundation model~(ViT-SAM~\cite{kirillov2023segment}). 
We find that both variants achieve better results than the weight training on imagenet labeled data, which indicates that advanced training strategies and huge-scale datasets allow models to learn generalized features for downstream tasks.
The proposed method also benefits by leveraging a stronger backbone~\cite{xie2021segformer}.
These results demonstrate the scalability of the proposed method.
Despite introducing only a small number of tunable parameters with the frozen backbone, we achieve non-trivial performance when compared with other well-designed domain-specific methods. 
We also show some visual comparisons with other methods for each task individually in Figure~\ref{fig:sota_result}. We can see the proposed method predicts more accurate masks compared to other approaches.

\textbf{Comparison with the efficient tuning methods.}
We evaluate our method with full fine-tuning and only tuning the decoder, which are the widely-used strategies for down-streaming task adaption. And similar methods from image classification, \emph{i.e.}, VPT~\cite{vpt} and AdaptFormer~\cite{chen2022adaptformer}. 
We adopt ViT-CLS and SegFormer for experiments and set the number of prompt tokens to 50/~50 for VPT, the middle dimension of AdaptMLP to 24/~24, the scale factor~($r$) of EVPv1 to 6/~4, the scale factor~($r$) of EVPv2 to 32/~16 in ViT-CLS/~SegFormer for fair comparisons in terms of the tunable parameters.
It can be seen from Table~\ref{tab:sota_finetune} that when only tuning the decoder, the performance drops largely. 
Further comparisons with similar methods indicate that introducing extra learnable tokens~\cite{vpt} or MLPs in transformer block~\cite{chen2022adaptformer} benefits the performance, with our proposed method achieving the best performance among them. 
Overall, our method outperforms full-tuning in all five datasets using ViT-CLS and four out of five datasets using SegFormer, and only worse than the recent advanced fine-tuning method in one dataset.
These results highlight the effectiveness of the proposed method in adapting to downstream tasks, which can be attributed to explicit visual prompting that enables knowledge transfer from the large-scale pre-training dataset to the downstream tasks effectively.
Moreover, we observe that the proposed method achieves good performance for both ViT-CLS and SegFormer, which indicates the generality and effectiveness of different architectures. 
The consistent improvements in different tasks show that we offer an effective solution for a wide range of downstream tasks, hence is a universal framework.

\begin{table*}[!t]
\caption{Ablation on the architecture designs described in Figure~\ref{fig:adaptor} and Figure~\ref{fig:freq_enhanced_adaptor}. We conduct evaluations on five datasets for five different tasks. 
}
\centering
\ablationtablestyle{1pt}{1.}
{
\begin{tabular}{l||c|cccc|cc|c|cc|cccc}
\toprule
\multirow{3}{*}{Method}& Trainable& \multicolumn{4}{c|}{\textbf{Salient}} & \multicolumn{2}{c|}{\textbf{Defocus Blur}} & \textbf{Shadow} & \multicolumn{2}{c|}{\textbf{Forgery }} & \multicolumn{4}{c}{\textbf{Camouflaged}}\\
      & Param. & \multicolumn{4}{c|}{DUTS~\cite{wang2017learning}}&  \multicolumn{2}{c|}{CUHK~\cite{shi2014discriminative}} & ISTD~\cite{wang2018stacked}& \multicolumn{2}{c|}{CASIA~\cite{dong2013casia}} & \multicolumn{4}{c}{CAMO~\cite{le2019anabranch}}\\ 
      & (M)  & $S_\alpha \uparrow$ & $E_\phi$ $\uparrow$  & $F_{\beta}^{m}$ $\uparrow$ & MAE $\downarrow$ &$F_{\beta}\uparrow$ & MAE $\downarrow$      & BER $\downarrow$    & $F_1\uparrow$ & AUC $\uparrow$  & $S_\alpha \uparrow$ & $E_\phi$ $\uparrow$  & $F_\beta^w$ $\uparrow$ & MAE $\downarrow$   \\ \hline
Full-tuning & 64.00 & .896 & .921 & .901  & .035  & .935 & .039 & 2.42 & .465 & .754 & .837 & .887 & .778 & .060 \\  
Only Decoder & 3.15 & .864 & .884 & .876  & .050  & .891  & .080 & 4.36 & .396 & .722 & .783 & .827 & .671 & .088 \\ \hline 
      \multicolumn{15}{c}{\textbf{Ablation study of EVP with Adaptor~(EVPv1)}} \\ \hline 
w/o $F_{pe}$ & 3.61 & .912 & .940 & .922 & .027 & .924 & .049 & 1.68  & .540 & .833  & .840 & .887 & .759 & .065 \\ 
w/o $F_{hfc}$ & 3.58 & .911 & .943 & .922 & .027 & .926  & .046  & 1.61 & .619   & .846 & .844 & .893 & .773 & .063 \\ 
w/ Shared $\mathtt{MLP^i_{tune}}$ & 3.49 & .911 & .940 & .921 & .028 & .928   & .048    & 1.77 & .619   & .860 & .837 & .889 & .763 & .064 \\
w/ Unshared $\mathtt{MLP_{up}}$ & 4.54 & \bf.914 & \bf.948 & \bf.923 & \bf.026 & .927   & \bf.045   & \bf1.33 & \bf.647   & \bf.875  & .844 & .893 & .774 & .060 \\ 
Full & 3.70 & .913 & .947 & \bf.923 & \bf.026 & \bf.928  & \bf.045  & 1.35 & .636   & .862 & \bf.846 & \bf.895 & \bf.777 & \bf.059 \\ \hline

      \multicolumn{15}{c}{\textbf{Ablation study of EVP with Frequency Enhanced Adaptor~(EVPv2)}} \\ \hline
w/o $F_{pe}$ & 3.65 & .912 & .942 & .921 & .028 & .925   & .048    & 1.75 & .646   & .875 & .848 & .897 & .782 & .059 \\
w/o $F_{hfc}$ & 3.63 & .914 & .945 & .922 & .028 & .928   & .044    & 1.47 & .636   & .864 & .846 & .898 & .781 & .060 \\
w/o FEAdaptor & 4.12 & .914 & .945 & .922 & \bf.027 & \bf.932   & \bf.040    & 1.40 & \bf.659   & .874 & .845 & .897 & .782 & .059 \\ 
Full & 3.68 & \bf.915 & \bf.948 & \bf.923 & \bf.027 & \bf.932   & .042  & \bf1.35 & .654   & \bf.876 & \bf.848 & \bf.899 & \bf.786 & \bf.058 \\ \bottomrule
\end{tabular}}
\label{tab:arch}
\end{table*}

\begin{table*}[!t]
\caption{Ablation on the tuning stages in SegFormer~\cite{xie2021segformer}. We conduct evaluations on five datasets for five different tasks. 
}
\centering
\ablationtablestyle{1pt}{1.}
{
\begin{tabular}{l||c|cccc|cc|c|cc|cccc}
\toprule
Tuning & Trainable & \multicolumn{4}{c|}{\textbf{Salient}}& \multicolumn{2}{c|}{\textbf{Defocus Blur}} & \textbf{Shadow} & \multicolumn{2}{c|}{\textbf{Forgery}} & \multicolumn{4}{c}{\textbf{Camouflaged}}\\
      Stage & Param.& \multicolumn{4}{c|}{DUTS~\cite{wang2017learning}}&  \multicolumn{2}{c|}{CUHK~\cite{shi2014discriminative}} & ISTD~\cite{wang2018stacked}& \multicolumn{2}{c|}{CASIA~\cite{dong2013casia}} & \multicolumn{4}{c}{CAMO~\cite{le2019anabranch}}\\ 
      & (M) & $S_\alpha \uparrow$ & $E_\phi$ $\uparrow$  & $F_{\beta}^{m}$ $\uparrow$ & MAE $\downarrow$    &$F_{\beta}\uparrow$ & MAE $\downarrow$      & BER $\downarrow$    & $F_1\uparrow$ & AUC $\uparrow$  & $S_\alpha \uparrow$ & $E_\phi$ $\uparrow$  & $F_\beta^w$ $\uparrow$ & MAE $\downarrow$   \\ \hline
Stage$_{1}$ & 3.15 & .876 & .896 & .888  & .045 & .895  & .072 & 3.99 & .405 & .724 & .791 & .834 & .674 & .089  \\ 
Stage$_{2}$ & 3.17 & .882 & .903 & .895  & .040 & .912  & .055 & 2.15 & .465 & .772 & .803 & .850 & .705 & .081  \\
Stage$_{3}$ & 3.55 & .911 & .930 & .917  & .031 & .926  & .045 & 1.44 & .625 & .862 & .842 & .892 & .774 & .061  \\
Stage$_{4}$ & 3.26 & .890 & .909 & .899  & .037 & .898  & .070 & 3.21 & .475 & .787 & .797 & .849 & .705 & .080  \\
Stage$_{1234}$ & 3.68 & \bf.915  & \bf.948  & \bf.923 & \bf.027   & \bf.932 & \bf.042 & \bf1.35 & \bf.654 & \bf.876 & \bf.848 & \bf.899 & \bf.786 & \bf.058  \\ 
\bottomrule
\end{tabular}}
\label{tab:tuning_stage}
\end{table*}

\begin{table*}[!t]
\caption{Ablation on the parameter scale factor $r$. We conduct evaluations on five datasets for five different tasks. 
}
\centering
\ablationtablestyle{1pt}{1.}
{
\begin{tabular}{l||c|cccc|cc|c|cc|cccc}
\toprule
\multirow{3}{*}{$r$}& Trainable & \multicolumn{4}{c|}{\textbf{Salient}} & \multicolumn{2}{c|}{\textbf{Defocus Blur}} & \textbf{Shadow} & \multicolumn{2}{c|}{\textbf{Forgery }} & \multicolumn{4}{c}{\textbf{Camouflaged}}\\
      & Param.& \multicolumn{4}{c|}{DUTS~\cite{wang2017learning}}&  \multicolumn{2}{c|}{CUHK~\cite{shi2014discriminative}} & ISTD~\cite{wang2018stacked}& \multicolumn{2}{c|}{CASIA~\cite{dong2013casia}} & \multicolumn{4}{c}{CAMO~\cite{le2019anabranch}}\\ 
      & (M) & $S_\alpha \uparrow$ & $E_\phi$ $\uparrow$  & $F_{\beta}^{m}$ $\uparrow$ & MAE $\downarrow$ &$F_{\beta}\uparrow$ & MAE $\downarrow$      & BER $\downarrow$    & $F_1\uparrow$ & AUC $\uparrow$  & $S_\alpha \uparrow$ & $E_\phi$ $\uparrow$  & $F_\beta^w$ $\uparrow$ & MAE $\downarrow$   \\ \hline
64 & 3.30 & .913 & .942 & .920  & .030 & .928  & .046 & 1.69 & .637 & .864 & .840 & .888 & .771 & .061 \\ 
32 & 3.42 & .915 & .944 & .922  & .029 & .929  & .043 & 1.47 & .644 & .866 & .847 & .898 & .780 & .061 \\ 
16  & 3.68 & .915  & \bf.948  & \bf.923 & \bf.027   & .932 & .042 & \bf1.35 & \bf.654 & \bf.876 & \bf.848 & \bf.899 & \bf.786 & \bf.058 \\
8  & 4.23  & .913 & .942 & .921  & .029 & .933  & .039 & 1.38 & .640 & .864 & .847 & .897 & .785 & .059 \\
4 & 5.50  & \bf.916 & .945 & .922  & .028 & \bf.935  & \bf.038 & 1.53 & .652 & .874 & .845 & .893 & .785 & .059 \\ \bottomrule
\end{tabular}}
\label{tab:model_size}
\end{table*}

\newcommand\wwfreq{0.095\textwidth}
\newcommand\hhfreq{0.095\textwidth}

\begin{figure*}[t]
\centering
\subfigure{\centering\includegraphics[width=\wwfreq, height=\hhfreq]{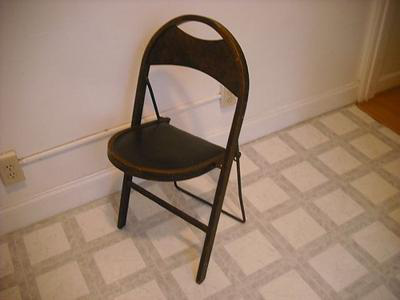}}
\subfigure{\centering\includegraphics[width=\wwfreq, height=\hhfreq]{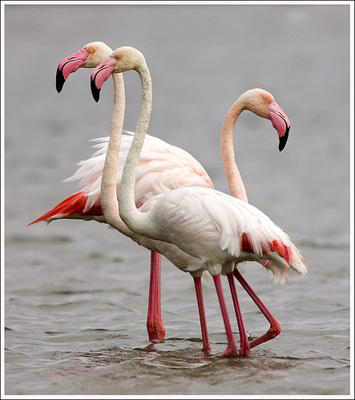}}
\subfigure{\centering\includegraphics[width=\wwfreq, height=\hhfreq]{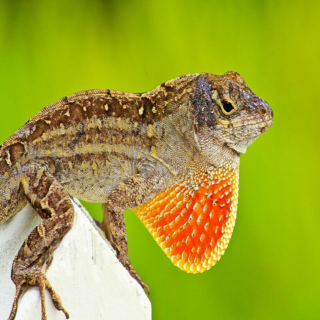}}
\subfigure{\centering\includegraphics[width=\wwfreq, height=\hhfreq]{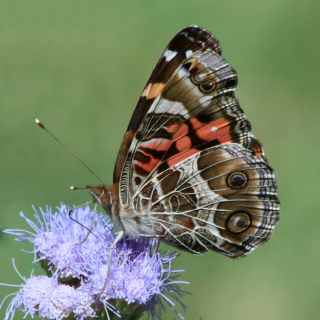}}
\subfigure{\centering\includegraphics[width=\wwfreq, height=\hhfreq]{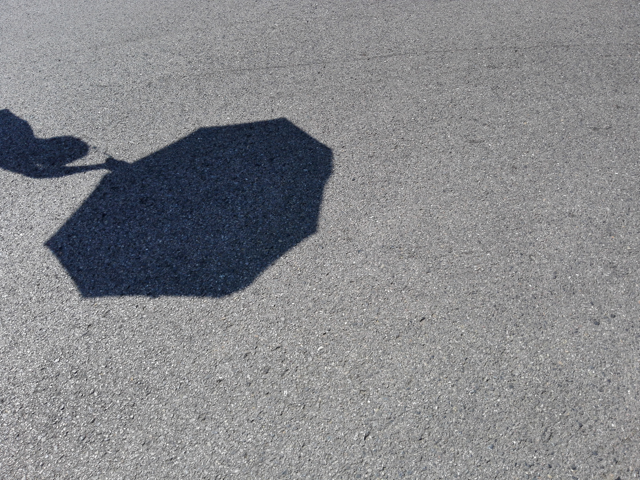}}
\subfigure{\centering\includegraphics[width=\wwfreq, height=\hhfreq]{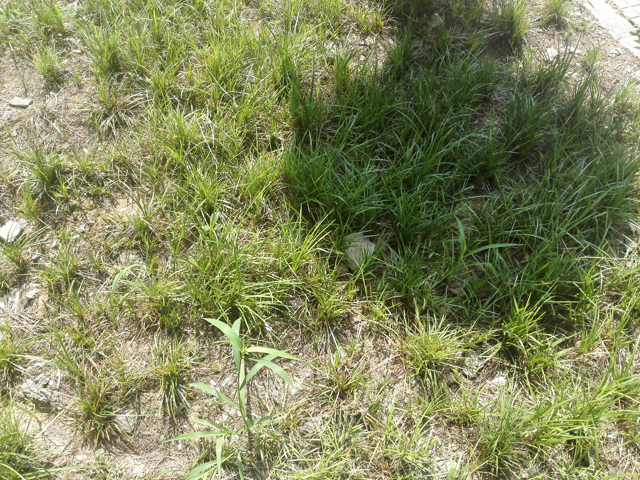}}
\subfigure{\centering\includegraphics[width=\wwfreq, height=\hhfreq]{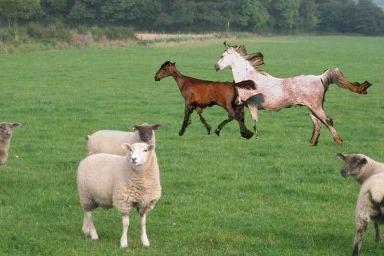}}
\subfigure{\centering\includegraphics[width=\wwfreq, height=\hhfreq]{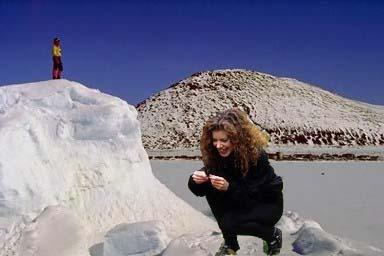}}
\subfigure{\centering\includegraphics[width=\wwfreq, height=\hhfreq]{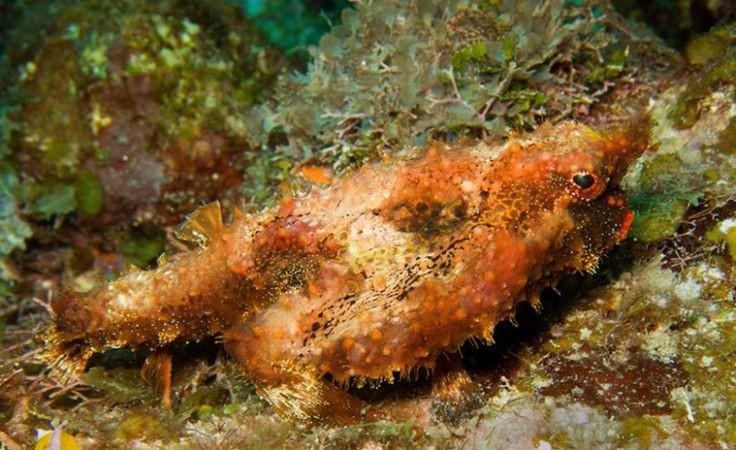}}
\subfigure{\centering\includegraphics[width=\wwfreq, height=\hhfreq]{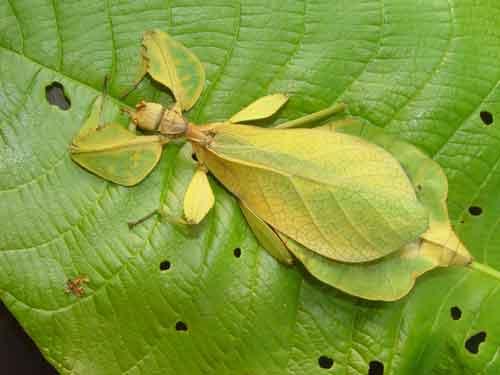}}

\par\bigskip\vspace*{-2em}
\subfigure{\centering\includegraphics[width=\wwfreq, height=\hhfreq]{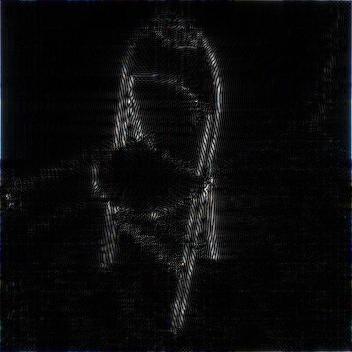}}
\subfigure{\centering\includegraphics[width=\wwfreq, height=\hhfreq]{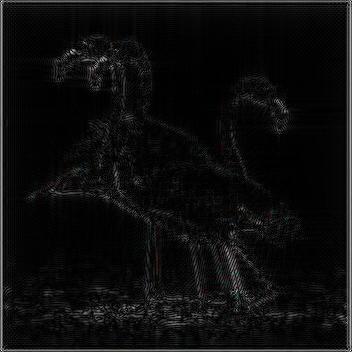}}
\subfigure{\centering\includegraphics[width=\wwfreq, height=\hhfreq]{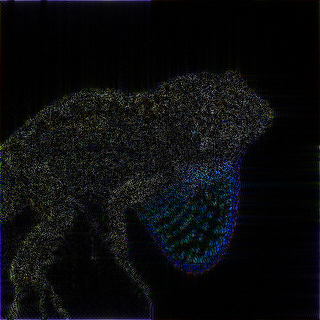}}
\subfigure{\centering\includegraphics[width=\wwfreq, height=\hhfreq]{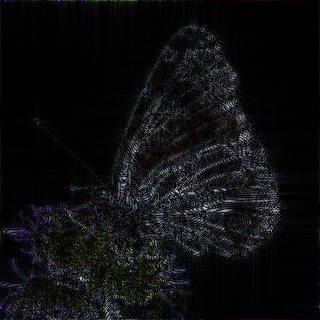}}
\subfigure{\centering\includegraphics[width=\wwfreq, height=\hhfreq]{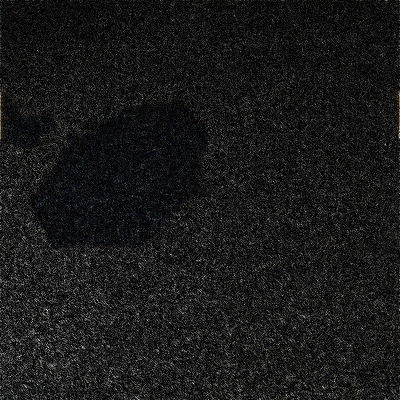}}
\subfigure{\centering\includegraphics[width=\wwfreq, height=\hhfreq]{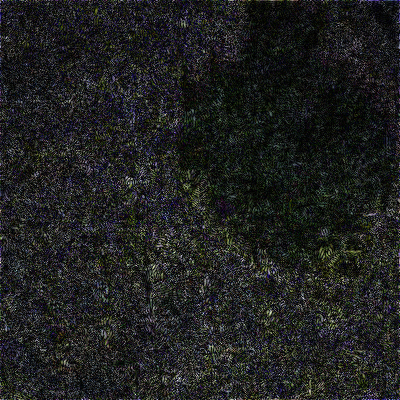}}
\subfigure{\centering\includegraphics[width=\wwfreq, height=\hhfreq]{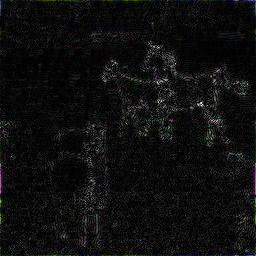}}
\subfigure{\centering\includegraphics[width=\wwfreq, height=\hhfreq]{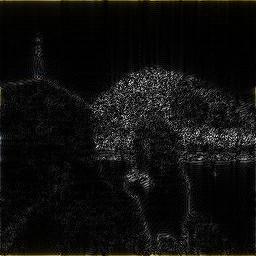}}
\subfigure{\centering\includegraphics[width=\wwfreq, height=\hhfreq]{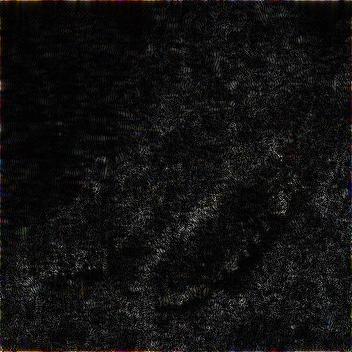}}
\subfigure{\centering\includegraphics[width=\wwfreq, height=\hhfreq]{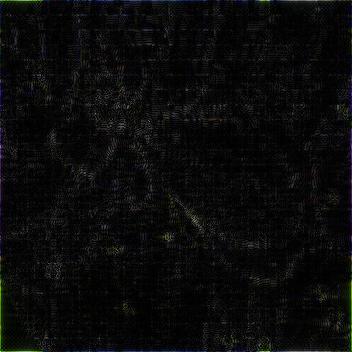}}

\par\bigskip\vspace*{-2em}
\subfigure{\centering\includegraphics[width=\wwfreq, height=\hhfreq]{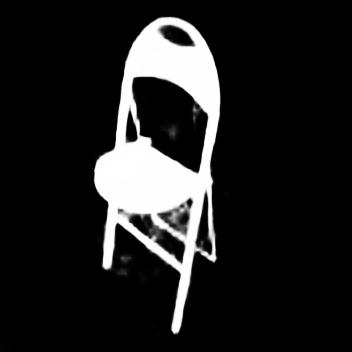}}
\subfigure{\centering\includegraphics[width=\wwfreq, height=\hhfreq]{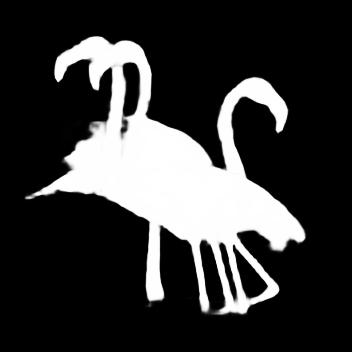}}
\subfigure{\centering\includegraphics[width=\wwfreq, height=\hhfreq]{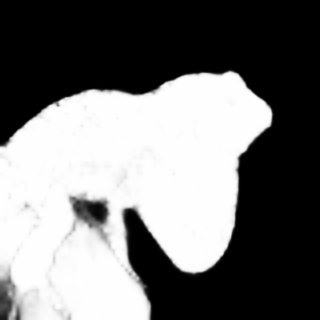}}
\subfigure{\centering\includegraphics[width=\wwfreq, height=\hhfreq]{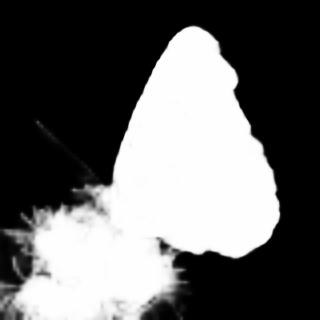}}
\subfigure{\centering\includegraphics[width=\wwfreq, height=\hhfreq]{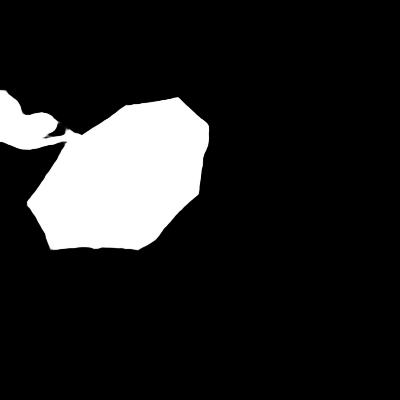}}
\subfigure{\centering\includegraphics[width=\wwfreq, height=\hhfreq]{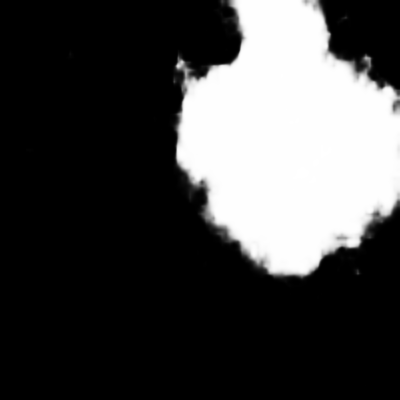}}
\subfigure{\centering\includegraphics[width=\wwfreq, height=\hhfreq]{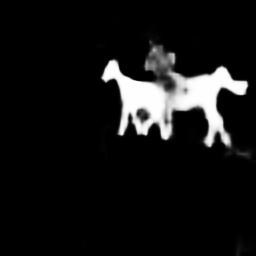}}
\subfigure{\centering\includegraphics[width=\wwfreq, height=\hhfreq]{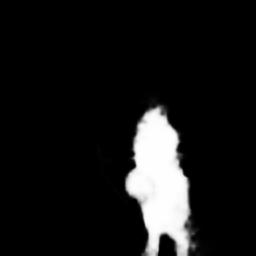}}
\subfigure{\centering\includegraphics[width=\wwfreq, height=\hhfreq]{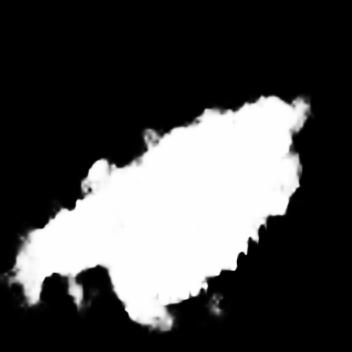}}
\subfigure{\centering\includegraphics[width=\wwfreq, height=\hhfreq]{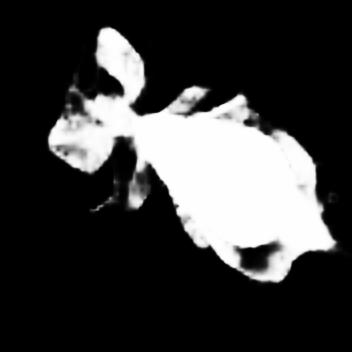}}

\par\bigskip\vspace*{-2em}
\subfigure{\centering\includegraphics[width=\wwfreq, height=\hhfreq]{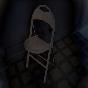}}
\subfigure{\centering\includegraphics[width=\wwfreq, height=\hhfreq]{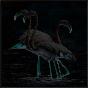}}
\subfigure{\centering\includegraphics[width=\wwfreq, height=\hhfreq]{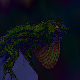}}
\subfigure{\centering\includegraphics[width=\wwfreq, height=\hhfreq]{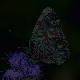}}
\subfigure{\centering\includegraphics[width=\wwfreq, height=\hhfreq]{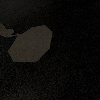}}
\subfigure{\centering\includegraphics[width=\wwfreq, height=\hhfreq]{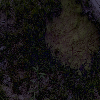}}
\subfigure{\centering\includegraphics[width=\wwfreq, height=\hhfreq]{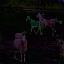}}
\subfigure{\centering\includegraphics[width=\wwfreq, height=\hhfreq]{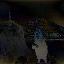}}
\subfigure{\centering\includegraphics[width=\wwfreq, height=\hhfreq]{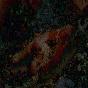}}
\subfigure{\centering\includegraphics[width=\wwfreq, height=\hhfreq]{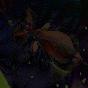}}

\par\bigskip\vspace*{-2em}
\subfigure{\centering\includegraphics[width=\wwfreq, height=\hhfreq]{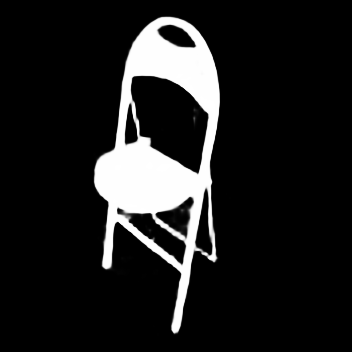}}
\subfigure{\centering\includegraphics[width=\wwfreq, height=\hhfreq]{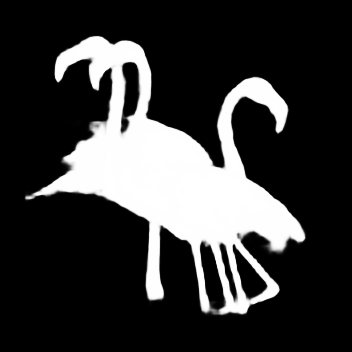}}
\subfigure{\centering\includegraphics[width=\wwfreq, height=\hhfreq]{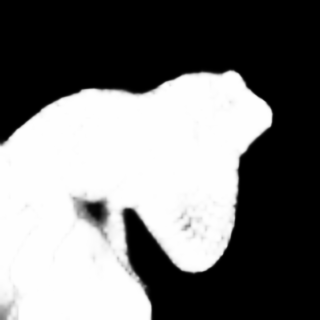}}
\subfigure{\centering\includegraphics[width=\wwfreq, height=\hhfreq]{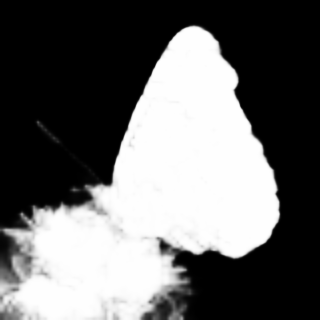}}
\subfigure{\centering\includegraphics[width=\wwfreq, height=\hhfreq]{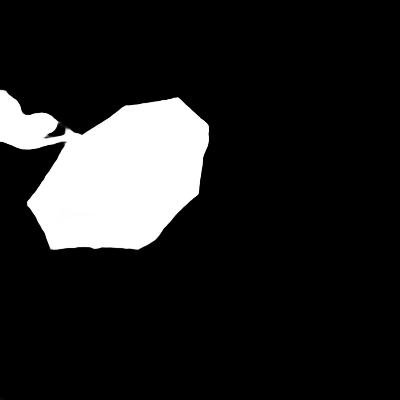}}
\subfigure{\centering\includegraphics[width=\wwfreq, height=\hhfreq]{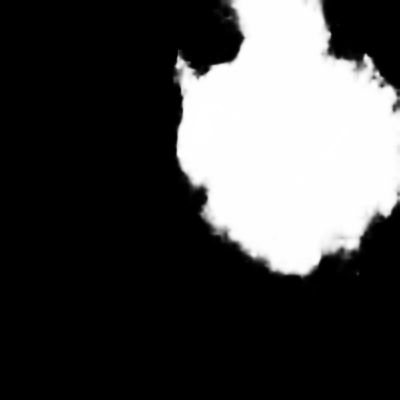}}
\subfigure{\centering\includegraphics[width=\wwfreq, height=\hhfreq]{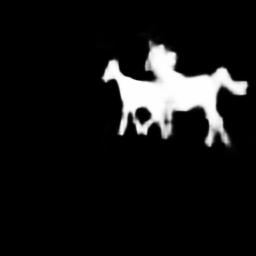}}
\subfigure{\centering\includegraphics[width=\wwfreq, height=\hhfreq]{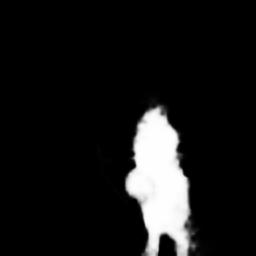}}
\subfigure{\centering\includegraphics[width=\wwfreq, height=\hhfreq]{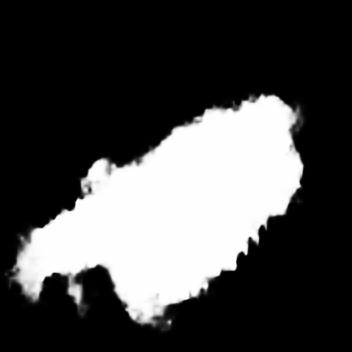}}
\subfigure{\centering\includegraphics[width=\wwfreq, height=\hhfreq]{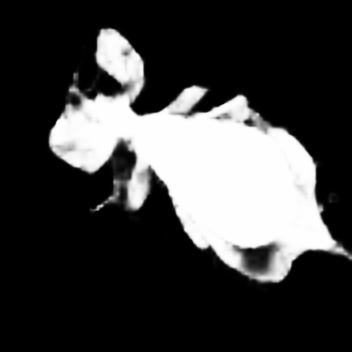}}

\caption{Visualization of prompt and prediction for five tasks.
From top to bottom are input, EVPv1 prompt in the frequency domain, EVPv1 prediction, EVPv2 prompt in the frequency domain, and EVPv2 prediction.
}
\label{fig:freq_vis}
\end{figure*}

\subsection{Ablation Study}
\label{sec:ablation_study}
We conduct the ablation to show the effectiveness of each component.
The following experiments are conducted with SegFormer-B4~\cite{xie2021segformer}.

\textbf{Architecture Design.}
We verify the effectiveness of the proposed framework through two different perspectives.
The scale factor $r$ for EVPv1 and EVPv2 is set to 4 and 16, respectively.
As shown in Table~\ref{tab:arch}, firstly, removing image embeddings or frequency features leads to a decline in performance, thereby indicating that they are both effective as explicit visual prompts for foreground segmentations. 
Secondly, we investigate the design of different adaptors. 
Sharing $\mathtt{MLP^i_{tune}}$ in different Adaptors only saves a small number of parameters~(0.55M \textit{v.s.} 0.34M) but leads to a significant performance drop, and It cannot obtain consistent performance improvement when using different $\mathtt{MLP_{up}}$ in different Adaptors, moreover introducing a large number of parameters~(0.55M \textit{v.s.} 1.39M).
Similarly, using two MLP blocks consisting of 
Low-Dimensional Mapping$\rightarrow$GELU$\rightarrow$High-Dimensional Mapping
for image embeddings and frequency features respectively instead of the proposed Frequency Enhanced Adaptor consume more parameters~(0.97M \textit{v.s.} 0.53M), and obtain no improvement against the proposed design.
Despite the motivation of EVPv1 and EVPv2 are both to learn explicit visual prompts from image embeddings and frequency features, the implementations are totally different.
EVPv1 manually extracts high-frequency components with a fixed mask applied in the spectrum of the input image, while EVPv2 automatically produces frequency features by adaptive masking the spectrum of image embeddings via Fourier MLP, thereby EVPv2 is a stronger method.
We also show the prompts extract in the frequency domain for comparison in Figure~\ref{fig:freq_vis}, which are the high-frequency components by applying the fixed mask~(EVPv1) and adaptive mask~(EVPv2) to the spectrum of the input image, respectively. It can be seen that EVPv1 tends to generate edge prompts, while EVPv2 provides denser prompts and hence can lead to the targets easier.

\textbf{Tuning Stage.}
We conduct an analysis of the contributions of each stage of the hierarchical backbone SegFormer-B4. 
SegFormer-B4 consists of four stages, each containing a varying number of transformer blocks with different dimensionalities. 
The number of transformer blocks of each stage in SegFormer-B4 is 3, 8, 27, and 3, respectively, with corresponding dimensions of 64, 128, 320, and 512. 
The experiments are based on EVP with Frequency Enhanced Adaptor~($r=16$).
We show the results of our tuning method by applying it in every single stage.
We mark the Stage$_{x}$ where the tunable prompting is added in Stage $x$. 
Table~\ref{tab:tuning_stage} shows we get the best performance when tuning all stages, and the second-best is tuning Stage$_{3}$.
It demonstrates that better performance can be obtained via the tunable number of transformer blocks increasing. 
In addition, similar performances are obtained when tuning the same number of blocks in Stage$_{1}$ and Stage$_{4}$, which indicates that the proposed prompting method has no preference for the top or bottom part of the model.

\textbf{Scale Factor $r$~(Section \ref{sec:explicit_visual_prompting_for_foreground_segmentation}).}
We introduce the parameter $r$ to control the number of tunable parameters in our approach in Section~\ref{sec:explicit_visual_prompting_for_foreground_segmentation}. 
A larger value of $r$ implies fewer parameters for tuning. 
We conduct experiments using EVP with Frequency Enhanced Adaptor.
The results in Table~\ref{tab:model_size} suggest that as the value of $r$ decreases from 64 to 16, the performance improves considerably across various tasks. Nevertheless, as we further decrease the value of $r$ to 8 or 4, the model does not exhibit consistent improvement, even if the model becomes larger. These results indicate that employing $r=16$ would be a reasonable choice to strike a balance between the model performance and its size and show the critical role of $r$ in controlling model complexity and performance.

\section{Conclusion}
\label{sec:conclusion}

In this paper, we present explicit visual prompting as a unified foreground segmentation framework. 
This approach re-modulates the transformer features by tuning the frozen features from patch embedding and the learned high-frequency features, thereby adapting the model pre-trained on large-scale datasets to different foreground segmentation tasks.
Equipped with our method, we find that different frozen vision transformer backbones with limited tunable parameters can achieve superior performance than full-tuning, also state-of-the-art performance compared with the other task-specific methods and parameter-efficient fine-tuning methods. 
We hope this work can promote further exploration of general frameworks for computer vision.

\ifCLASSOPTIONcaptionsoff
  \newpage
\fi

\bibliographystyle{IEEEtran.bst}
\bibliography{11_references}

\end{document}


\title{\paperTitle \\ Supplemental Material}
\author{\authorBlock}
\maketitle

\appendix
\label{sec:appendix}
\renewcommand\thetable{\Alph{section}\arabic{table}}   
\renewcommand\thefigure{\Alph{section}\arabic{figure}}  



 
\section{Implementation Details}
\label{sec:implementation_details}
We give more implementation details in the main paper of the "comparison with the task-specific methods" and "comparison with the efficient tuning methods". 

\paragraph{Basic Setting.} Our method contains a backbone for feature extraction and a decoder for segmentation prediction. We initialize the weight of the backbone via ImageNet classification pre-training, and the weight of the decoder is randomly initialized. Below, we give the details of each variant.

\paragraph{Full-tuning.} We follow the basic setting above, and then, fine-tune all the parameters of the encoder and decoder.

\paragraph{Only Decoder.} We follow the basic setting above, and then, fine-tune the parameters in the decoder only.

\paragraph{VPT~\cite{vpt}.} We first initialize the model following the basic setting. Then, we concatenate the prompt embeddings in each transformer block of the backbone only. Notice that, their prompt embeddings are implicitly shared across the whole dataset. We follow their  original paper and optimize the parameters in the prompt embeddings and the decoder. 

\paragraph{AdaptFormer~\cite{chen2022adaptformer}.} We first initialize the model following the basic setting above. Then, the AdaptMLP is added to each transformer block of the backbone for feature adaptation. We fine-tune the parameters in the decoder and the newly introduced AdaptMLP. 

\paragraph{EVP~(Ours).} We also initialize the weight following the basic setting. Then, we add the explicit prompting as described in the main paper of Figure~\ref{fig:arch}. 

\paragraph{Metric.}
AUC calculates the area of the ROC curve. ROC curve is a function of true positive rate~($\frac{tp}{tp + fn}$) in terms of false positive rate~($\frac{fp}{fp + tn}$), where $tp$, $tn$, $fp$, $fn$ represent the number of pixels which are classified as true positive, true negative, false positive, and false negative, respectively. $F_{1}$ score is defined as $F_{1} = \frac{2 \times precision \times recall}{precision + recall}$, where $precision = \frac{tp}{tp + fp}$ and $recall = \frac{tp}{tp + fn}$. 
The balance error rate~(BER) $ = \left(1-\frac{1}{2}\left(\frac{tp}{tp+fn}+\frac{tn}{tn+fp}\right)\right) \times 100$.
F-measure is calculated as $F_{\beta} = \frac{\left(1+\beta^2\right) \times precision \times recall}{\beta^2 \times precision + recall}$, where $\beta^2=0.3$. MAE computes pixel-wise average distance.
Weighted F-measure ($F_\beta^w$) weighting the quantities TP, TN, FP, and FN according to the errors at their location and their neighborhood information:
$F_\beta^w = \frac{\left(1+\beta^2\right) \times precision^w \times recall^w}{\beta^2 \times precision^w + recall^w}$.
E-measure~($E_\phi$) jointly considers image statistics and local pixel matching:
$E_\phi=\frac{1}{W \times H} \sum_{i=1}^W \sum_{j=1}^H \phi_S(i, j)$,
where $\phi_S$ is the alignment matrix depending on the similarity of the prediction and ground truth.

\paragraph{Training Data.}
Note that most forgery detection methods~(ManTraNet~\cite{wu2019mantra}, SPAN~\cite{hu2020span}, PSCCNet~\cite{liu2022pscc}, and ObjectFormer~\cite{wang2022objectformer} in Table~\ref{tab:sota_forgery}) and one shadow detection method~(MTMT~\cite{mtmt} in Table~\ref{tab:sota_shadow}) use extra training data to get better performance. We only use the training data from the standard datasets and obtain SOTA performance.

\section{More Results}
\label{sec:more_results}
We provide more experimental results in addition to the main paper.

\subsection{High-Frequency Prompting}
Our method gets the knowledge from the explicit content of the image itself, hence we also discuss other similar explicit clues of images as the prompts. Specifically, we choose the common-used Gaussian filter, the noise-filter~\cite{fridrich2012rich}, the all-zero image, and the original image as experiments. From Table~\ref{tab:filter}, we find the Gaussian filter shows a better performance in defocus blur since 
it is also a kind of blur. Also, the noise filter~\cite{fridrich2012rich} from forgery detection also boosts the performance. Interestingly, we find that simply replacing the original image with an all-zero image also boosts the performance, since it can also be considered as a kind of implicitly learned embeddings across the full dataset as in VPT~\cite{vpt}. Differently, the high-frequency components of the image achieve consistent performance improvement to other methods on these several benchmarks.

\subsection{HFC \textit{v.s.} LFC} We conduct the ablation study on choosing of high-frequency features or the low-frequency features in Table~\ref{tab:hfc}. From the table, using the low-frequency components as the prompting just show some trivial improvement on these datasets. Differently, the high-frequency components are more general solutions and show a much better performance in shadow detection, forgery detection, and camouflaged detection. Similar to the Gaussian filter as we discussed above, the LFC is also a kind of blur, which makes the advantage of LFC in the defocus blur detection.

\subsection{Mask Ratio $\tau$} We further evaluate the hyper-parameter mask ratio~$\tau$ introduced in Section~\ref{Preliminary}. From Table~\ref{tab:hfc}, when we mask out 25\% of the central pixels in the spectrum, it achieves consistently better performance in all the tasks. We also find that the performance may drop when the increasing of mask ratio~(all 0 images), especially in shadow detection, forgery detection, and camouflaged object detection.

\begin{table*}[t]
\centering
\begin{tabular}{l|cc|c|cc|cccc}
\toprule
\multirow{3}{*}{Method}& \multicolumn{2}{c|}{\textbf{Defocus Blur}} & \textbf{Shadow} & \multicolumn{2}{c|}{\textbf{Forgery}}  & \multicolumn{4}{c}{\textbf{Camouflaged}}\\
&  \multicolumn{2}{c|}{CUHK~\cite{shi2014discriminative}} & ISTD~\cite{wang2018stacked} & \multicolumn{2}{c|}{CASIA~\cite{dong2013casia}}& \multicolumn{4}{c}{CAMO~\cite{le2019anabranch}} \\ 
&$F_{\beta}\uparrow$ & MAE $\downarrow$      & BER $\downarrow$    & $F_1\uparrow$ & AUC $\uparrow$  & $S_\alpha \uparrow$ & $E_\phi$ $\uparrow$  \\ \hline
Gaussian Blur & \bf.928  & .046   & 1.52  & .631  & .860 & .842 & .893   \\ 
Noise Filter  & .923 & .046   & 1.47 & \bf.637  & \bf.866 & .843 & .894   \\
All 0 Image   & .922 & .047  & 1.45 & .630 & .862  & .844 & .895 \\ 
Original  & .922 & .047  & 1.59 & .630 & .860  & .840 & .891  \\ 
\rowcolor{lightgray!30} HFC  &\bf .928 & \bf.045  & \bf 1.35 &  .636 &  .862  & \bf.846 & \bf.895  \\
\bottomrule
\end{tabular}
\caption{Ablation on the explicit visual prompting with other relatives. We compare with the widely-used image filter as prompting to verify the effectiveness of the proposed HFC.
}
\label{tab:filter}
\end{table*}

\begin{table*}[t]
\centering
\begin{tabular}{lc|cc|c|cc|cccc}
\toprule
\multirow{3}{*}{Method}& Mask & \multicolumn{2}{c|}{\textbf{Defocus Blur}} & \textbf{Shadow} & \multicolumn{2}{c|}{\textbf{Forgery}}  & \multicolumn{4}{c}{\textbf{Camouflaged}}\\
      & Ratio &  \multicolumn{2}{c|}{CUHK~\cite{shi2014discriminative}} & ISTD~\cite{wang2018stacked}& \multicolumn{2}{c|}{CASIA~\cite{dong2013casia}}& \multicolumn{4}{c}{CAMO~\cite{le2019anabranch}} \\ 
      & $\tau$ (\%) &$F_{\beta}\uparrow$ & MAE $\downarrow$      & BER $\downarrow$    & $F_1\uparrow$ & AUC $\uparrow$  & $S_\alpha \uparrow$ & $E_\phi$ $\uparrow$  \\ \hline
\multicolumn{11}{c}{\textbf{Low Frequency Components (LFC) with FFT}} \\
LFC* & 0 & .922 & .047  & 1.45 & .630 & .862  & .844 & .895 \\ 
LFC         & 10 & .927 & .046  & 1.58 & .631 & .862  & .845 & \bf.895 \\
LFC         & 25 & .924 & .047 & 1.49 & .630 & .860  & .842 & .891 \\
LFC         & 50 & .923 & .048  & 1.48 & .630 & .860  & .841 & .893 \\
LFC         & 75 & .924 & 048  & 1.56 & .627 & .859  & .840 & .894 \\
LFC         & 90 & .925 & .046  & 1.47 & .626 & .859  & .841 & .894 \\
LFC** & 100  & .922 & .047  & 1.59 & .630 & .860  & .840 & .891  \\ \hline
\multicolumn{11}{c}{\textbf{High Frequency Components (HFC) with FFT}} \\
HFC & 10 & .926 & .046  &  1.60 &  .631 & .854  & .843 & .894 \\
\rowcolor{lightgray!30} HFC & 25 &\bf .928 & \bf.045  & \bf 1.35 &  \bf .636 &  \bf .862  & \bf.846 & \bf.895  \\
HFC & 50 & .925 & .047  & 1.51 & .631 & .862  & .843 & .894  \\
HFC & 75 & .923 & .048 & 1.52 & .629 & .858  & .842 & .892  \\
HFC & 90 & .924 & .047  & 1.49 & .630 & .861  & .842 & .893\\
\bottomrule
\end{tabular}
\caption{Ablation on HFC, LFC, and mask ratio $\tau$. Leveraging FFT to extract high-frequency components consistently outperforms LFC. *LFC~($\tau$=0) equals to a full zero image for prompting. **LFC~($\tau$=100) means we learn an embedding from the original input image directly. }
\vspace{-1em}
\label{tab:hfc}
\end{table*}

\section{Additional Visual Results}
\label{sec:visulation}

\begin{figure*}[t]
  \captionsetup[subfigure]{position=b}
  \centering

  \setlength{\tabcolsep}{0pt}
  \subcaptionbox{\scriptsize{Input}}{%
    \begin{tabular}{c}
      \includegraphics[width=0.19\textwidth, height=0.15\textwidth]{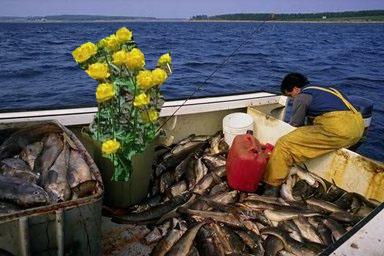} \\[0.1em]
      \includegraphics[width=0.19\textwidth, height=0.15\textwidth]{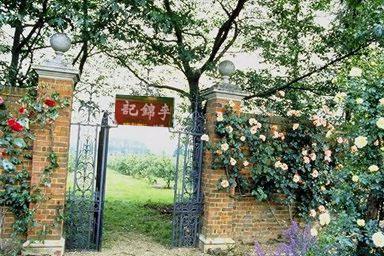} \\[0.1em]
      \includegraphics[width=0.19\textwidth, height=0.15\textwidth]{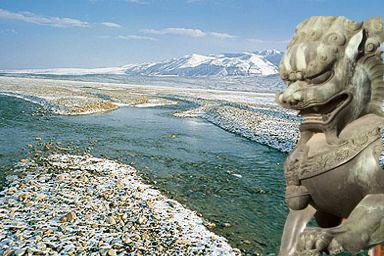}\\[0.1em]
      \includegraphics[width=0.19\textwidth, height=0.15\textwidth]{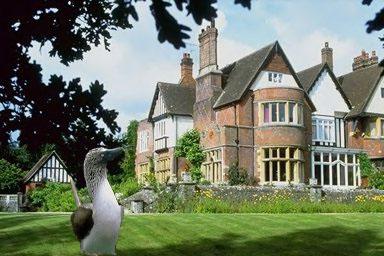}\\[0.1em]
      \includegraphics[width=0.19\textwidth, height=0.15\textwidth]{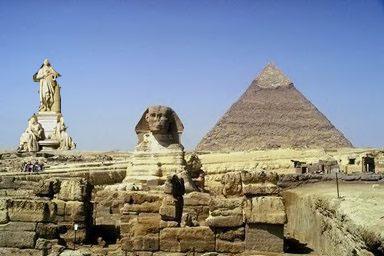}
    \end{tabular}}%
  \hspace{0.1em}%
  \subcaptionbox{\scriptsize{GT}}{%
    \begin{tabular}{c}
      \includegraphics[width=0.19\textwidth, height=0.15\textwidth]{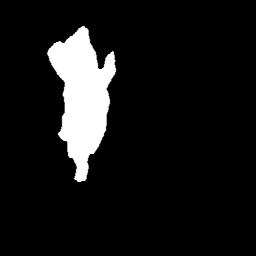}\\[0.1em]
      \includegraphics[width=0.19\textwidth, height=0.15\textwidth]{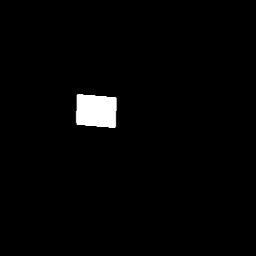} \\[0.1em]
      \includegraphics[width=0.19\textwidth, height=0.15\textwidth]{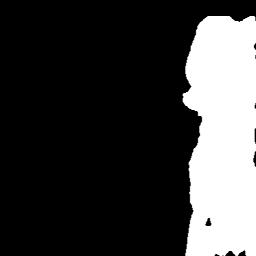}\\[0.1em]
      \includegraphics[width=0.19\textwidth, height=0.15\textwidth]{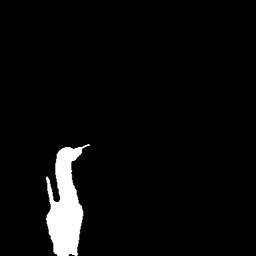}\\[0.1em]
      \includegraphics[width=0.19\textwidth, height=0.15\textwidth]{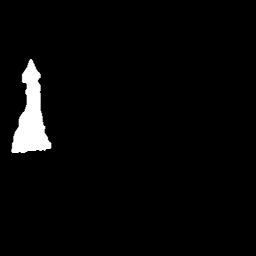}
    \end{tabular}}%
  \hspace{0.1em}%
  \subcaptionbox{\scriptsize{Ours}}{%
    \begin{tabular}{c}
      \includegraphics[width=0.19\textwidth, height=0.15\textwidth]{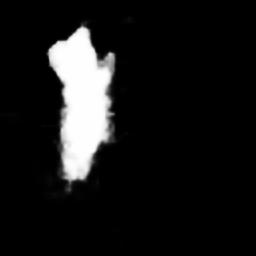} \\[0.1em]
      \includegraphics[width=0.19\textwidth, height=0.15\textwidth]{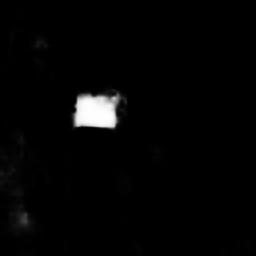}\\[0.1em]
      \includegraphics[width=0.19\textwidth, height=0.15\textwidth]{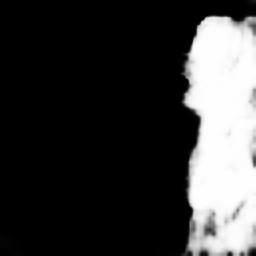}\\[0.1em]
      \includegraphics[width=0.19\textwidth, height=0.15\textwidth]{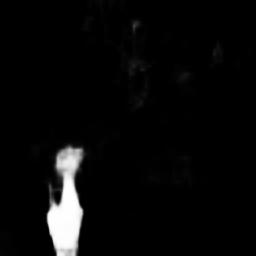}\\[0.1em]
      \includegraphics[width=0.19\textwidth, height=0.15\textwidth]{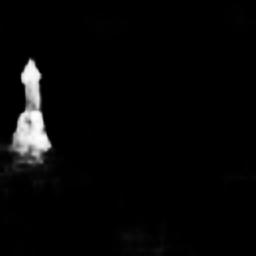}
    \end{tabular}}%
  \hspace{0.1em}%
  \subcaptionbox{\scriptsize{ManTraNet}}{%
    \begin{tabular}{c}
      \includegraphics[width=0.19\textwidth, height=0.15\textwidth]{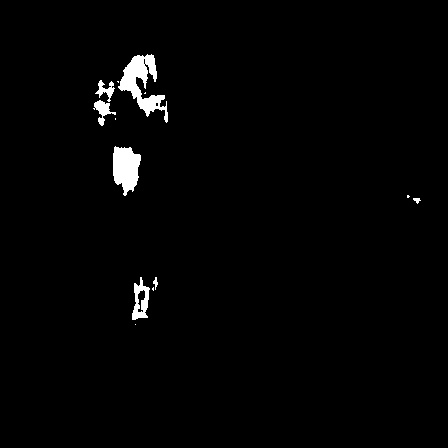} \\[0.1em]
      \includegraphics[width=0.19\textwidth, height=0.15\textwidth]{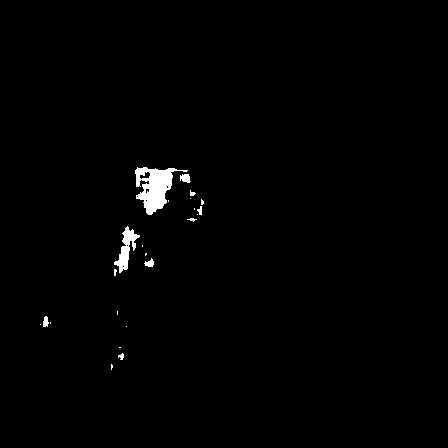} \\[0.1em]
      \includegraphics[width=0.19\textwidth, height=0.15\textwidth]{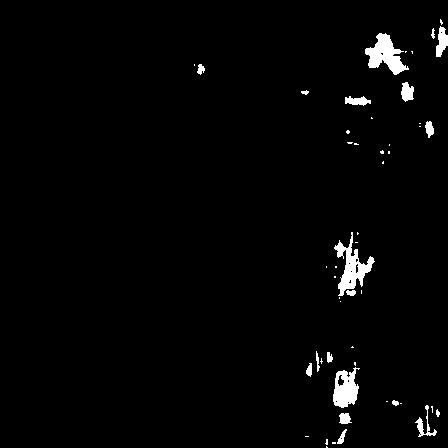}\\[0.1em]
      \includegraphics[width=0.19\textwidth, height=0.15\textwidth]{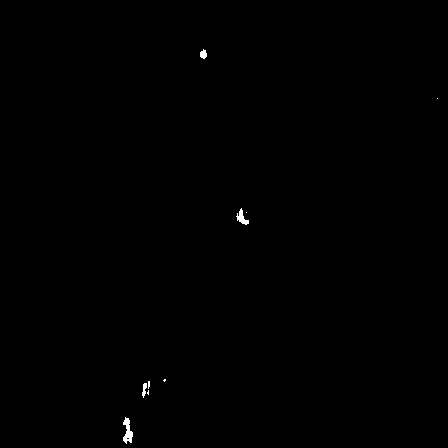}\\[0.1em]
      \includegraphics[width=0.19\textwidth, height=0.15\textwidth]{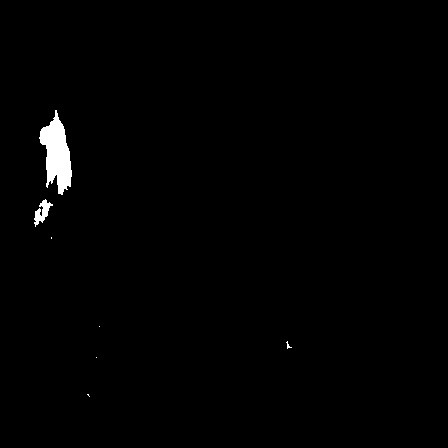}
    \end{tabular}}%
  \hspace{0.1em}%
  \subcaptionbox{\scriptsize{SPAN}}{%
    \begin{tabular}{c}
      \includegraphics[width=0.19\textwidth, height=0.15\textwidth]{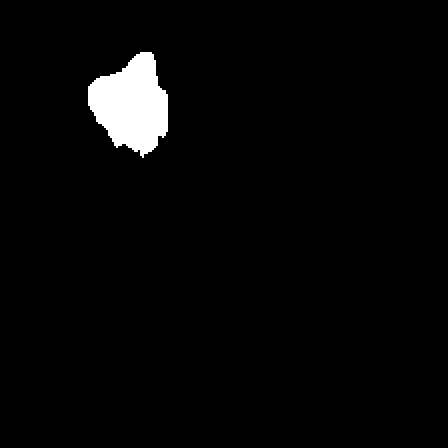}\\[0.1em]
      \includegraphics[width=0.19\textwidth, height=0.15\textwidth]{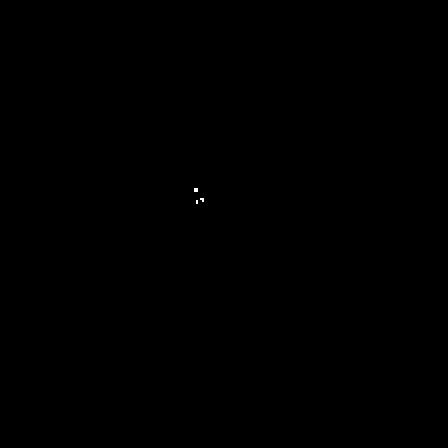} \\[0.1em]
      \includegraphics[width=0.19\textwidth, height=0.15\textwidth]{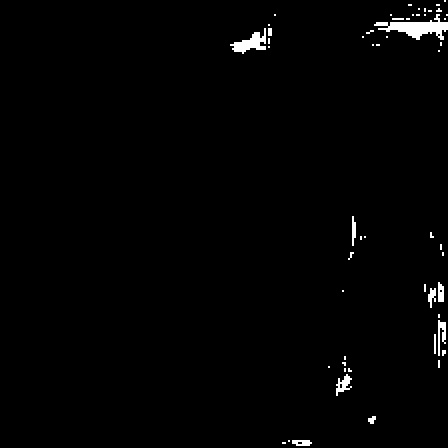}\\[0.1em]
      \includegraphics[width=0.19\textwidth, height=0.15\textwidth]{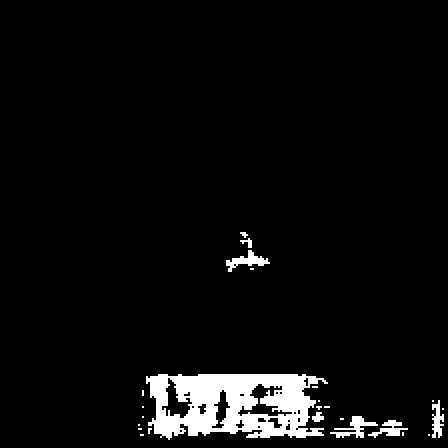}\\[0.1em]
      \includegraphics[width=0.19\textwidth, height=0.15\textwidth]{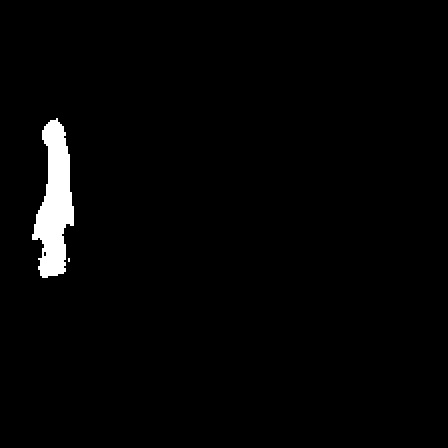}
    \end{tabular}}%
  \hspace{0.0em}%
  
\caption{More results on CAISA~\cite{dong2013casia} dataset for forgery detection. We compare to ManTraNet~\cite{wu2019mantra} and SPAN~\cite{hu2020span}.}
\label{fig:supp_forgery}
\end{figure*}

\begin{figure*}[t]
  \captionsetup[subfigure]{position=b}
  \centering
  \setlength{\tabcolsep}{0pt}
  \subcaptionbox{\scriptsize{Input}}{%
    \begin{tabular}{c}
      \includegraphics[width=0.15\textwidth, height=0.15\textwidth]{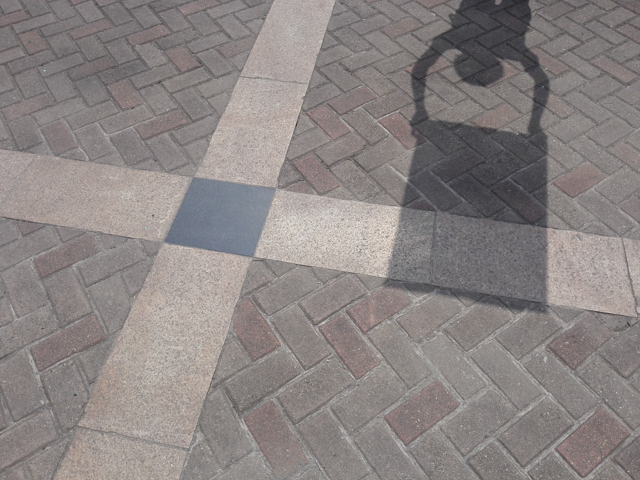} \\[0.1em]
      \includegraphics[width=0.15\textwidth, height=0.15\textwidth]{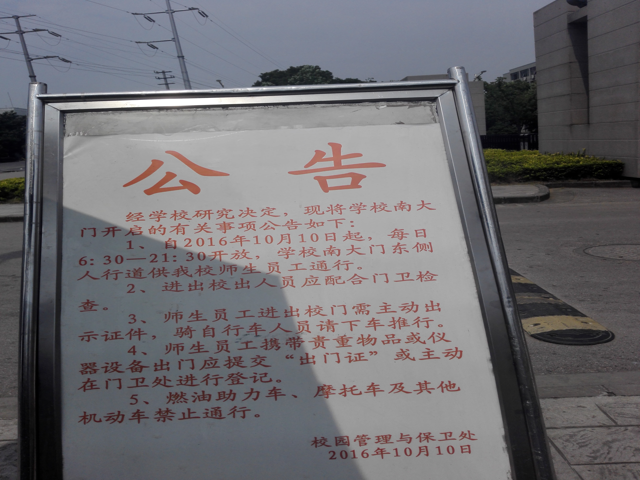} \\[0.1em]
      \includegraphics[width=0.15\textwidth, height=0.15\textwidth]{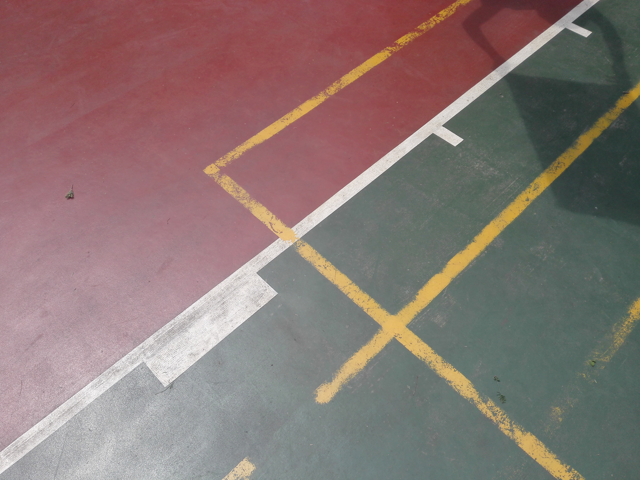}\\[0.1em]
      \includegraphics[width=0.15\textwidth, height=0.15\textwidth]{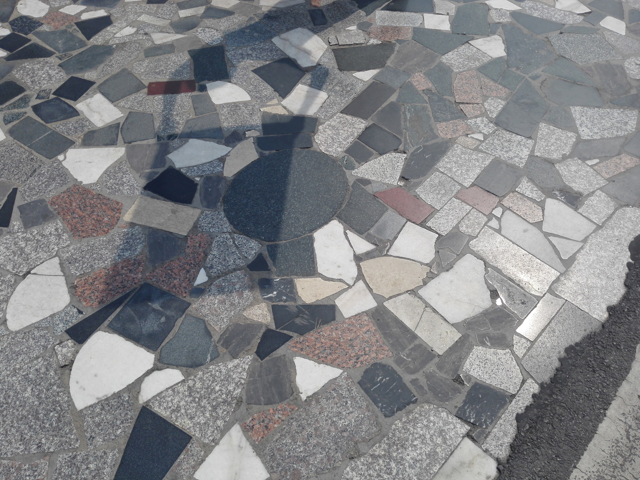}\\[0.1em]
      \includegraphics[width=0.15\textwidth, height=0.15\textwidth]{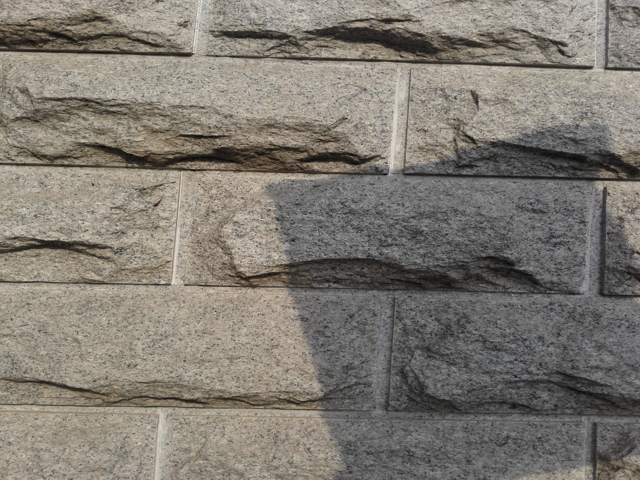}
    \end{tabular}}%
  \hspace{0.1em}%
  \subcaptionbox{\scriptsize{GT}}{%
    \begin{tabular}{c}
      \includegraphics[width=0.15\textwidth, height=0.15\textwidth]{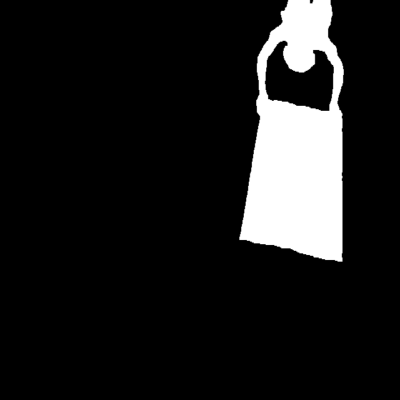} \\[0.1em]
      \includegraphics[width=0.15\textwidth, height=0.15\textwidth]{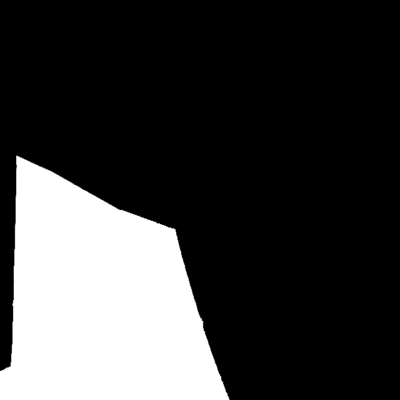} \\[0.1em]
      \includegraphics[width=0.15\textwidth, height=0.15\textwidth]{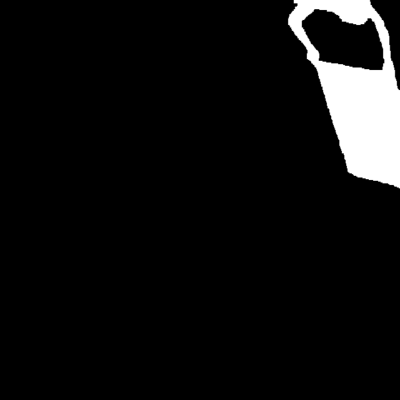}\\[0.1em]
      \includegraphics[width=0.15\textwidth, height=0.15\textwidth]{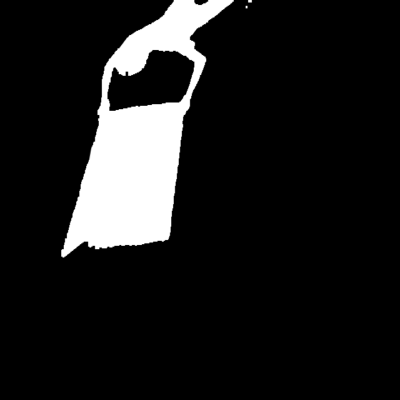}\\[0.1em]
      \includegraphics[width=0.15\textwidth, height=0.15\textwidth]{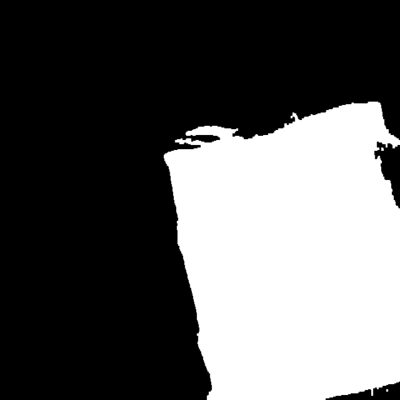}
    \end{tabular}}%
  \hspace{0.1em}%
  \subcaptionbox{\scriptsize{Ours}}{%
    \begin{tabular}{c}
      \includegraphics[width=0.15\textwidth, height=0.15\textwidth]{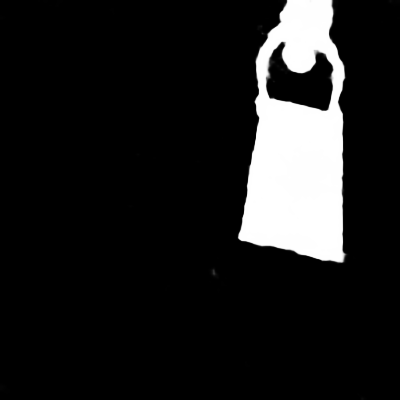} \\[0.1em]
      \includegraphics[width=0.15\textwidth, height=0.15\textwidth]{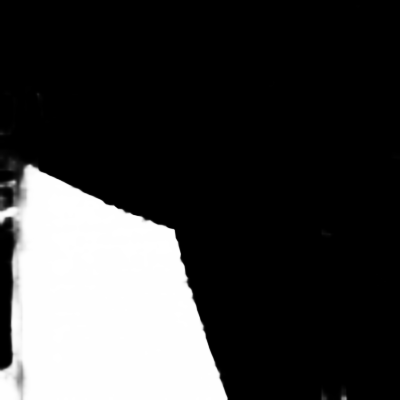} \\[0.1em]
      \includegraphics[width=0.15\textwidth, height=0.15\textwidth]{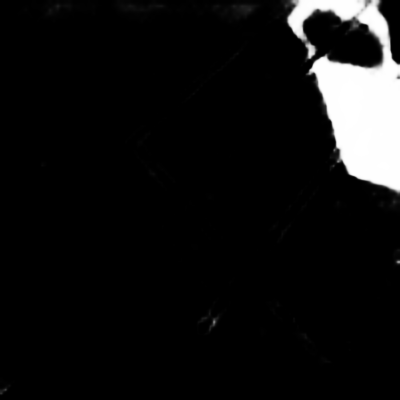}\\[0.1em]
      \includegraphics[width=0.15\textwidth, height=0.15\textwidth]{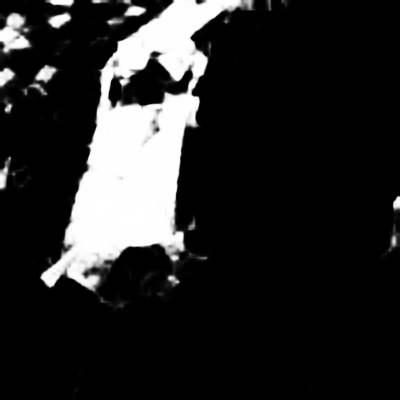}\\[0.1em]
      \includegraphics[width=0.15\textwidth, height=0.15\textwidth]{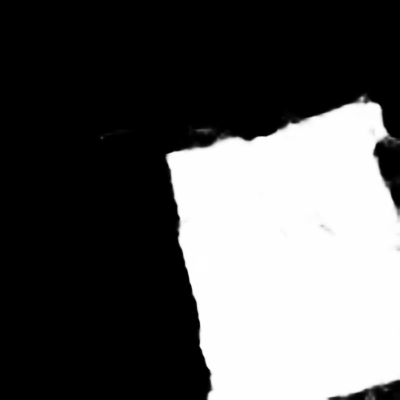}
    \end{tabular}}%
  \hspace{0.1em}%
  \subcaptionbox{\scriptsize{DSD}}{%
    \begin{tabular}{c}
      \includegraphics[width=0.15\textwidth, height=0.15\textwidth]{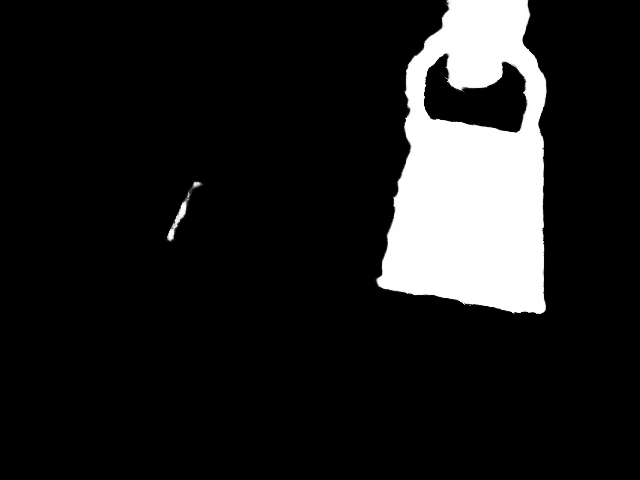} \\[0.1em]
      \includegraphics[width=0.15\textwidth, height=0.15\textwidth]{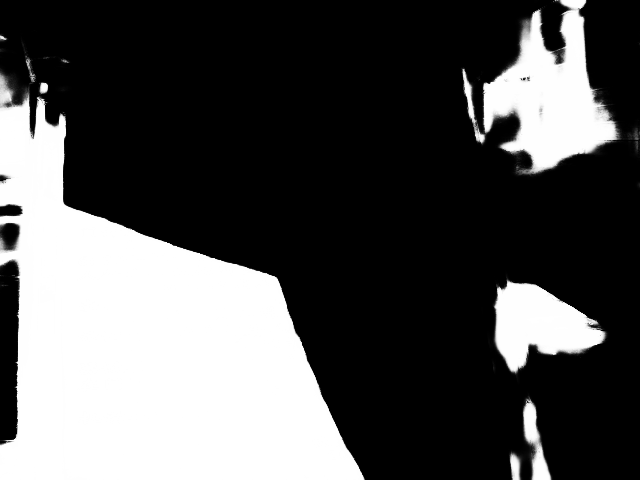} \\[0.1em]
      \includegraphics[width=0.15\textwidth, height=0.15\textwidth]{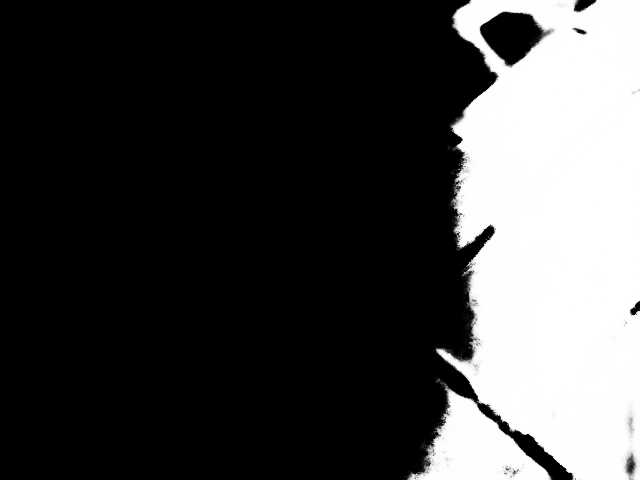}\\[0.1em]
      \includegraphics[width=0.15\textwidth, height=0.15\textwidth]{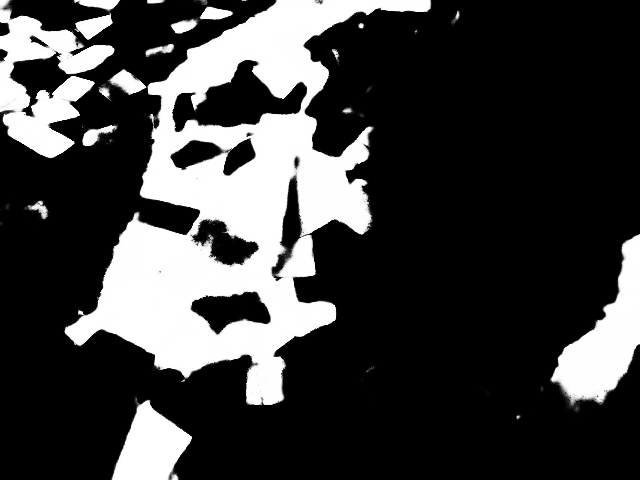}\\[0.1em]
      \includegraphics[width=0.15\textwidth, height=0.15\textwidth]{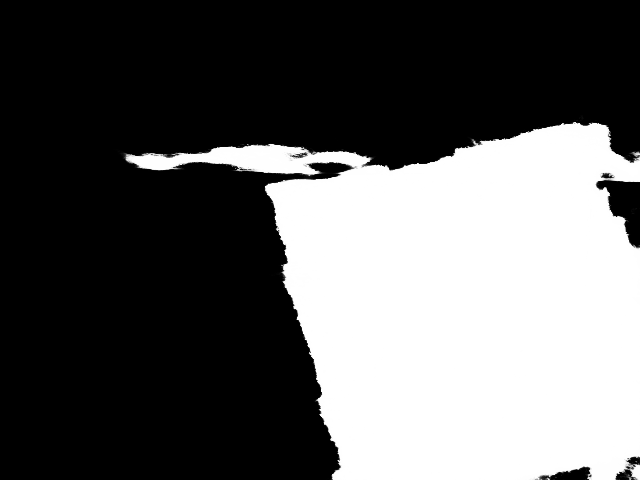}
    \end{tabular}}%
  \hspace{0.1em}%
  \subcaptionbox{\scriptsize{MTMT}}{%
    \begin{tabular}{c}
      \includegraphics[width=0.15\textwidth, height=0.15\textwidth]{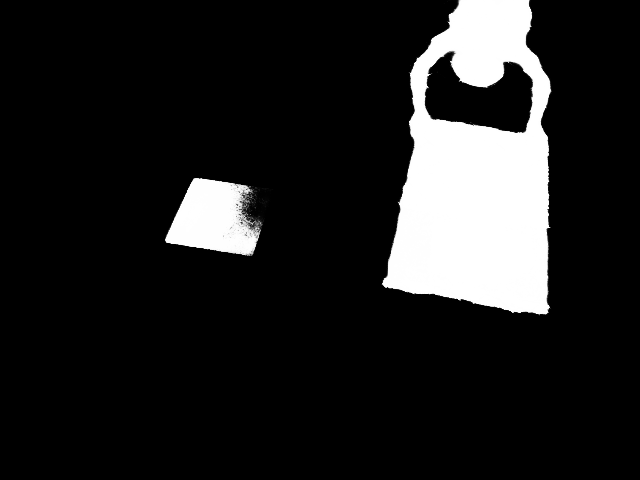} \\[0.1em]
      \includegraphics[width=0.15\textwidth, height=0.15\textwidth]{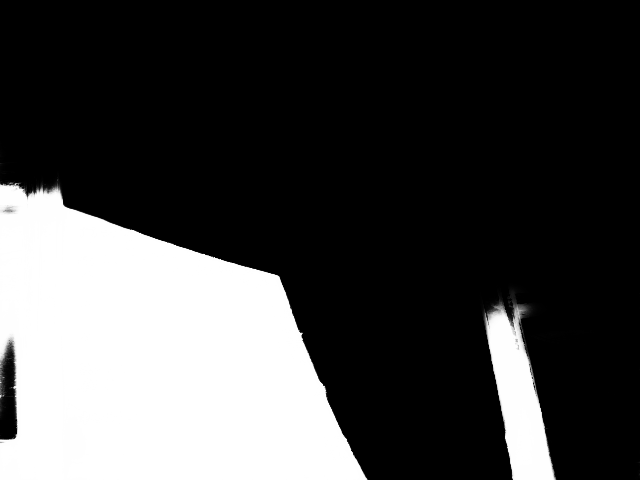} \\[0.1em]
      \includegraphics[width=0.15\textwidth, height=0.15\textwidth]{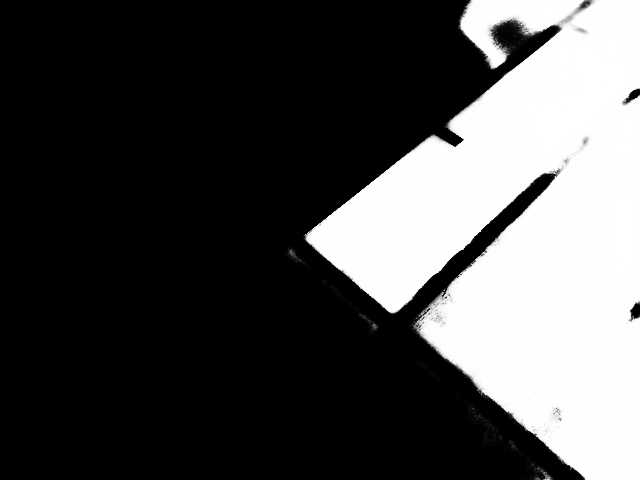}\\[0.1em]
      \includegraphics[width=0.15\textwidth, height=0.15\textwidth]{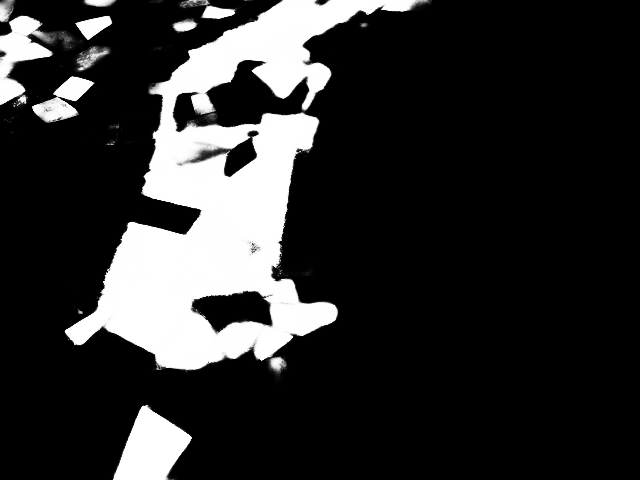}\\[0.1em]
      \includegraphics[width=0.15\textwidth, height=0.15\textwidth]{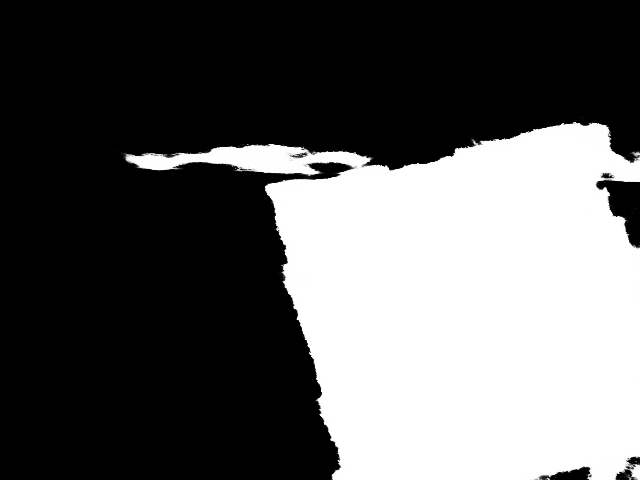}
    \end{tabular}}%
  \hspace{0.1em}%
  \subcaptionbox{\scriptsize{FDRNet}}{%
    \begin{tabular}{c}
      \includegraphics[width=0.15\textwidth, height=0.15\textwidth]{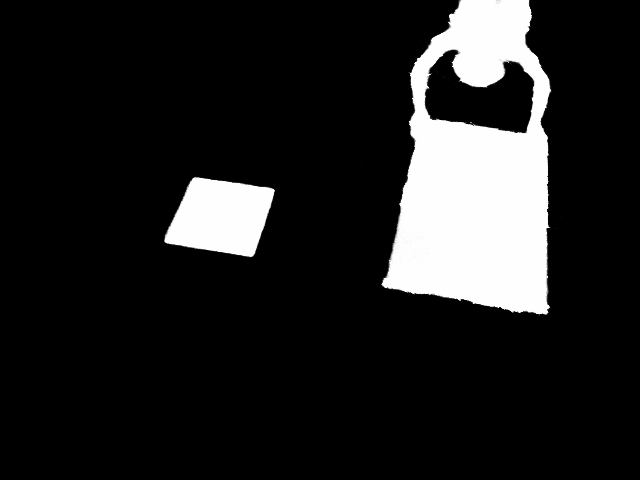} \\[0.1em]
      \includegraphics[width=0.15\textwidth, height=0.15\textwidth]{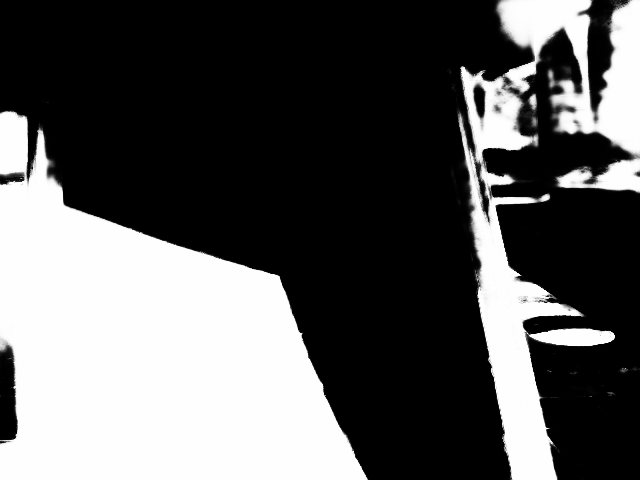} \\[0.1em]
      \includegraphics[width=0.15\textwidth, height=0.15\textwidth]{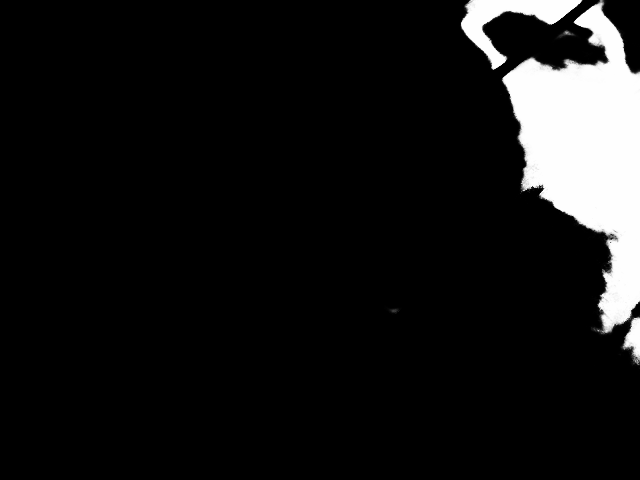}\\[0.1em]
      \includegraphics[width=0.15\textwidth, height=0.15\textwidth]{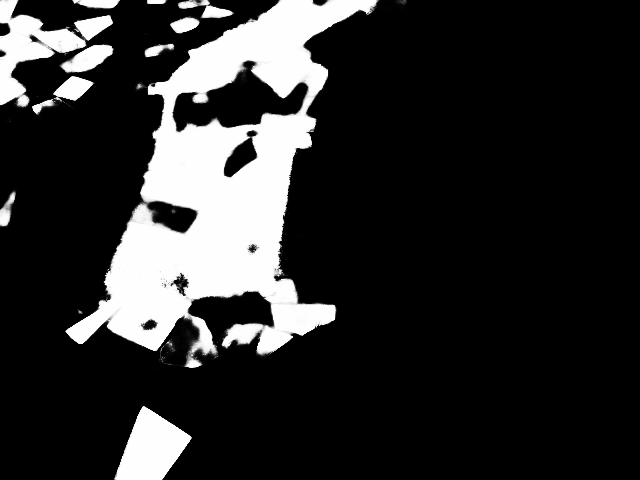}\\[0.1em]
      \includegraphics[width=0.15\textwidth, height=0.15\textwidth]{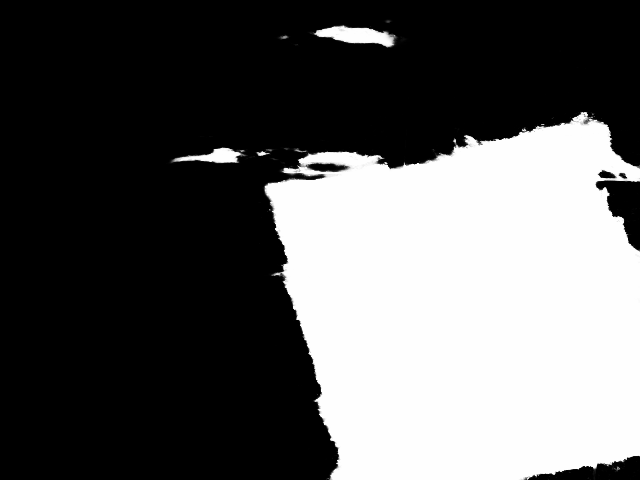}
    \end{tabular}}%
  \hspace{0.0em}%
  
\caption{More results on ISTD~\cite{wang2018stacked} dataset for shadow detection. We compare to DSD~\cite{zhao2021self}, MTMT~\cite{mtmt}, FDRNet~\cite{zhu2021mitigating}.}
\label{fig:supp_shadow}
\end{figure*}

\begin{figure*}[t]
  \captionsetup[subfigure]{position=b}
  \centering
  \setlength{\tabcolsep}{0pt}
  \subcaptionbox{\scriptsize{Input}}{%
    \begin{tabular}{c}
      \includegraphics[width=0.15\textwidth, height=0.15\textwidth]{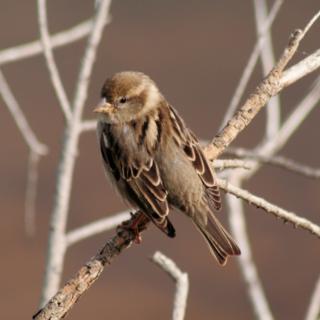} \\[0.1em]
      \includegraphics[width=0.15\textwidth, height=0.15\textwidth]{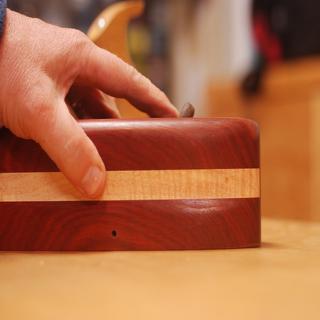} \\[0.1em]
      \includegraphics[width=0.15\textwidth, height=0.15\textwidth]{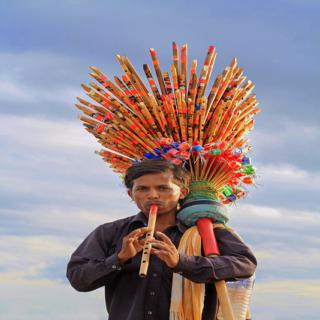} \\[0.1em]
      \includegraphics[width=0.15\textwidth, height=0.15\textwidth]{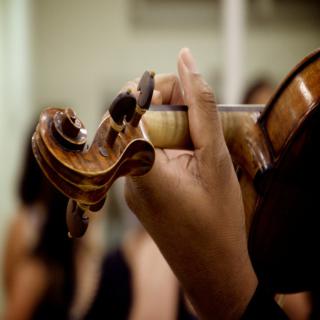} \\[0.1em]
      \includegraphics[width=0.15\textwidth, height=0.15\textwidth]{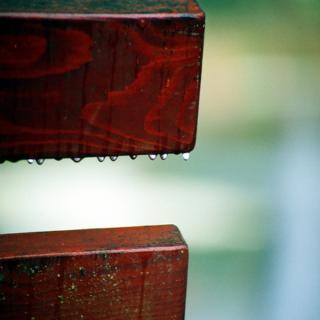}
    \end{tabular}}%
  \hspace{0.1em}%
  \subcaptionbox{\scriptsize{GT}}{%
    \begin{tabular}{c}
      \includegraphics[width=0.15\textwidth, height=0.15\textwidth]{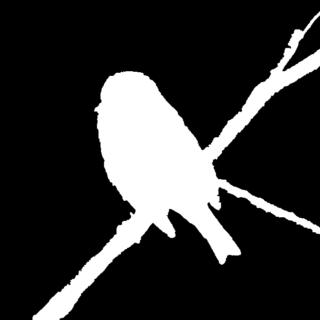} \\[0.1em]
      \includegraphics[width=0.15\textwidth, height=0.15\textwidth]{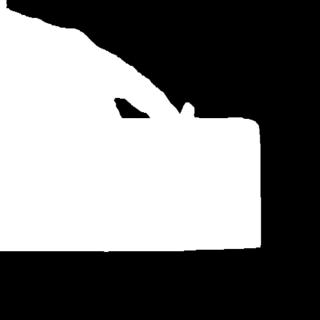} \\[0.1em]
      \includegraphics[width=0.15\textwidth, height=0.15\textwidth]{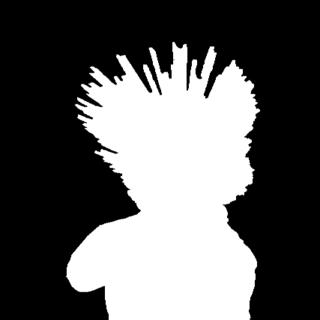} \\[0.1em]
      \includegraphics[width=0.15\textwidth, height=0.15\textwidth]{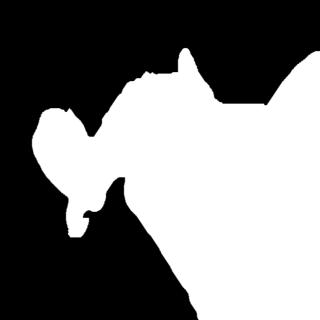} \\[0.1em]
      \includegraphics[width=0.15\textwidth, height=0.15\textwidth]{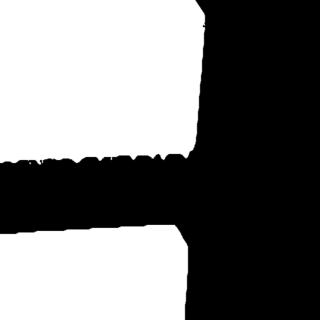}
    \end{tabular}}%
  \hspace{0.1em}%
  \subcaptionbox{\scriptsize{Ours}}{%
    \begin{tabular}{c}
      \includegraphics[width=0.15\textwidth, height=0.15\textwidth]{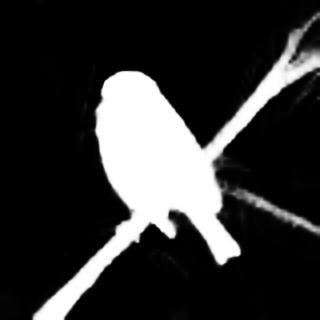} \\[0.1em]
      \includegraphics[width=0.15\textwidth, height=0.15\textwidth]{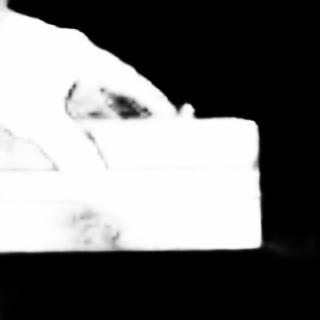} \\[0.1em]
      \includegraphics[width=0.15\textwidth, height=0.15\textwidth]{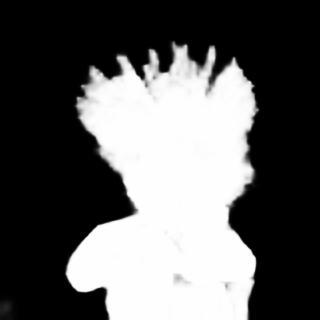} \\[0.1em]
      \includegraphics[width=0.15\textwidth, height=0.15\textwidth]{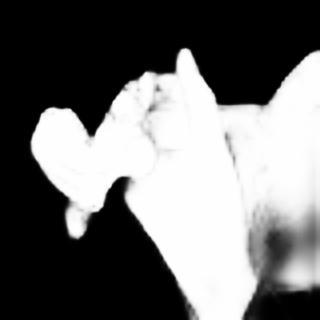} \\[0.1em]
      \includegraphics[width=0.15\textwidth, height=0.15\textwidth]{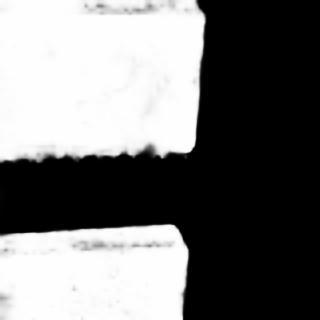}
    \end{tabular}}%
  \hspace{0.1em}%
  \subcaptionbox{\scriptsize{BTBNet}}{%
    \begin{tabular}{c}
      \includegraphics[width=0.15\textwidth, height=0.15\textwidth]{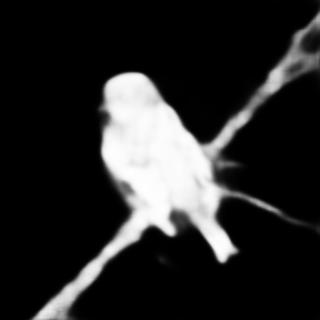} \\[0.1em]
      \includegraphics[width=0.15\textwidth, height=0.15\textwidth]{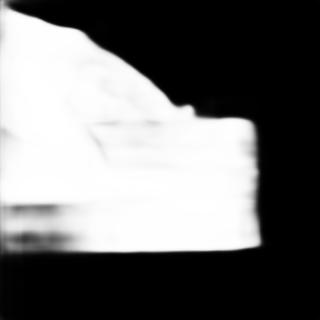} \\[0.1em]
      \includegraphics[width=0.15\textwidth, height=0.15\textwidth]{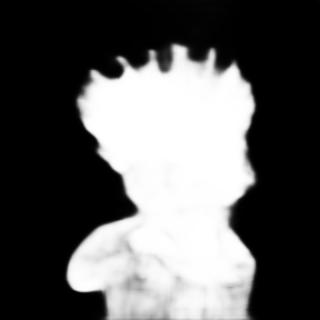} \\[0.1em]
      \includegraphics[width=0.15\textwidth, height=0.15\textwidth]{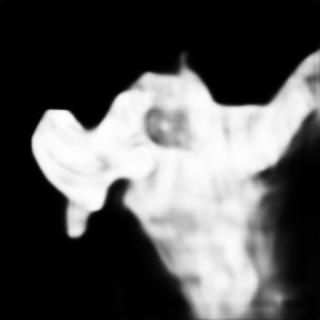} \\[0.1em]
      \includegraphics[width=0.15\textwidth, height=0.15\textwidth]{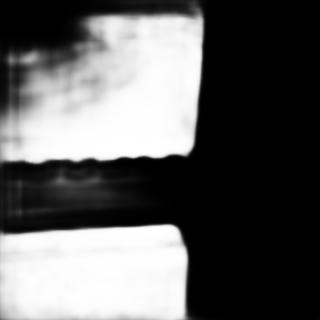}
    \end{tabular}}%
  \hspace{0.1em}%
  \subcaptionbox{\scriptsize{CENet}}{%
    \begin{tabular}{c}
      \includegraphics[width=0.15\textwidth, height=0.15\textwidth]{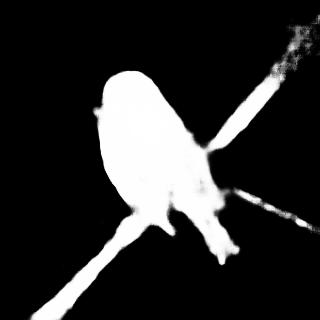} \\[0.1em]
      \includegraphics[width=0.15\textwidth, height=0.15\textwidth]{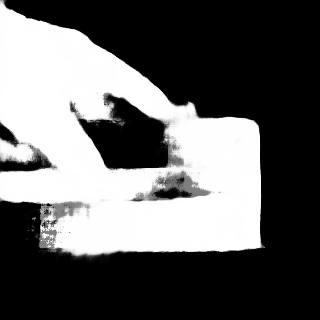} \\[0.1em]
      \includegraphics[width=0.15\textwidth, height=0.15\textwidth]{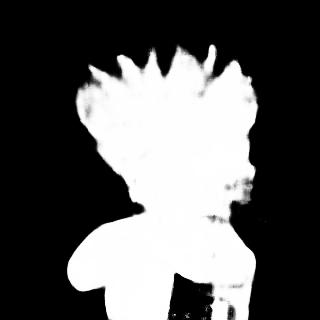} \\[0.1em]
      \includegraphics[width=0.15\textwidth, height=0.15\textwidth]{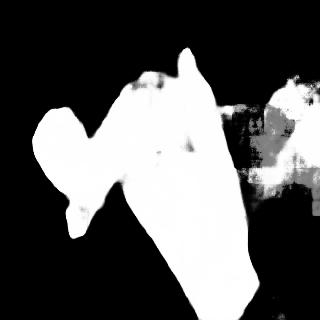} \\[0.1em]
      \includegraphics[width=0.15\textwidth, height=0.15\textwidth]{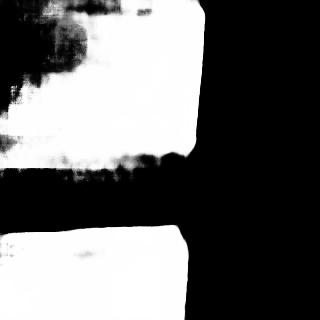}
    \end{tabular}}%
  \hspace{0.1em}%
  \subcaptionbox{\scriptsize{EFENet}}{%
    \begin{tabular}{c}
      \includegraphics[width=0.15\textwidth, height=0.15\textwidth]{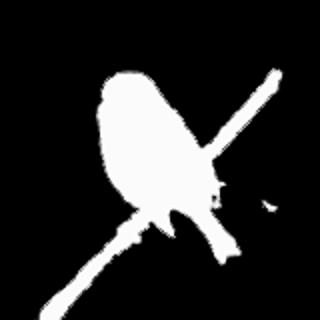} \\[0.1em]
      \includegraphics[width=0.15\textwidth, height=0.15\textwidth]{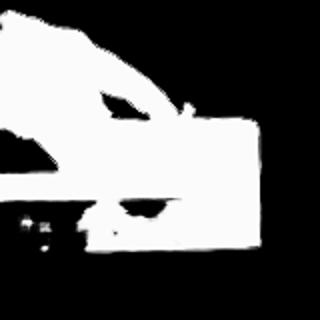} \\[0.1em]
      \includegraphics[width=0.15\textwidth, height=0.15\textwidth]{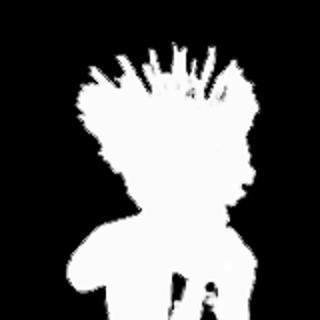} \\[0.1em]
      \includegraphics[width=0.15\textwidth, height=0.15\textwidth]{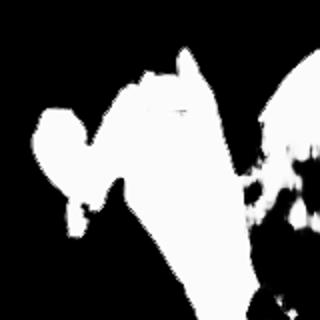} \\[0.1em]
      \includegraphics[width=0.15\textwidth, height=0.15\textwidth]{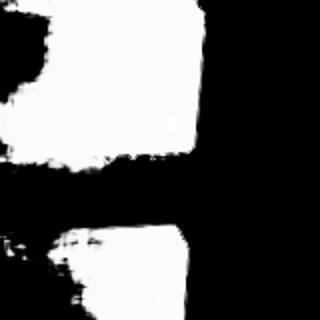}
    \end{tabular}}%
  \hspace{0.0em}%
  
\caption{More results on CUHK~\cite{shi2014discriminative} dataset for defocus blur detection.We compare to BTBNet~\cite{zhao2019btbnet}, CENet~\cite{zhao2019cenet} and EFENet~\cite{zhao2021defocus}.}
\label{fig:supp_defocus}
\end{figure*}

\begin{figure*}[t]
  \captionsetup[subfigure]{position=b}
  \centering
  \setlength{\tabcolsep}{0pt}
  \subcaptionbox{\scriptsize{Input}}{%
    \begin{tabular}{c}
      \includegraphics[width=0.13\textwidth, height=0.13\textwidth]{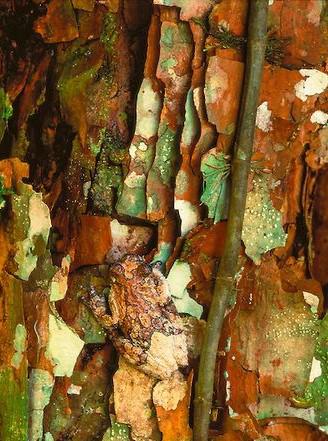} \\[0.1em]
      \includegraphics[width=0.13\textwidth, height=0.13\textwidth]{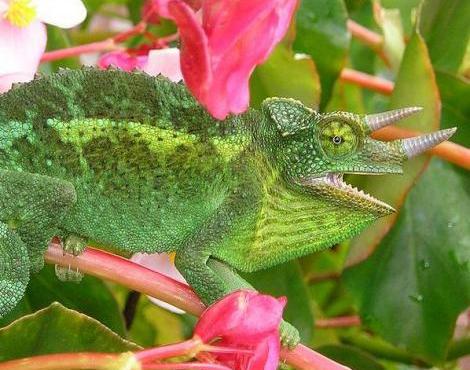} \\[0.1em]
      \includegraphics[width=0.13\textwidth, height=0.13\textwidth]{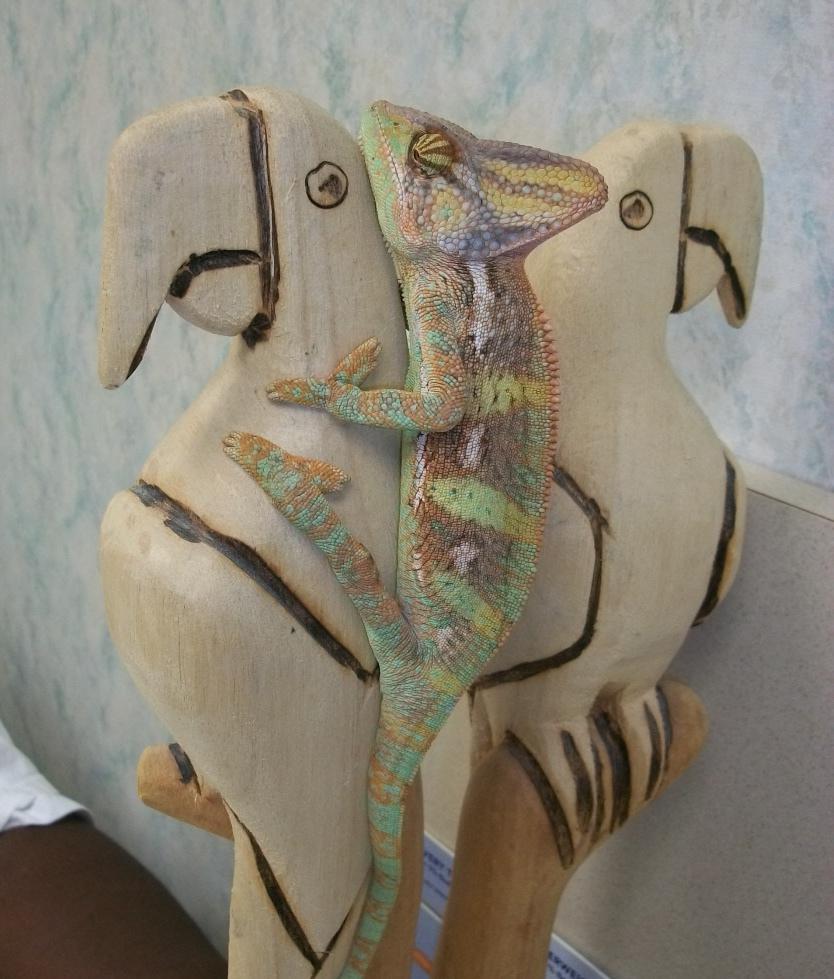}\\[0.1em]
      \includegraphics[width=0.13\textwidth, height=0.13\textwidth]{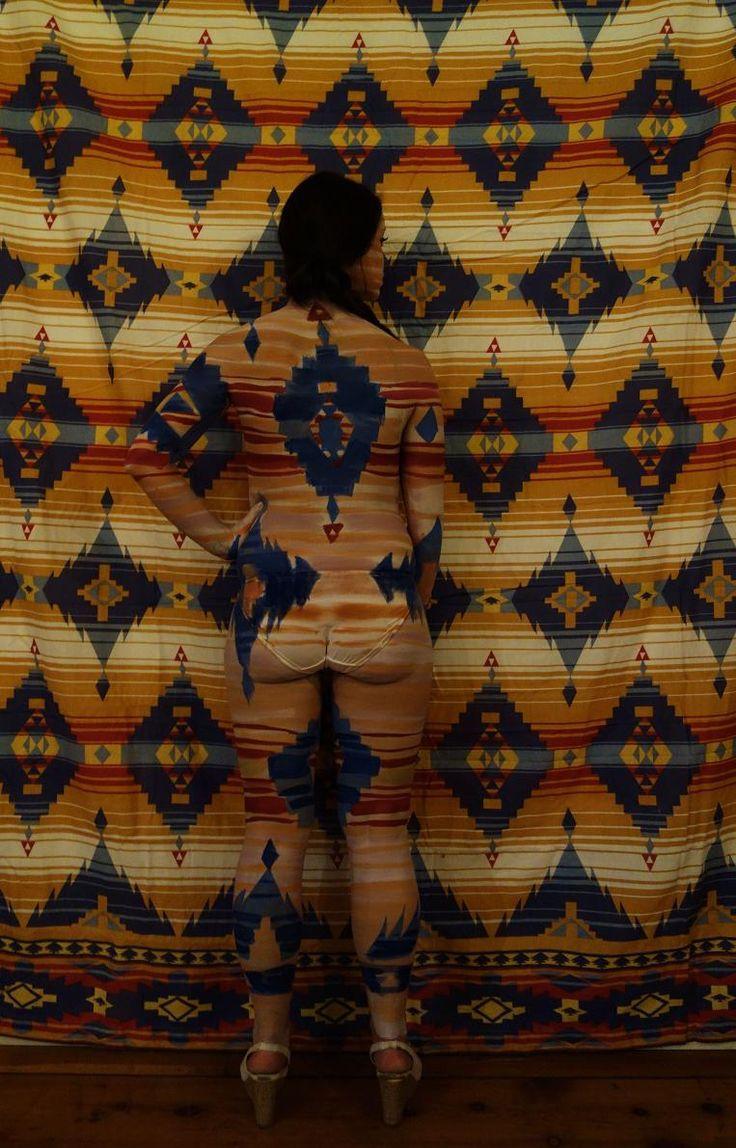}\\[0.1em]
      \includegraphics[width=0.13\textwidth, height=0.13\textwidth]{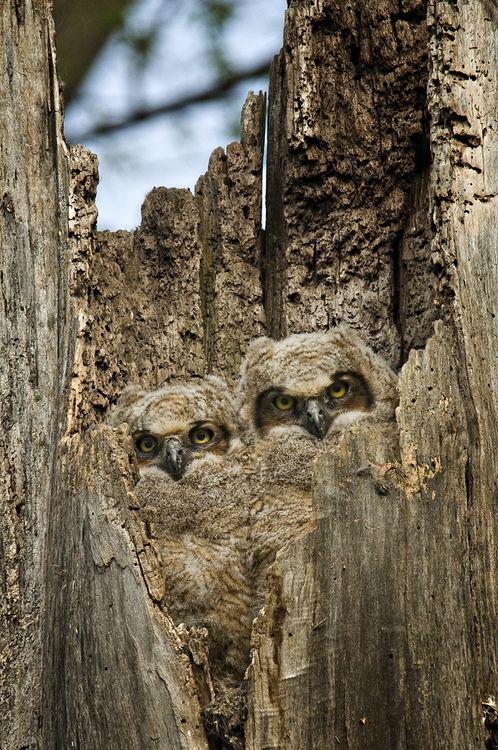}
    \end{tabular}}%
  \hspace{0.1em}%
  \subcaptionbox{\scriptsize{GT}}{%
    \begin{tabular}{c}
      \includegraphics[width=0.13\textwidth, height=0.13\textwidth]{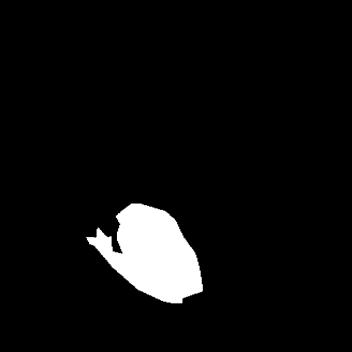} \\[0.1em]
      \includegraphics[width=0.13\textwidth, height=0.13\textwidth]{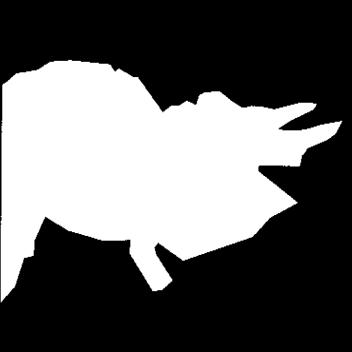} \\[0.1em]
      \includegraphics[width=0.13\textwidth, height=0.13\textwidth]{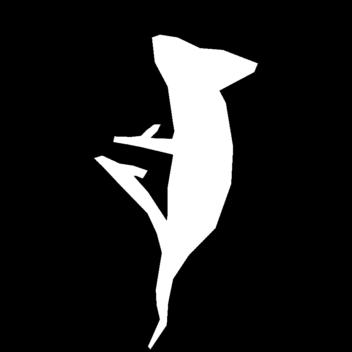}\\[0.1em]
      \includegraphics[width=0.13\textwidth, height=0.13\textwidth]{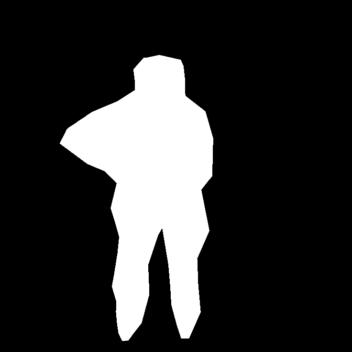}\\[0.1em]
      \includegraphics[width=0.13\textwidth, height=0.13\textwidth]{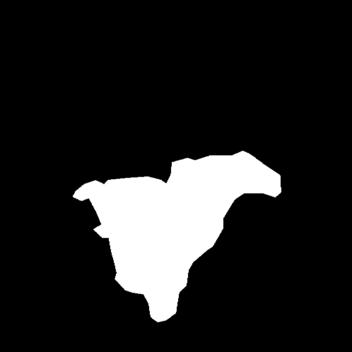}
    \end{tabular}}%
  \hspace{0.1em}%
  \subcaptionbox{\scriptsize{Ours}}{%
    \begin{tabular}{c}
      \includegraphics[width=0.13\textwidth, height=0.13\textwidth]{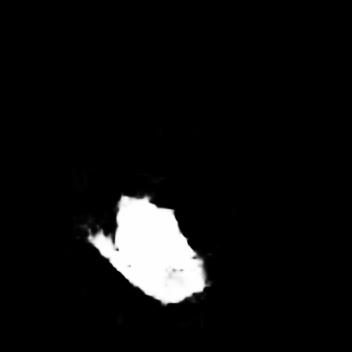} \\[0.1em]
      \includegraphics[width=0.13\textwidth, height=0.13\textwidth]{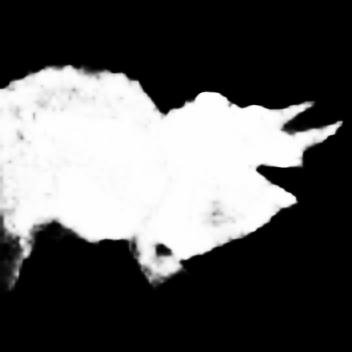} \\[0.1em]
      \includegraphics[width=0.13\textwidth, height=0.13\textwidth]{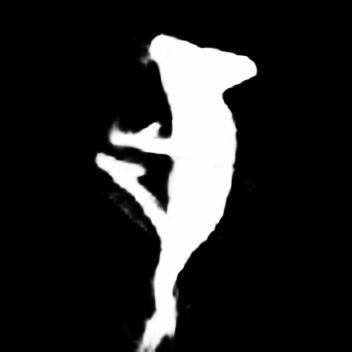}\\[0.1em]
      \includegraphics[width=0.13\textwidth, height=0.13\textwidth]{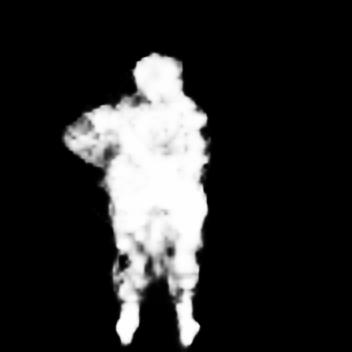}\\[0.1em]
      \includegraphics[width=0.13\textwidth, height=0.13\textwidth]{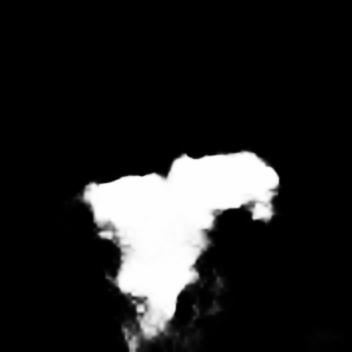}
    \end{tabular}}%
  \hspace{0.1em}%
  \subcaptionbox{\scriptsize{SINet}}{%
    \begin{tabular}{c}
      \includegraphics[width=0.13\textwidth, height=0.13\textwidth]{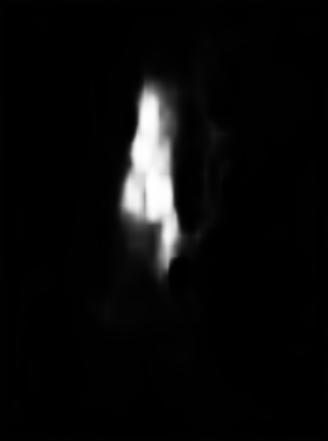} \\[0.1em]
      \includegraphics[width=0.13\textwidth, height=0.13\textwidth]{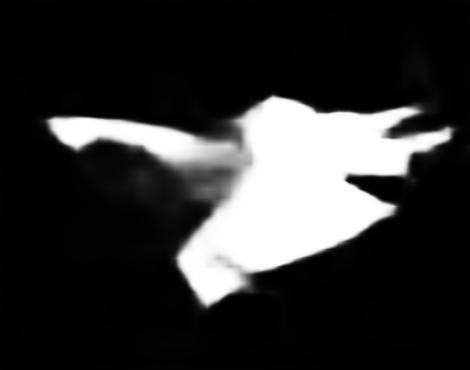} \\[0.1em]
      \includegraphics[width=0.13\textwidth, height=0.13\textwidth]{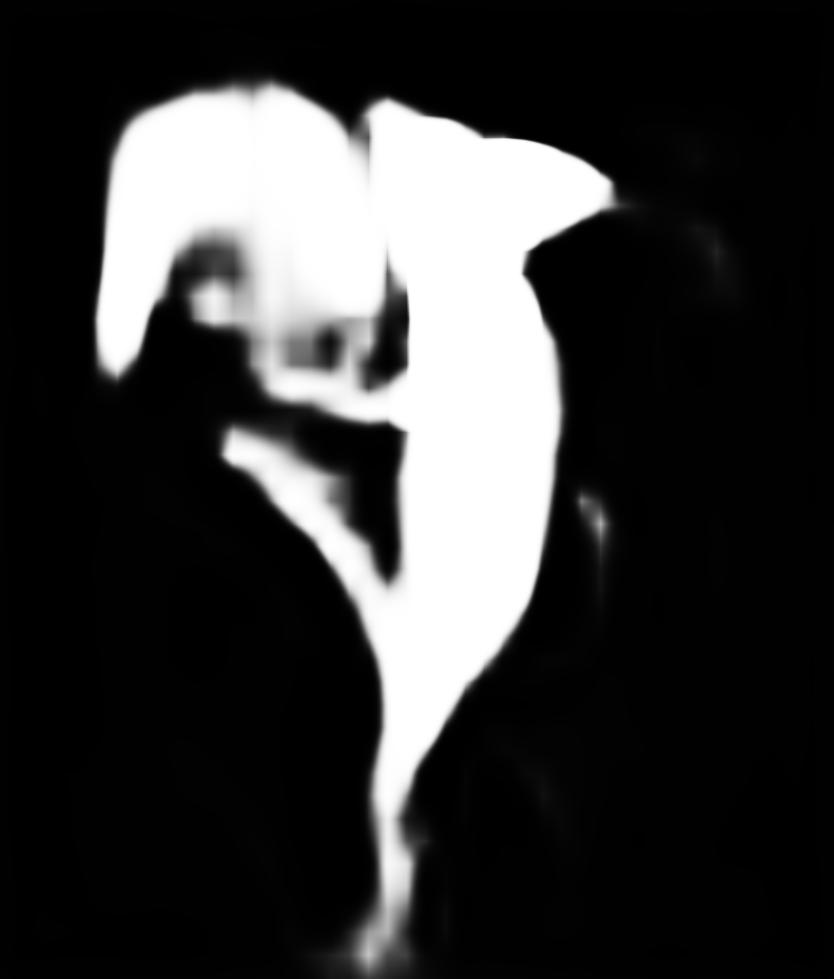}\\[0.1em]
      \includegraphics[width=0.13\textwidth, height=0.13\textwidth]{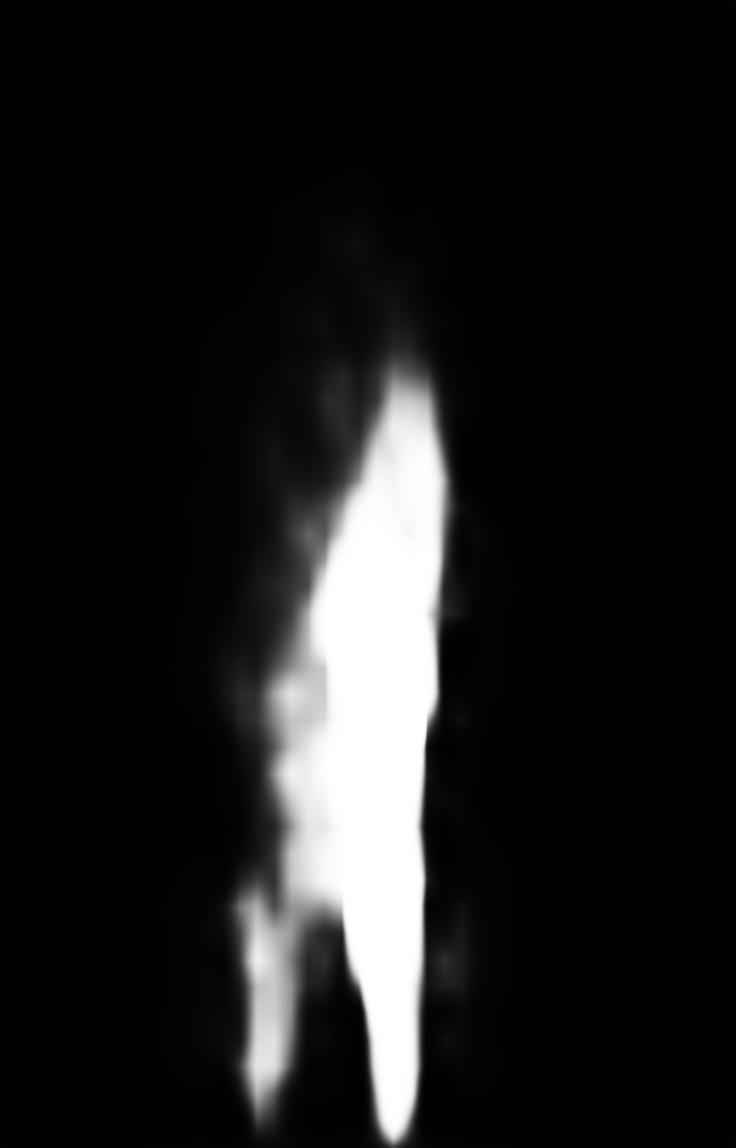}\\[0.1em]
      \includegraphics[width=0.13\textwidth, height=0.13\textwidth]{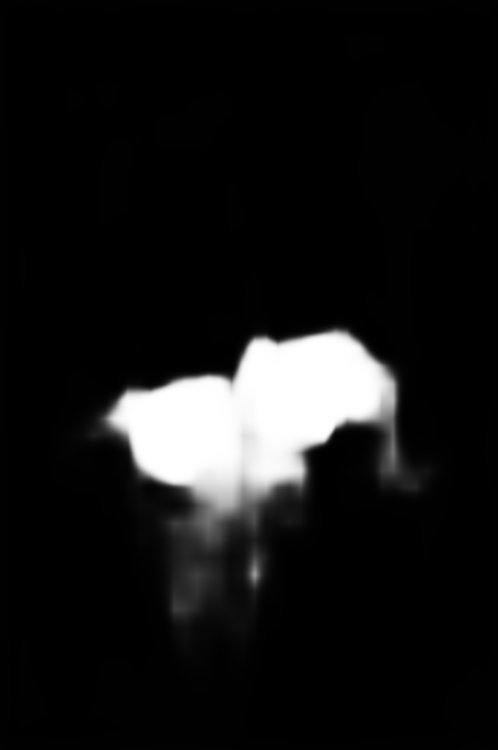}
    \end{tabular}}%
  \hspace{0.1em}%
  \subcaptionbox{\scriptsize{PFNet}}{%
    \begin{tabular}{c}
      \includegraphics[width=0.13\textwidth, height=0.13\textwidth]{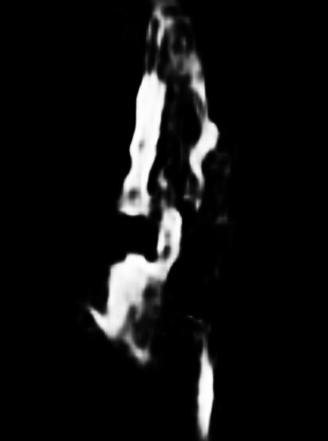} \\[0.1em]
      \includegraphics[width=0.13\textwidth, height=0.13\textwidth]{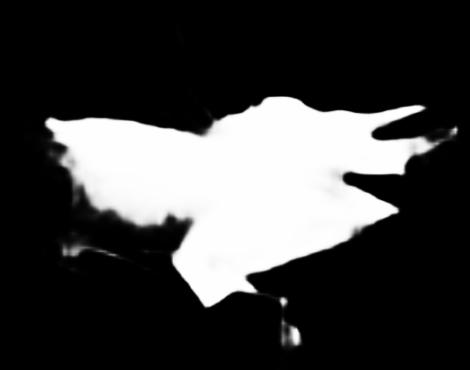} \\[0.1em]
      \includegraphics[width=0.13\textwidth, height=0.13\textwidth]{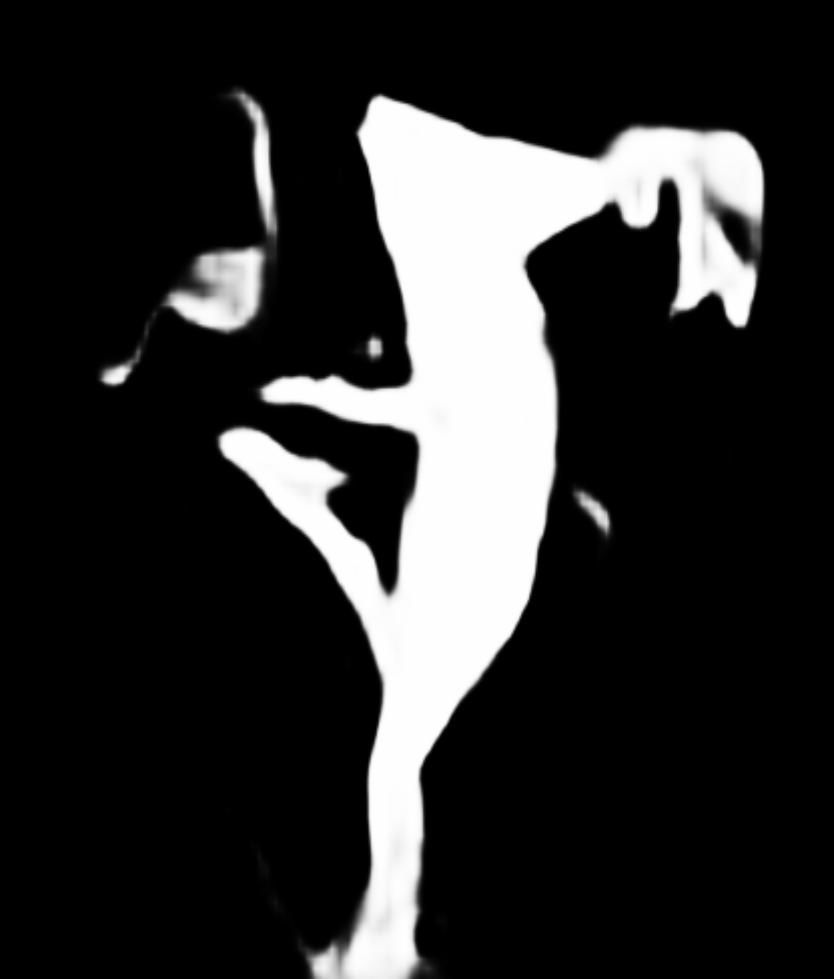}\\[0.1em]
      \includegraphics[width=0.13\textwidth, height=0.13\textwidth]{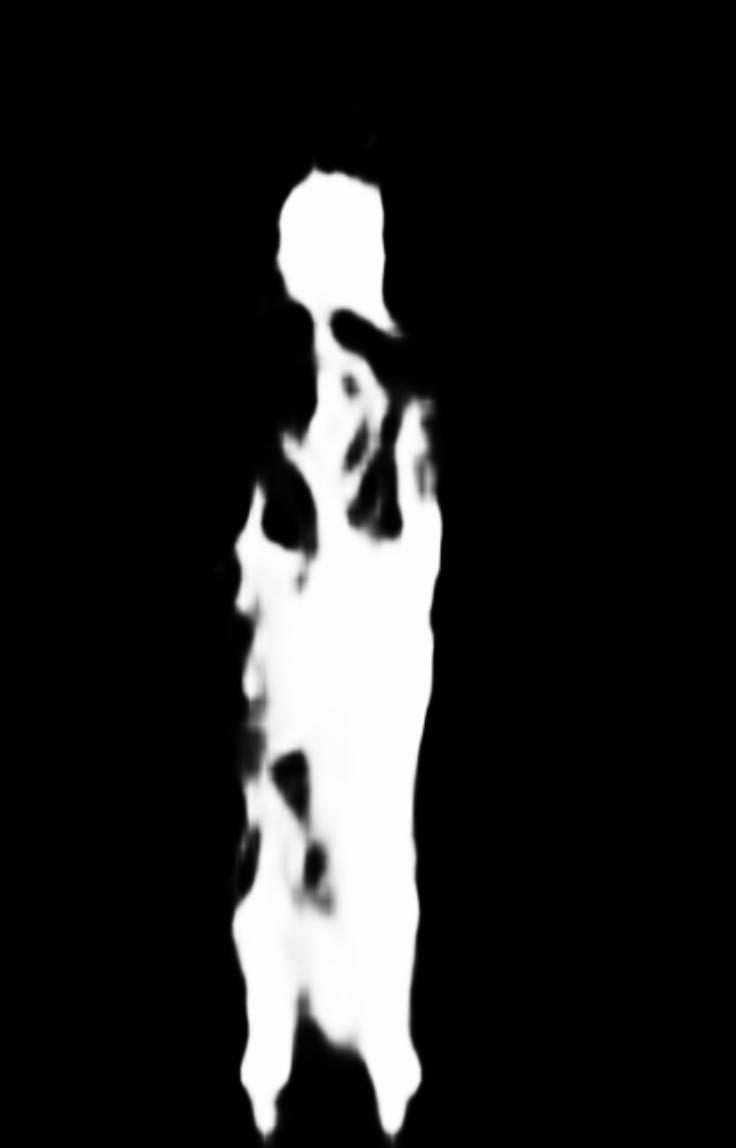}\\[0.1em]
      \includegraphics[width=0.13\textwidth, height=0.13\textwidth]{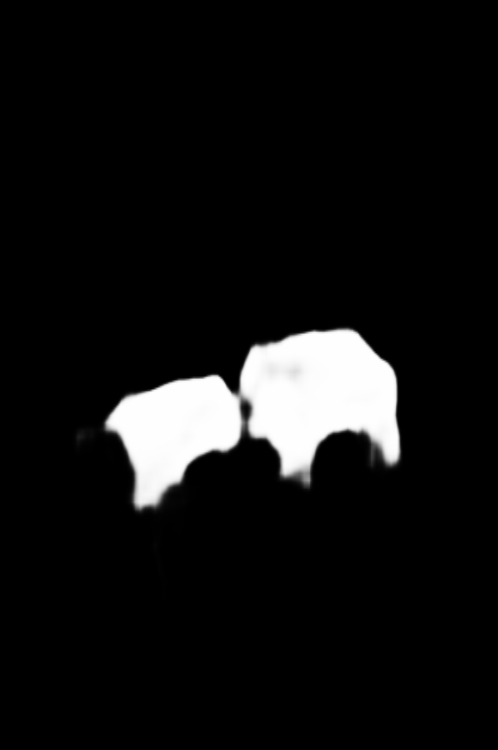}
    \end{tabular}}%
  \hspace{0.1em}%
  \subcaptionbox{\scriptsize{JCOD}}{%
    \begin{tabular}{c}
      \includegraphics[width=0.13\textwidth, height=0.13\textwidth]{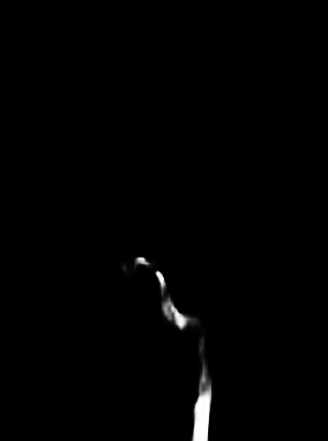} \\[0.1em]
      \includegraphics[width=0.13\textwidth, height=0.13\textwidth]{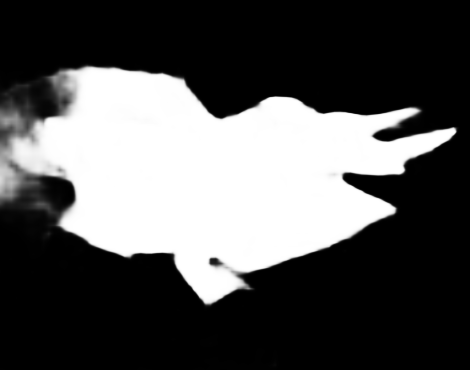} \\[0.1em]
      \includegraphics[width=0.13\textwidth, height=0.13\textwidth]{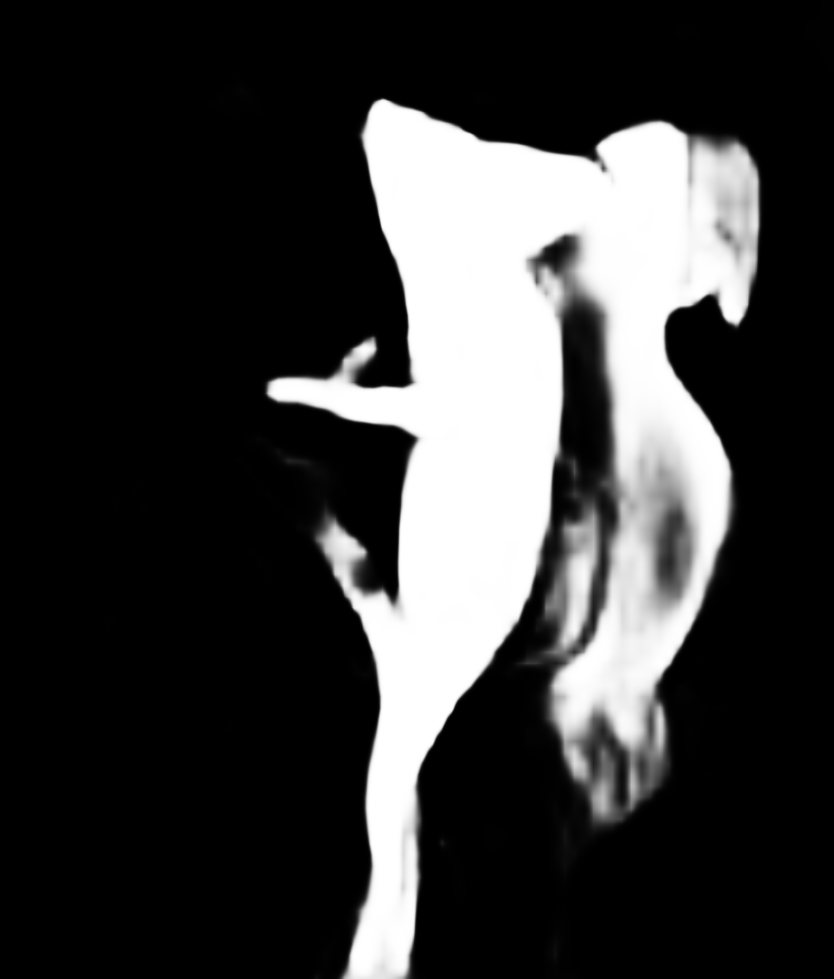}\\[0.1em]
      \includegraphics[width=0.13\textwidth, height=0.13\textwidth]{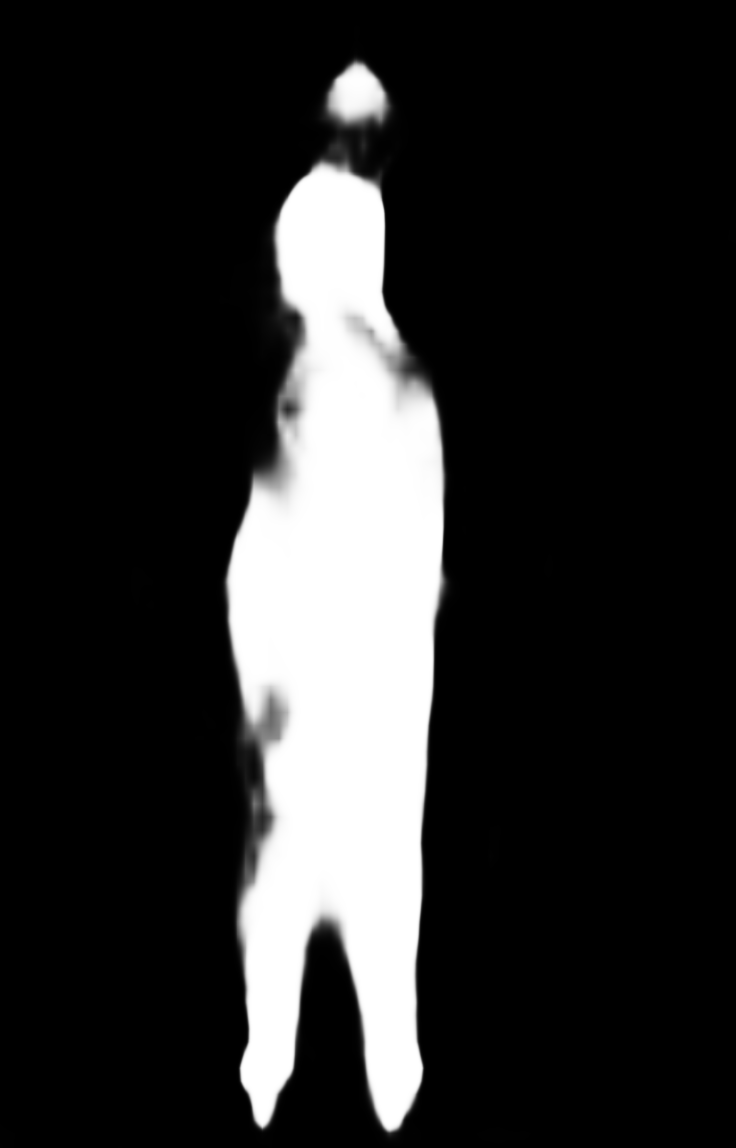}\\[0.1em]
      \includegraphics[width=0.13\textwidth, height=0.13\textwidth]{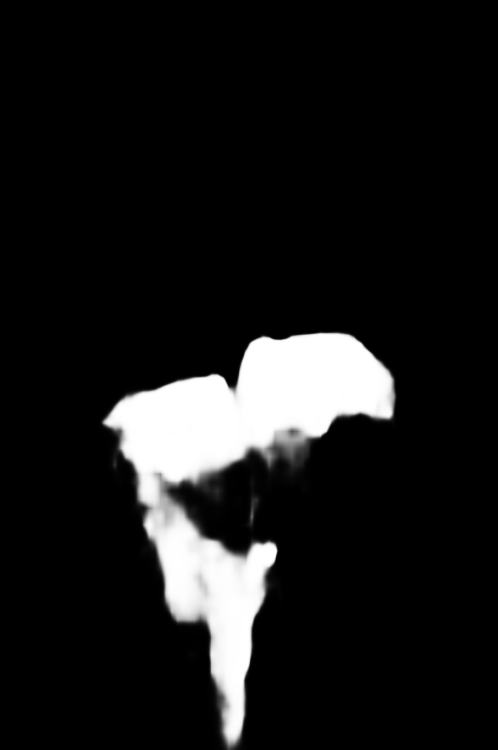}
    \end{tabular}}%
  \hspace{0.1em}%
  \subcaptionbox{\scriptsize{RankNet}}{%
    \begin{tabular}{c}
      \includegraphics[width=0.13\textwidth, height=0.13\textwidth]{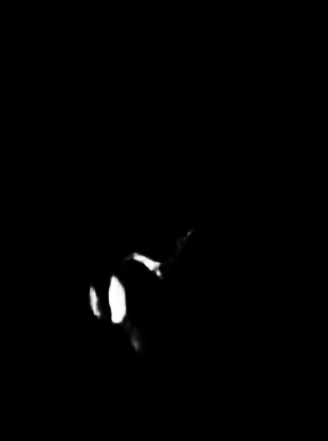} \\[0.1em]
      \includegraphics[width=0.13\textwidth, height=0.13\textwidth]{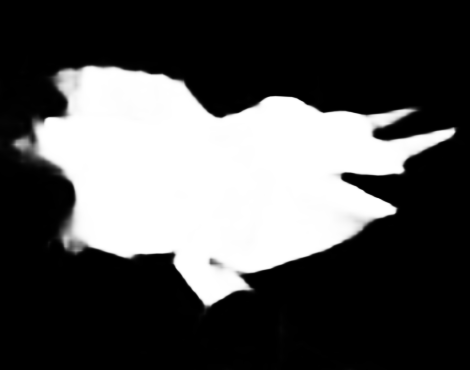} \\[0.1em]
      \includegraphics[width=0.13\textwidth, height=0.13\textwidth]{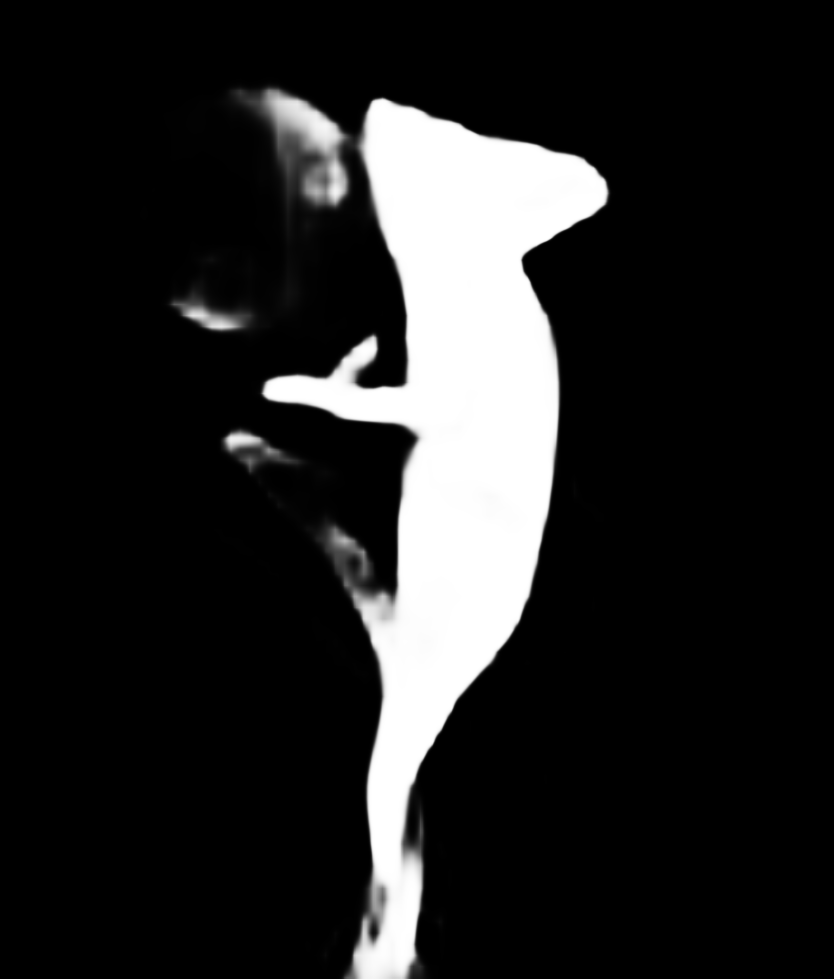}\\[0.1em]
      \includegraphics[width=0.13\textwidth, height=0.13\textwidth]{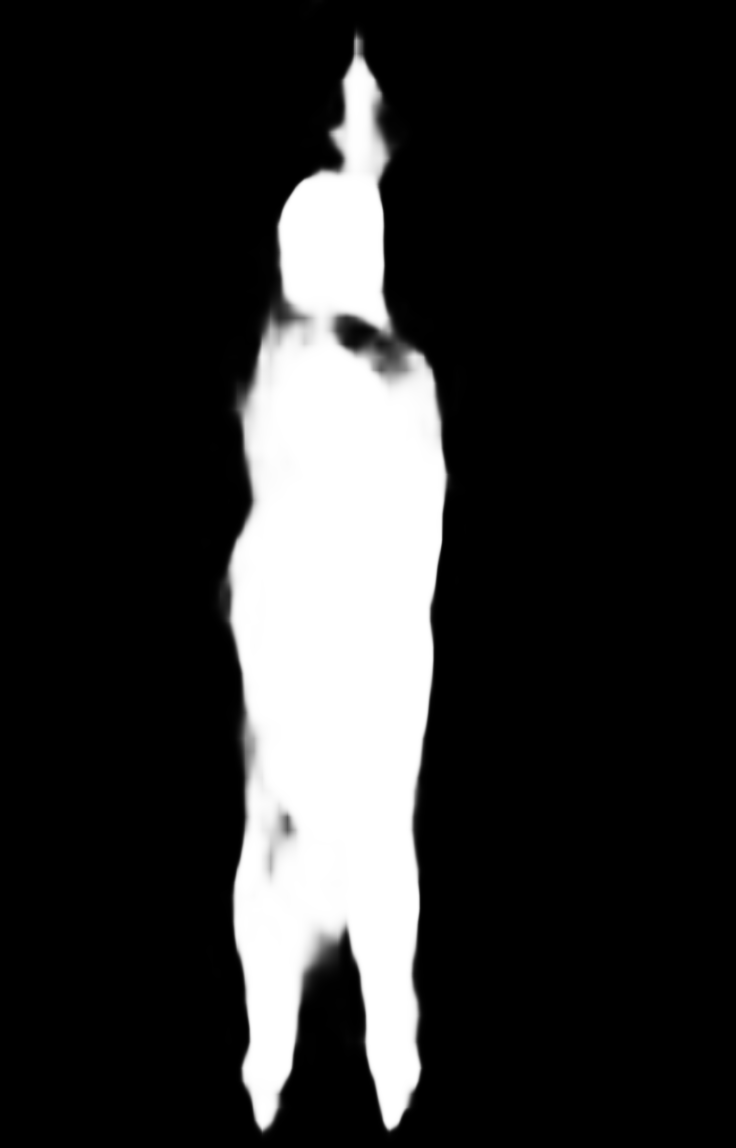}\\[0.1em]
      \includegraphics[width=0.13\textwidth, height=0.13\textwidth]{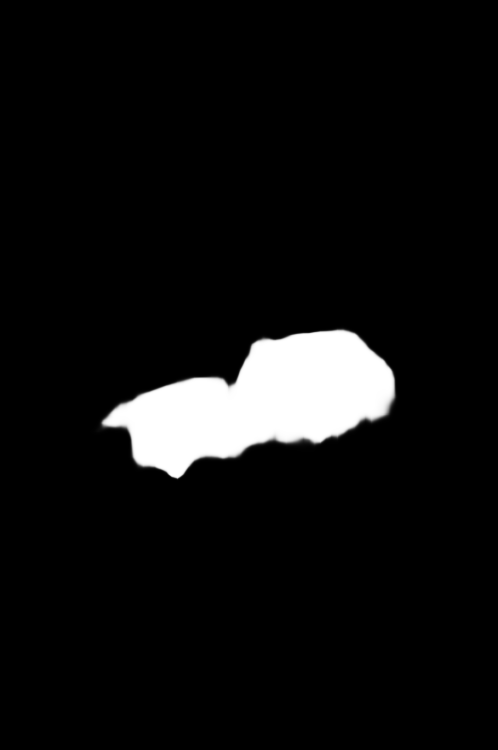}
    \end{tabular}}%
  \hspace{0.0em}%
  
\caption{More results on CAMO~\cite{le2019anabranch} dataset for camouflaged object detection. We compare to SINet~\cite{fan2020camouflaged}, PFNEt~\cite{mei2021camouflaged}, JCOD~\cite{li2021uncertainty} and RankNet~\cite{lv2021simultaneously}.}
\label{fig:supp_cod}
\end{figure*}

We give more visual results of EVP and other task-specific methods on the four tasks in Figure~\ref{fig:supp_forgery},~\ref{fig:supp_shadow},~\ref{fig:supp_defocus}, and~\ref{fig:supp_cod} as supplementary to the visual results in the main paper.

{\small
\bibliographystyle{ieee_fullname}
\bibliography{11_references}
}